\documentclass{article}
\usepackage{authblk}   
\usepackage{graphicx} 
\usepackage[margin=1in]{geometry}
\usepackage{booktabs}
\usepackage[colorlinks=true, linkcolor=blue, urlcolor=blue, citecolor=blue]{hyperref}
\usepackage{xcolor}
\usepackage{soul} 
\usepackage{lscape}
\usepackage{natbib}
\usepackage{amsmath}
\usepackage[capitalise,nameinlink]{cleveref}
\usepackage{caption}
\usepackage{enumitem}
\usepackage{xspace}
\usepackage{listings}
\usepackage{algorithm}
\usepackage{tabularx}
\usepackage{tabulary}
\usepackage{algpseudocode}
\usepackage{parskip}  
\usepackage{pdfpages}
\usepackage{multirow}
\usepackage{wrapfig}
\usepackage{multicol}
\usepackage{placeins}
\usepackage{rotating}
\usepackage{subcaption}
\usepackage{float}

\newcommand{\spu}[1]{\textsuperscript{#1}}

\newcommand\rolloutStartDate{January 30 2025\xspace}
\newcommand\rolloutEndDate{April 18 2025\xspace}
\newcommand\inductionEndDate{February 28 2025\xspace}
\newcommand\restStudyStartDate{March 1 2025\xspace}
\newcommand\nTotalVisits{87931\xspace}
\newcommand\nEligibleVisits{52409\xspace}
\newcommand\nIneligibleVisits{35522\xspace}
\newcommand\nOutsideNairobiCounty{3878\xspace}
\newcommand\nIneligibleVisitType{29210\xspace}
\newcommand\nNotSeenByProvider{2434\xspace}
\newcommand\nConsentedVisits{40745\xspace}
\newcommand\nSeenByProviderInBothGroups{1166\xspace}

\newcommand\nSilentVisits{18990\xspace}
\newcommand\nActiveVisits{20589\xspace}
\newcommand\nDocsSilentVisits{18990\xspace}
\newcommand\nOutcomesSilentVisits{7331\xspace}
\newcommand\nDocsActiveVisits{20589\xspace}
\newcommand\nOutcomesActiveVisits{7918\xspace}
\newcommand\pctEligible{59.6\%\xspace}
\newcommand\pctConsented{77.7\%\xspace}
\newcommand\pctIneligible{40.4\%\xspace}
\newcommand\pctOutsideNairobiCounty{4.4\%\xspace}
\newcommand\pctIneligibleVisitType{33.2\%\xspace}
\newcommand\pctNotSeenByProvider{2.8\%\xspace}

\newcommand\pctSeenByProviderInBothGroups{2.2\%\xspace}
\newcommand\pctSilentVisits{46.6\%\xspace}
\newcommand\pctDocsSilentVisits{100.0\%\xspace}
\newcommand\pctOutcomesSilentVisits{38.6\%\xspace}
\newcommand\pctActiveVisits{50.5\%\xspace}
\newcommand\pctDocsActiveVisits{100.0\%\xspace}
\newcommand\pctOutcomesActiveVisits{38.5\%\xspace}

\newcommand\pctEastlandsSilent{43.6\%\xspace}
\newcommand\pctEastlandsActive{38.4\%\xspace}
\newcommand\pctSouthwestSilent{22.2\%\xspace}
\newcommand\pctSouthwestActive{18.8\%\xspace}
\newcommand\pctThikaSilent{34.2\%\xspace}
\newcommand\pctThikaActive{42.8\%\xspace}
\newcommand\nSilentAI{49\xspace}
\newcommand\nActiveAI{57\xspace}

\newcommand\visitsPerProviderMedSilent{428\xspace}

\newcommand\visitsPerProviderMedActive{395\xspace}

\newcommand\nPhysicianReviewers{108\xspace}
\newcommand\nPhysicianReviewersKenya{29\xspace}

\newcommand\nUniqueVisitsHumanRated{5666\xspace}
\newcommand\nVisitsSingleRated{4279\xspace}
\newcommand\nVisitsDoubleRated{1387\xspace}

\newcommand\inductionRRRTreatment{4.3\%\xspace}
\newcommand\inductionRRRTreatmentLowCI{-3.0\%\xspace}
\newcommand\inductionRRRTreatmentHighCI{11.1\%\xspace}

\newcommand\mainPeriodLikertErrorRatesHistoryPVal{0.000\xspace}

\newcommand\mainPeriodLikertErrorRatesRRRHistory{31.8\%\xspace}
\newcommand\mainPeriodLikertErrorRatesRRRHistoryLowCI{21.9\%\xspace}
\newcommand\mainPeriodLikertErrorRatesRRRHistoryHighCI{40.5\%\xspace}

\newcommand\mainPeriodLikertErrorRatesInvestigationsPVal{0.034\xspace}

\newcommand\mainPeriodLikertErrorRatesRRRInvestigations{10.3\%\xspace}
\newcommand\mainPeriodLikertErrorRatesRRRInvestigationsLowCI{1.0\%\xspace}
\newcommand\mainPeriodLikertErrorRatesRRRInvestigationsHighCI{18.8\%\xspace}

\newcommand\mainPeriodLikertErrorRatesDiagnosisPVal{0.001\xspace}

\newcommand\mainPeriodLikertErrorRatesRRRDiagnosis{16.0\%\xspace}
\newcommand\mainPeriodLikertErrorRatesRRRDiagnosisLowCI{6.9\%\xspace}
\newcommand\mainPeriodLikertErrorRatesRRRDiagnosisHighCI{24.2\%\xspace}

\newcommand\mainPeriodLikertErrorRatesNNTDiagnosis{18.1\xspace}

\newcommand\mainPeriodLikertErrorRatesRRRTreatment{12.7\%\xspace}
\newcommand\mainPeriodLikertErrorRatesRRRTreatmentLowCI{6.8\%\xspace}
\newcommand\mainPeriodLikertErrorRatesRRRTreatmentHighCI{18.3\%\xspace}

\newcommand\mainPeriodLikertErrorRatesNNTTreatment{13.9\xspace}

\newcommand\totalErrorReductionDiagnosis{22102\xspace}
\newcommand\totalErrorReductionTreatment{28880\xspace}

\newcommand\mainPeriodRedOnlyRRRDiagnosis{31.5\%\xspace}

\newcommand\mainPeriodRedOnlyRRRTreatment{18.0\%\xspace}

\newcommand\redVYellowFirstGroupDiagnosisRate{46.2\%\xspace}
\newcommand\redVYellowFirstGroupDiagnosisLowerCI{40.7\%\xspace}
\newcommand\redVYellowFirstGroupDiagnosisUpperCI{51.7\%\xspace}

\newcommand\redVYellowSecondGroupDiagnosisRate{35.3\%\xspace}
\newcommand\redVYellowSecondGroupDiagnosisLowerCI{32.8\%\xspace}
\newcommand\redVYellowSecondGroupDiagnosisUpperCI{37.8\%\xspace}

\newcommand\yellowVGreenSecondGroupDiagnosisRate{20.5\%\xspace}
\newcommand\yellowVGreenSecondGroupDiagnosisLowerCI{18.4\%\xspace}
\newcommand\yellowVGreenSecondGroupDiagnosisUpperCI{22.9\%\xspace}

\newcommand\fleissKappaHistory{0.260\xspace}
\newcommand\fleissKappaInvestigations{0.285\xspace}
\newcommand\fleissKappaDiagnosis{0.232\xspace}
\newcommand\fleissKappaTreatment{0.223\xspace}

\newcommand\historyLikertOnePointHigh{79.4\%\xspace}
\newcommand\historyLikertOnePointLow{76.2\%\xspace}
\newcommand\historyLikertOnePointProp{77.8\%\xspace}

\newcommand\investigationsLikertOnePointHigh{67.8\%\xspace}
\newcommand\investigationsLikertOnePointLow{64.2\%\xspace}
\newcommand\investigationsLikertOnePointProp{66.0\%\xspace}

\newcommand\diagnosisLikertOnePointHigh{70.8\%\xspace}
\newcommand\diagnosisLikertOnePointLow{67.3\%\xspace}
\newcommand\diagnosisLikertOnePointProp{69.1\%\xspace}

\newcommand\treatmentLikertOnePointHigh{68.9\%\xspace}
\newcommand\treatmentLikertOnePointLow{65.3\%\xspace}
\newcommand\treatmentLikertOnePointProp{67.1\%\xspace}

\newcommand\mainStudyGEEhistoryRRRPoint{25.3\%\xspace}

\newcommand\mainStudyGEEinvestigationsRRRPoint{9.8\%\xspace}

\newcommand\mainStudyGEEdiagnosisRRRPoint{16.8\%\xspace}

\newcommand\mainStudyGEEtreatmentRRRPoint{12.2\%\xspace}

\newcommand\seekCareGivenInappropriateMedsRate{7.9\%\xspace}
\newcommand\seekCareGivenInappropriateMedsRateLowerCI{6.7\%\xspace}
\newcommand\seekCareGivenInappropriateMedsRateUpperCI{9.4\%\xspace}
\newcommand\seekCareGivenNotInappropriateMedsRate{12.3\%\xspace}
\newcommand\seekCareGivenNotInappropriateMedsRateLowerCI{10.4\%\xspace}
\newcommand\seekCareGivenNotInappropriateMedsRateUpperCI{14.4\%\xspace}
\newcommand\seekCareVsAppropriateTableP{0.000\xspace}

\newcommand\GPTFourOneRRRTreatment{21.5\%\xspace}
\newcommand\GPTFourOneRRRTreatmentLowCI{19.4\%\xspace}
\newcommand\GPTFourOneRRRTreatmentHighCI{23.7\%\xspace}

\newcommand\GPTFourOneHistoryWithinOneAgreement{87.0\%\xspace}

\newcommand\GPTFourOneHistoryFleissKappa{0.283\xspace}

\newcommand\OThreeRRRTreatment{19.1\%\xspace}
\newcommand\OThreeRRRTreatmentLowCI{17.1\%\xspace}
\newcommand\OThreeRRRTreatmentHighCI{21.1\%\xspace}

\newcommand\OThreeHistoryWithinOneAgreement{86.6\%\xspace}

\newcommand\OThreeHistoryFleissKappa{0.306\xspace}

\newcommand\notFeelingBetterFirstGroupRate{4.3\%\xspace}
\newcommand\notFeelingBetterFirstGroupLowerCI{3.7\%\xspace}
\newcommand\notFeelingBetterFirstGroupUpperCI{4.9\%\xspace}

\newcommand\notFeelingBetterSecondGroupRate{3.8\%\xspace}
\newcommand\notFeelingBetterSecondGroupLowerCI{3.3\%\xspace}
\newcommand\notFeelingBetterSecondGroupUpperCI{4.4\%\xspace}
\newcommand\notFeelingBetterPVal{0.234\xspace}

\newcommand\nSilentAISurveyResponses{23\xspace}
\newcommand\nActiveAISurveyResponses{36\xspace}
\newcommand\pctSilentAISurveyResponses{47\%\xspace}
\newcommand\pctActiveAISurveyResponses{63\%\xspace}
\newcommand\emrQualityUPVal{0.046\xspace}

\newcommand\activeAISpecificNPS{78\xspace}
\newcommand\activeAISatisfactionPctFour{58\%\xspace}
\newcommand\activeAISatisfactionPctFive{42\%\xspace}

\newcommand\activeAISpecificQualityPctFive{75\%\xspace}
\newcommand\nonAIMedianDuration{13.01\xspace}
\newcommand\AIMedianDuration{16.43\xspace}
\newcommand\MedianDurationP{0.000\xspace}
\newcommand\nAICalls{155450\xspace}
\newcommand\nAICallsThumbsAny{19493\xspace}
\newcommand\pctAICallsThumbsAny{12.5\%\xspace}
\newcommand\nAICallsThumbsUp{18424\xspace}
\newcommand\nAICallsThumbsDown{1069\xspace}
\newcommand\pctAICallsThumbsUp{94.5\%\xspace}
\newcommand\pctAICallsThumbsDown{5.5\%\xspace}

\title{AI-based Clinical Decision Support \\for Primary Care: A Real-World Study}

\author{%
Robert Korom\spu{*,1},
Sarah Kiptinness\spu{*,1},
Najib Adan\spu{1},
Kassim Said\spu{1},
Catherine Ithuli\spu{1},
Oliver Rotich\spu{1},
Boniface Kimani\spu{1},
Irene King'ori\spu{1},
Stellah Kamau\spu{1},
Elizabeth Atemba\spu{1},
Muna Aden\spu{2},
Preston Bowman\spu{3},
Michael Sharman\spu{3},
Rebecca Soskin Hicks\spu{3},
Rebecca~Distler\spu{3},
Johannes Heidecke\spu{3},
Rahul K.\ Arora\spu{*,3},
Karan Singhal\spu{*,3}
\\
\spu{1}Penda Health \qquad
\spu{2}Nairobi County \qquad
\spu{3}OpenAI
}
\date{}  
\usepackage{natbib}
\setcitestyle{authoryear,open={(},close={)}}

\begin{document}

\maketitle

\begingroup
  \renewcommand{\thefootnote}{\fnsymbol{footnote}}
  \footnotetext[1]{Correspondence to: \texttt{robert@pendahealth.com sarah.kiptinness@pendahealth.com rahul@openai.com karan@openai.com}}
\endgroup

\begin{abstract}
    We evaluate the impact of large language model-based clinical decision support in live care. In partnership with Penda Health, a network of primary care clinics in Nairobi, Kenya, we studied \emph{AI Consult}, a tool that serves as a safety net for clinicians by identifying potential documentation and clinical decision-making errors. AI Consult integrates into clinician workflows, activating only when needed and preserving clinician autonomy. We conducted a quality improvement study, comparing outcomes for 39,849 patient visits performed by clinicians with or without access to AI Consult across 15 clinics. Visits were rated by independent physicians to identify clinical errors. Clinicians with access to AI Consult made relatively fewer errors: 16\% fewer diagnostic errors and 13\% fewer treatment errors. In absolute terms, the introduction of AI Consult would avert diagnostic errors in 22,000 visits and treatment errors in 29,000 visits annually at Penda alone. In a survey of clinicians with AI Consult, all clinicians said that AI Consult improved the quality of care they delivered, with 75\% saying the effect was ``substantial''. These results required a clinical workflow-aligned AI Consult implementation and active deployment to encourage clinician uptake. We hope this study demonstrates the potential for LLM-based clinical decision support tools to reduce errors in real-world settings and provides a practical framework for advancing responsible adoption.\footnote{This quality improvement study was conducted with the approval of Kenya's Ministry of Health, Kenya's Digital Health Agency, Nairobi County, and AMREF Health Africa Ethical and Scientific Review Committee (approval ESRC P1795/2024).}
\end{abstract}

\section{Introduction}
Artificial intelligence (AI) systems have the potential to widen access to reliable health information and high-quality care \citep{beam2018bigdata,topol2019highperf,rajkomar2019machine}. Large language models (LLMs) have recently experienced significant leaps in performance, reliability, and safety for health applications \citep{arora2025healthbench,nori2025sequential,singhal2023large,singhal2025toward}. These advances suggest new opportunities for improving healthcare delivery—including supporting clinicians in delivering better care.

Despite research progress, scaled real-world deployment of AI tools in clinical environments remains limited. State-of-the-art LLMs now often outperform physicians on benchmarks \citep{goh2025gpt,arora2025healthbench,nori2025sequential,vanveenAdaptedLargeLanguage2024b}, but these gains have yet to translate into measurable benefits for patients and clinicians in live care settings. The most critical bottleneck in the health AI ecosystem is no longer better models, but rather the \emph{model-implementation gap}: the chasm between model capabilities and real-world implementation.

Closing the model-implementation gap necessitates the responsible study of LLM implementations in frontier health AI use cases. One example is clinical decision support (CDS) systems \citep{suttonOverviewClinicalDecision2020,middletonClinicalDecisionSupport2016}, which provide clinicians with relevant knowledge at the point of care. Efforts to measure how well LLMs can help with clinical decisions so far have used offline evaluations, often measuring model capabilities on clinical vignettes without capturing the unique challenges of designing and deploying an implementation for real-world care \citep{benaryLeveragingLargeLanguage2023,goh2025gpt,onianiEnhancingLargeLanguage2024}.

\begin{figure}[t]
  \centering
  \includegraphics[width=\linewidth]{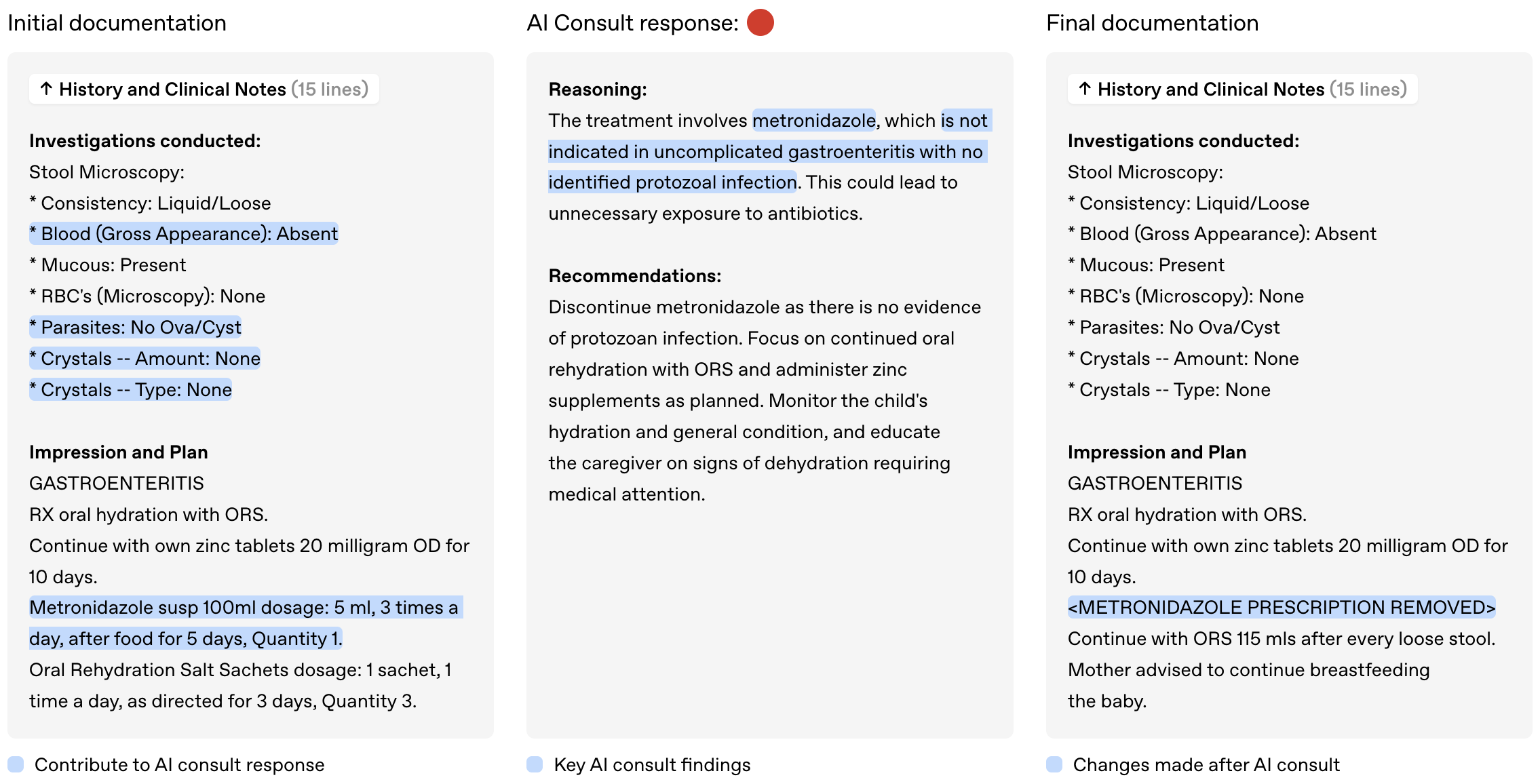}
  \caption{AI Consult is a safety net that runs in the background of a patient visit to identify potential errors. It was iteratively designed with clinicians, providing outputs with green/yellow/red severity and issuing alerts only when needed to reduce errors. In this example, AI Consult provided a red flag that helped a clinician identify and remove an unnecessary antibiotic prescription.}
  \label{fig:example}
\end{figure}

In this study, we examine the impact of an LLM-based clinical decision support tool in live care. Penda Health, where several authors are affiliated, is a network of high-volume clinics in Nairobi, Kenya that delivers 24-hour primary and urgent care to a broad range of Nairobi residents. We studied Penda's \emph{AI Consult}, which serves as a clinical safety net to prevent errors. The system is triggered asynchronously during key clinical workflow decision points in the electronic medical record (e.g., diagnosis, treatment). It surfaces guidance through a tiered traffic-light interface (green: no action, yellow: advisory, red: requires review), and is explicitly designed to minimize cognitive burden and preserve clinician autonomy. The tool was developed through iterative co-design with frontline clinicians and tailored to local epidemiology, Kenyan clinical guidelines, and Penda's care protocols.

To assess the tool’s impact, we conducted a pragmatic cluster-assigned study of 39,849 visits, comparing outcomes for patient visits managed by clinicians with and without access to AI Consult. We aimed to evaluate three primary domains: (i) clinical quality, as rated by independent physicians reviewing clinical documentation with patient identification removed; (ii) use and usability, based on a clinician survey and AI Consult usage data; and (iii) patient-reported outcomes collected via routine follow-up calls. We find meaningful reductions in clinical errors for clinicians with the tool (``AI group'') vs those without (``non-AI group'') and encouraging feedback from clinicians using AI Consult. We did not detect a significant difference in patient-reported outcomes. This study was conducted with the approval and consultation of Kenya's Ministry of Health, Kenya's Digital Health Agency, Nairobi County, AMREF Health Africa Ethical and Scientific Review Committee, Kenya's National Commission for Science, Technology and Innovation (NACOSTI), and other local stakeholders to ensure it aligned with national priorities, ethical standards, and data protection requirements.

This study makes three key contributions:
\begin{itemize}
    \item We describe a live deployment and study of an LLM-powered clinical decision support tool across 39,849 patient visits, 106 clinicians, and 15 clinics.
    \item We report findings:
        \begin{itemize}[itemsep=0.25em, parsep=0em, topsep=0em, partopsep=0em]
          \item We observe significant relative reductions in errors, including 32\% for history-taking errors (number needed to treat [NNT] 11.3), 10\% for investigation errors (NNT 27.8), 16\% for diagnostic errors (NNT 18.1), and 13\% for treatment errors (NNT 13.9) for clinicians in the AI vs the non-AI group. In absolute terms, the introduction of AI Consult would avert diagnostic errors in 22,000 visits and treatment errors in 29,000 visits annually at Penda alone.
          \item AI group clinicians saw significant reductions in important clinical failure modes, including incorrect primary diagnosis, inappropriate medications, missing patient education and follow-up plan, key history details missing, and key investigations missing.
          \item The effect of the tool became more pronounced after an initial induction period, when Penda rolled out active strategies to drive clinician uptake.
          \item LLMs evaluating study visits found a greater difference in clinical errors between the AI group and the non-AI group than the difference found by physician evaluators (e.g., 22\% reduction in treatment errors and 19\% in diagnostic errors according to \texttt{GPT-4.1}).
          \item In routine follow-up calls, 3.8\% of patients treated by AI group clinicians said they were not feeling better, compared to 4.3\% for the non-AI group, a difference that was not statistically significant.
          \item{All survey respondents in the AI group said AI Consult helped them improve the quality of care they could deliver, with 75\% saying the effect was ``substantial''.}
          \item{Over the study, AI group clinicians learned to avoid ``red'' outputs even before receiving them (the fraction of AI group visits with initial red outputs decreased from 45\% to 35\% during the study), suggesting the tool helped clinicians improve their own practice.}
          \item{In patient safety reports, there were no cases where AI Consult advice actively caused harm.}
        \end{itemize}
    \item We describe the key factors for success: a capable model, a clinically-aligned implementation, and active deployment strategies.
\end{itemize}

This work offers an early demonstration of the potential for LLM-based tools to serve as real-time copilots for delivering care and a practical framework for advancing responsible adoption in real-world health systems.

\section{Background}

\subsection{Primary care}
Primary care clinicians see patients across every age group, organ system, and disease type, often in the same day, requiring broad knowledge. The breadth of practice contributes to primary care quality challenges worldwide, with the WHO reporting substantial rates of preventable patient harm \citep{PatientSafety}. This suggests that AI systems could be especially useful in primary care.

In Kenya, primary care is largely delivered by clinical officers: clinicians who complete three years of academic training followed by a one-year supervised internship. They manage the full breadth of acute and chronic conditions across the life course. Structural challenges in Kenyan primary care (late presentation, high patient volumes, limited diagnostics) compound this wide scope of practice to create a sizable quality gap: Studies suggest low adherence to national guidelines by healthcare workers across multiple levels of Kenya's healthcare system, with frequent errors such as missed comorbidities, antibiotic overprescription, and diagnostic delays \citep{marete2020clinicians,kruger2017imci,kiener2025antibiotic}.

\subsection{Penda Health}

Penda Health is a Nairobi-based social enterprise founded in 2012 that delivers comprehensive, 24-hour primary and urgent care services through a network of fully-licensed medical centers distributed across the city. The organization presently operates 16 clinics and records over 1000 patient visits a day, supported by a clinical workforce of more than 100 licensed clinical officers. For a video and photos depicting Penda's care context and AI Consult, see \href{https://openai.com/index/ai-clinical-copilot-penda-health/}{the blog post} that accompanies this paper.

\subsection{Digital infrastructure and clinical decision support at Penda}
\label{sec:digital-infra-penda}

Penda has invested substantially in its digital infrastructure and quality improvement programs over the years, and has been a pioneer in implementing clinical decision support tools.

\paragraph {Electronic medical record (2017).} A cloud-hosted electronic medical record (EMR), Easy Clinic, was introduced in 2017, supporting all patient visits and enabling real-time monitoring of quality metrics and operations.

\paragraph {Rule-based system (2019-2020).} Penda implemented an early non-AI CDS system before its first iteration of AI Consult \citep{Korom2020}. In this system, decision trees embedded in the EMR provided point-of-care reminders for some common conditions. Similar early approaches have been employed and studied in other contexts for many years \citep{papadopoulos2022systematic,brightEffectClinicalDecisionSupport2012,musenClinicalDecisionSupportSystems2021}. 

The rule-based system and concurrent quality improvement efforts had a large effect. Within 12 months of deployment, guideline adherence at Penda rose from the national baseline of 40\% to over 90\%. While the rule-based system was very effective for improving adherence to specific national practice guidelines, it was narrow in scope: there was still a significant quality gap in clinical officers' history taking, diagnostic accuracy, and patient management. Rule-based systems struggled to effectively support the wide variety of situations a Penda clinician faces daily.

\paragraph {AI Consult v1 (February 2024).} Penda Health implemented an early version of an LLM copilot prior to the version studied in this work. AI Consult v1 provided feedback from an LLM on the current visit at clinician request. Clinicians clicked a button within the EMR during a patient visit, chose an area to receive feedback on (including documentation, patient management, and overall visit), and received structured feedback from an LLM. Similar to Penda's other CDS iterations, clinicians reviewed the output of the tool and made all clinical decisions.

During the early deployment of AI Consult v1, Penda performed an internal safety audit of 100 randomly selected cases. Each of these cases included (1) patient documentation state before AI Consult use, (2) AI Consult response, and (3) final documentation state, including any changes resulting from AI Consult. These cases were reviewed by Penda's quality team. Each AI Consult output was scored from 1–5, where 5 was outstanding feedback from the LLM on the case (relevant, locally-appropriate, comprehensive, and actionable); 3 was neutral; and 1 was actively harmful (e.g., encouraging the clinician to perform unnecessary tests, offering an inappropriate diagnosis, or an incorrect or not locally appropriate treatment plan). Cases were also annotated with qualitative notes on how clinicians may have acted on AI Consult responses.

In that audit, Penda assigned 64 outputs a rating of 5, 21 a rating of 4, and 14 a rating of 3 (one visit lacked sufficient clinical documentation to be analyzed). No AI responses were unsafe, and the team did not find any instances where the effect of AI Consult was harmful. Penda did find some qualitative improvements in care after clinicians received AI feedback. 

Despite showing early promise in terms of patient safety and quality improvement, AI Consult v1 only achieved adoption in about 60\% of visits. Qualitative notes showed many cases in which AI feedback was not heeded despite being correct and clinically actionable. There was a need to further optimize the AI Consult workflow to seamlessly intervene at key decision points without creating alert fatigue, and to work closely with clinician users to increase uptake. These analyses gave Penda’s quality team the confidence they needed to further develop and test AI Consult.

\paragraph {AI Consult v2 (January 2025).} To create a universal “safety net” in the EMR workflow without increasing cognitive load, AI Consult was re-engineered to run silently in the background at key workflow inflection points (documentation of vitals and chief complaint, documentation of history and physical examinations, ordering diagnostic tests, diagnosis, management plan). Outputs are surfaced through a traffic-light interface: green: no action, yellow: advisory, red: mandatory review before proceeding. This design couples high coverage with minimal interruption, and leaves the clinician with ultimate control over all clinical decisions. For a video of AI Consult, see \href{https://openai.com/index/ai-clinical-copilot-penda-health/}{the blog post} that accompanies this paper. In this work, we refer to this version of the tool as ``AI Consult''. The tool is described further in \cref{sec:methods-ai-consult}.

\subsection{Motivation for the present study}

Collectively, Penda's large patient volumes and a highly variable disease mix, in combination with its strong quality program, digital maturity, and CDS experience make Penda's clinics an informative setting for evaluating the impact of LLM-based clinical decision support on patient care.

In addition to evaluating the impact of LLMs in real-world settings, this study focuses on two additional factors driving the successful uptake of AI-based CDS: \emph{clinically-aligned implementation} (\cref{sec:clinically-aligned-implementation}), or highly iterative development of a tool well-integrated into clinical workflows,  and \emph{active deployment} (\cref{sec:methods-active-deployment}), or strategies to build clinician understanding of and buy-in for a tool. We find all three factors (model performance, clinically-aligned implementation, and active deployment) are crucial for successful implementation and adoption.

\section{Methods}

Here, we describe AI Consult, how Penda integrated it into its clinical workflow, Penda's rollout of the tool to half of clinicians, and the design and methods for our study of that rollout. For images of AI Consult, see \cref{fig:ss_yellow_click,fig:ss_red,fig:ss_green_main,fig:ss_green_click,fig:ss_yellow_main}. For a video of AI Consult, see \href{https://openai.com/index/ai-clinical-copilot-penda-health/}{the blog post} that accompanies this paper.

\subsection{AI Consult}
\label{sec:methods-ai-consult}

\paragraph{Design rationale.} Penda’s AI Consult tool is conceived as a continuously-running safety net. Its core objectives are to:
\begin{enumerate}
    \item \textbf{Maximize coverage:} the model reviews every visit and each major decision node, and this review does not require active clinician requests.
    \item \textbf{Minimize cognitive load:} model feedback interrupts the clinical workflow only when it identifies material risk.
    \item \textbf{Maintain clinician autonomy:} the system issues recommendations, but all final decisions remain the clinician's.
\end{enumerate}

There are three types of responses that can be returned, following a three-color traffic light interface:
\begin{itemize}
    \item Green: indicates no concerns; appears as a green checkmark.
    \item Yellow: indicates moderate concerns; appears as a yellow ringing bell that clinicians can choose whether to view.
    \item Red: indicates safety-critical issues; appears as a pop-up that clinicians are required to view and acknowledge before continuing.
\end{itemize}

Classification thresholds must balance sensitivity against alert fatigue. The traffic-light approach helps create this balance: red alerts are cases with high probability or severity of harm, meaning that alerts are likely to be true positives and therefore can safely interrupt the clinician workflow. Yellow events are in an ambiguous middle region, and the bell helps engage clinician judgment without interrupting. Green events confirm that AI Consult is running correctly while fading into the background. 

\paragraph{Asynchronous, event-driven architecture.} AI Consult is embedded in Penda’s cloud-hosted EMR (Easy Clinic). The EMR triggers AI Consult calls in response to predefined events: whenever the user finishes typing and navigates away from a critical field (i.e., chief complaint, clinical notes, investigations, diagnosis, and medications), AI Consult will run in the background with the documentation state up until that point and return a response to the clinician.

\paragraph{Prompt engineering.} The LLM prompt contextualizes the patient visit and contains Penda-specific context as well as a summary of relevant clinical practice guidelines. It then includes the task for the model, the definition of each clinical category, and few-shot examples of red, yellow, and green responses for each category. The model is asked to return a color (alert severity level), a rationale for that color, and an action for the clinician to consider. For the full prompts used in AI Consult, see \cref{app:ai_consult_prompts}.

\paragraph{Model-agnostic.} The design of AI Consult—in particular, its reliance on a prompted general model rather than a specialized or fine-tuned model—permits any model to easily be used. This allows more performant models, cheaper models, or models that meet other specific needs to easily be substituted in.

Penda opted to use GPT-4o as the default model for AI Consult due to its strong few-shot reasoning and low latency. At the time of the study, more performant models like \texttt{GPT-4.1}, \texttt{o3}, and \texttt{o4-mini} were not yet available. While reasoning models offer advantages in terms of nuanced performance on challenging health-related questions, Penda found that minimizing latency was more important for the tool to give feedback timely enough to be actionable to ensure clinician adoption, and thus maximize downstream impact of the tool.

\paragraph{Development.} AI Consult was developed in partnership with Penda's EMR vendor and Penda's clinical quality and IT teams.

\subsection{Iteration towards clinically-aligned implementation}
\label{sec:clinically-aligned-implementation}

To achieve the clinically-aligned implementation of AI Consult used in the present study, Penda went through numerous iterative development cycles. Penda's clinical quality team initially documented the proposed product specifications and user acceptance criteria for the end-to-end tool. After initial development and prior to deployment to the production environment, Penda's clinical quality and IT teams used, tested, and red-teamed AI Consult extensively in order to maximize safety and usability for frontline clinicians. During the study's induction period (described in \cref{sec:methods-quality-improvement-rollout}), Penda's teams also continued to iterate on AI Consult with real user feedback.

Hundreds of design decisions were made during this process; here, we document the most important categories.

\paragraph{AI Consult triggers.}
A fundamental challenge in designing AI Consult is knowing when to call the model (i.e., when to have the model review documentation and return a response). If it is called prematurely, feedback is returned before it is useful to the clinician. If it is called (or returns feedback) too late, the clinician's decision-making moment has passed, and it can be challenging to reverse decisions already made. Penda initially explored the possibility of making model calls only at the point that a patient is sent to a different physical location (e.g., pharmacy or laboratory).  Testing revealed that by then, the clinician has often already explained the next steps to the patient; it can be uncomfortable for clinicians to walk those next steps back if AI Consult recommendations conflict, making AI Consult less useful. After several iterations, Penda decided to trigger when users navigate away (``focus out'') from specific EMR fields.

One example of a specific technical challenge Penda faced in its implementation of triggers: if a user was typing in a decision-triggering box (say, the clinical notes) and a red response appeared for a previous workflow stage, acknowledging that red alert was considered an event which triggers AI Consult for the current workflow. This results in another model call, even if the clinician user was not done with that section. In testing, this behavior could lead to a painful cycle of red-alert pop-ups that were clinically inappropriate and led to alert fatigue. Identifying this in testing allowed engineers to modify the criteria for AI Consult triggers to exclude the acknowledgment of previous AI Consult alerts.

\paragraph{Threshold-setting.}
In live testing with clinicians, the overall usability of the tool was highly dependent on the red/yellow/green severity thresholds. When the threshold for problems is set too low, over-triggering of the system becomes apparent immediately and clinicians may begin to ignore alerts. 

Given the design of AI Consult, threshold-setting to avoid alert fatigue while still surfacing the most critical clinical problems is primarily a prompt engineering problem. Clear explanations and few-shot prompting allowed Penda to precisely define which gaps ought to trigger a red alert. For example, Penda included few-shot examples to ensure that missing vital signs would trigger red alerts. Vital signs are so critical to choosing diagnostic tests and making a diagnosis that a history and physical exam could not be considered complete if vital signs were absent. On the other hand, Penda had to moderate its expectations on the comprehensiveness of history and physical examination. In initial testing, red alerts were over-triggering for missing components of the clinical history. While the missing history components were not unreasonable, fully acting on these alerts would have required too dramatic of a shift in the documentation of history for Penda's practice setting, so a more lenient threshold was selected here.

\paragraph{User interface.}
Subtle design decisions for the user interface can substantially impact the user experience and adoption of a new tool. Penda's first iteration of the tool focused only on red-alert pop-ups for serious problems. However, many opportunities for clinical quality improvement are of intermediate (yellow) severity.  Pre-deployment testing showed a need to allow these quality improvement opportunities to surface to clinicians at the right moment without forcing a pop-up. Similarly, the initial UI did not include a green checkmark for green model outputs, which caused clinicians to wait in case a yellow or red alert was incoming. The final iteration of AI Consult included this green checkmark to reduce cognitive overhead for clinicians.

Penda also made final implementation improvements early in the rollout of the AI Consult tool, as described in \cref{sec:methods-quality-improvement-rollout}.

\subsection{Quality improvement rollout}
\label{sec:methods-quality-improvement-rollout}

Penda's audit of AI Consult safety (\cref{sec:digital-infra-penda}) and the design of AI Consult as a safety net, with all final decisions made by Penda clinicians, gave the Penda leadership team the confidence in AI Consult it needed to pilot the tool more broadly. As part of its quality improvement practice, Penda decided to roll out AI Consult to half of its clinicians from \rolloutStartDate to \rolloutEndDate. It rolled out AI Consult at the clinician level: half of clinicians in each clinic were randomly assigned to have access to AI Consult (AI group), while their colleagues did not (non-AI group).\footnote{Clinicians were split at the clinic-level across 15 clinics because different Penda clinics serve populations with different demographics. The way Penda implemented this was equivalent to randomly allocating clinicians stratified by clinic, with block size 2 and a 1:1 allocation ratio at the clinic level. Note that some randomized providers left Penda before the rollout started and are therefore not included in the study analysis.} As a further assurance during this rollout, Penda actively monitored model outputs throughout the course of the study through its established patient safety reporting process, rapidly reviewing any case where a patient experienced an adverse event. This process found no cases where a model recommendation directly caused patient harm; see \cref{sec:patient-safety-reports} for full findings.

\begin{wrapfigure}
    {r}{0.55\textwidth}
    \vspace{-15pt}  
    \centering
    \includegraphics[width=\linewidth]{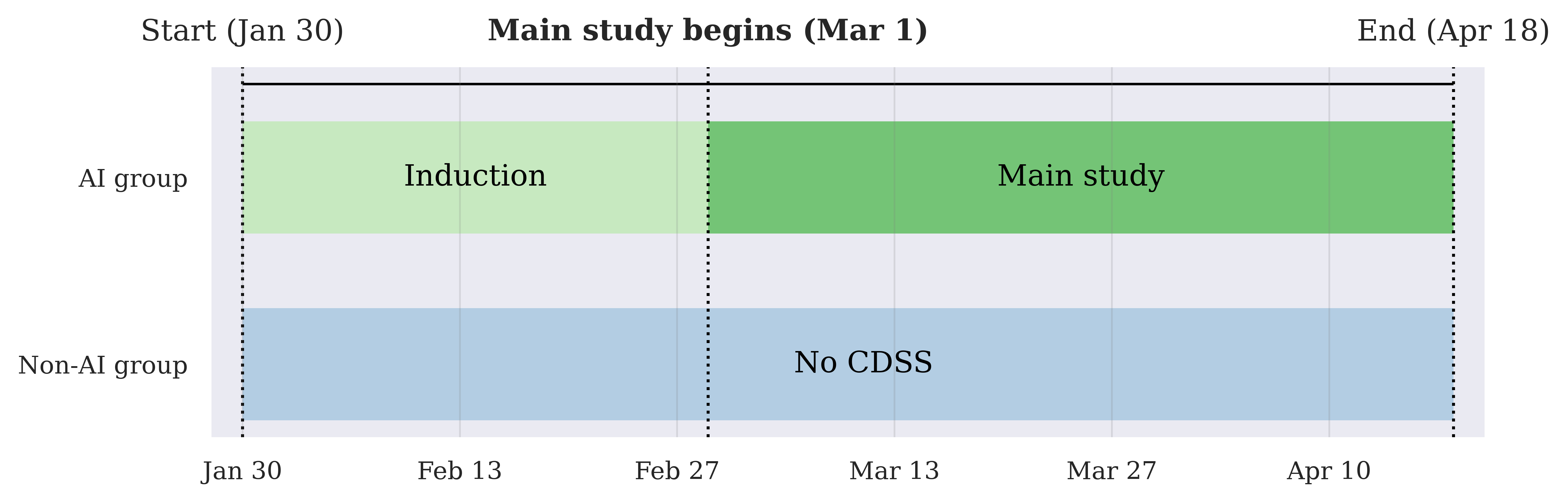}
    \caption{Timeline of AI Consult deployment and quality improvement evaluation.}
    \label{fig:timeline}
\end{wrapfigure}

The first part of this rollout, from \rolloutStartDate to \inductionEndDate, was an induction period for clinicians to familiarize themselves with the tool. It included up-front training but no active change management. The primary period of the quality improvement evaluation was from \restStudyStartDate to \rolloutEndDate, and included active deployment to drive adoption from Penda quality and branch leadership (\cref{sec:methods-active-deployment}). A diagram of the timeline is in \cref{fig:timeline}.

\paragraph{Shadow mode.}  AI Consult was also built with a \emph{shadow mode}, where AI Consult would operate as normal in the background of a visit, calling an LLM and logging responses, but no alerts (whether red, yellow, or green) were shown to clinicians. When Penda ultimately rolled out AI Consult to the AI group, it also used shadow mode for the non-AI group providers, enabling Penda to compare AI group clinicians with non-AI group clinicians by understanding triggers that \emph{would have} occurred if non-AI group clinicians had AI Consult. This is used in reporting several of the results in \cref{sec:results}.

\paragraph{Final implementation improvements during the induction period.}
Continuing to iterate on AI Consult after it was first deployed to clinicians in live care was also essential to make it as useful as possible. The first part of Penda's deployment was an induction period, intended to familiarize clinicians with the tool and to iterate given feedback.

Penda continuously collected feedback from clinicians through online surveys and virtual roundtable discussions. This feedback surfaced further areas for improvement on AI Consult thresholds and prompting to prevent alert fatigue. For example, this helped Penda ensure AI Consult did not ask for patient history not routinely collected at Penda or diagnostics unavailable in the Penda setting.

During this time, Penda also shadowed clinicians at every clinic to see how they engaged with the tool during their typical workflow. Clinician shadowing revealed a challenge not seen in pre-deployment testing: in real-world practice, many clinicians were facing system slowness that led to the AI Consult not providing near-real-time feedback. Due to a combination of technical factors, the time taken to return AI Consult responses had increased dramatically with the number of simultaneous API calls that were now being made. Penda re-engineered AI Consult's code to improve its speed and asynchronous functionality, allowing the API call to return, on average, in under three seconds.

\subsection{Active deployment}
\label{sec:methods-active-deployment}

Following the initial deployment of AI Consult to Penda's production EMR, Penda monitored the adoption of the tool, the safety and helpfulness of its outputs, and the extent to which clinicians acted on model outputs. Penda's approach had three pillars:
\begin{itemize}
    \item \textbf{Connection}: Peer champions and branch managers explained why the copilot mattered, walked colleagues through its strengths and limitations, and offered one-on-one coaching to support uptake.
    \item \textbf{Measurement}: Penda tracked how often clinicians interacted with AI Consult recommendations and reached out with personalized coaching.
    \item \textbf{Incentives}: Penda quality leadership recognized clinicians and clinics that used the tool well.
\end{itemize}

\paragraph{Connecting with clinicians to share AI Consult strengths and limitations.}
The Penda team made considerable efforts to connect clinicians with AI Consult's value. 

In Penda's continuing medical education sessions, clinical leaders identified real examples where an AI Consult clinician had received a red response and acted on it, and discussed with teams how this choice improved the quality of care delivery.  These examples built tangible buy-in for clinicians who could see in real practice and hear from their peers about how the tool improved quality. Penda also nominated high-performing peer champions at each clinic, who shared how they had learned to use AI Consult well and provided suggestions and feedback to other clinicians to encourage successful uptake.

Penda's work to connect with clinicians also identified other factors that made it difficult for clinicians to act on AI Consult.  For example, some clinicians were accustomed to documenting patient visits asynchronously, meaning a patient may have gone to receive medications before a provider documented medications and received AI Consult feedback. This and related patterns sometimes made AI Consult challenging to act on. Penda coached providers to document in real-time and trigger AI Consult before taking the next steps to ensure AI Consult recommendations were considered. While a major workflow change for some, this was essential to enabling clinicians to act on the feedback in real time.

\paragraph{Data and measurement.}
Penda's data infrastructure was crucial to building the metrics required for monitoring: the data backend allowed Penda clinical leadership to view AI inputs, red/yellow/green model outputs, and final patient documentation for over 8000 model calls per day. 

To summarize this data, Penda's clinical leadership team developed a single north star indicator: the ``left in red'' rate.  This metric tracked the fraction of patient visits where the final AI Consult model call for any category was red. Recall that when a red pop-up occurs, the clinician must acknowledge it, but then has the option to either leave things as they are or change a decision. If a decision is changed, AI Consult will run again, again returning a color for severity.  If the issue causing the red alert was addressed, AI Consult will likely return yellow or green, and would no longer be ``left in red.''  Thus, the left in red rate is a useful metric for understanding the extent to which clinicians with AI Consult are acting on the most severe alerts. A high left in red rate could reflect that clinicians were not seeing AI Consult alerts, that they were not reading these alerts, or that they were intentionally choosing not to act on the feedback from these results—each of which is valuable to understand to increase the tool's impact. Improving this single metric enabled Penda's team to identify and improve instances of each of these failure modes. 

In the first month of piloting the tool, Penda noticed that clinicians with AI Consult had only a slightly lower left in red rate compared to clinicians in the non-AI group (where the left in red rate could be calculated because AI Consult was running in shadow mode, with data logged but without outputs shown to providers). Penda reviewed AI Consult's red alerts and found them to be generally high quality, which made it concerning that clinicians were often not heeding AI recommendations. Penda therefore entered a period of active change management to further drive adoption.

\paragraph{Creating positive incentives.}
To socially incentivize use of AI Consult, Penda also shared individual left in red rates with each AI group clinician and included their decile of performance compared to their peers.  This approach provided positive encouragement for clinicians who were among the best in acting on AI Consult feedback. It also showed clinicians who were not acting on the AI Consult outputs that there was room for improvement relative to their peers—in many cases, clinicians were surprised about their relative performance. These steps, combined with peer champion weekly coaching feedback, helped Penda substantially reduce the left in red rate for clinicians with AI Consult (\cref{fig:reds_over_time_panels}).

\subsection{Study of AI Consult}

Penda and OpenAI embarked on a research study of the rollout across 15 clinics using routinely-collected patient documentation and outcomes. The study compared providers with and without access to AI Consult. We examined the effects of the tool on (i) quality of care, including diagnosis and treatment errors (using clinical documentation with patient identifiers removed); (ii) patient-reported outcomes (using routinely-collected patient outcomes data); and (iii) clinician workflows (using anonymous clinician surveys and clinical workflow data). 

\paragraph{Ethical considerations.} This study was approved by the AMREF Health Africa Ethical and Scientific Review Committee (approval number ESRC P1795/2024) and conducted under a research license from the National Commission for Science, Technology, and Innovation in Kenya (license number NACOSTI/P/25/415242). This research was also approved by the Ministry of Health in Kenya, Kenya's Digital Health Agency, and Nairobi County. 

Only patients who agreed to Penda's general patient consent form—which includes consent for use of data without patient identifiers for research purposes and to follow-up calls to collect patient-reported outcomes—were included in this analysis. Patients were also able to withdraw their consent for the use of data in this study until 15 days after the end of the study period.

Given that the research involved no deviation from the care that patients would otherwise receive during the phased rollout, all patient data used for study analysis was routinely-collected, and the analyzed data were stripped of patient identifiers, the AMREF Health Africa Ethical and Scientific Review Committee determined that additional consent particular to this study was not needed beyond Penda's existing consent form.

The study also included a survey of clinicians to understand their satisfaction with Penda's EMR and AI Consult. As these surveys are not ordinarily done, we sought explicit written consent from clinicians. These surveys were fully anonymous.

\paragraph{Funding.} Funding for this study was provided by OpenAI. OpenAI was involved in the study analysis and reporting.

\paragraph{Reporting.} The reporting of this quality improvement study was guided by the SQUIRE 2.0 statement \citep{Ogrinc2016squire}.

\subsection{Study population and data} 
For this study, we included data from all 15 Penda clinics in Nairobi County, Kenya.\footnote{The specific clinics included span Penda's three service regions: Eastlands (Tassia, Umoja 1, Umoja 2, Embakasi, and Pipeline); Southwest (Kangemi, Kawangware, Kimathi Street, and Lang'ata), and the Thika Road Corridor (Mathare North, Kasarani, Sunton, Lucky Summer, Zimmerman, and Kahawa West).  This is all but one of Penda's facilities; the remaining one, Githurai 45, is located in Kiambu County and was excluded as we sought approval for this study specifically in Nairobi County.} These centers provide both primary and urgent care services, and also have laboratory and pharmacy services onsite. In most cases, these centers are located within Nairobi’s urban low- and middle-income demographic communities.

We included in-person visits at Penda where clinicians actively document in the Penda EMR. This means that we excluded visit categories where clinicians generally do not actively document in the EMR, e.g., over-the-counter medication requests, laboratory self-requests, as well as patients in Penda's blood pressure chronic care management program ``BP Sawa'' and routine well-baby care. Finally, we excluded telemedicine visits because they are not in person, and dental visits because of their more narrow focus.

\begin{figure}[htbp]
  \centering
  \includegraphics[width=\linewidth]{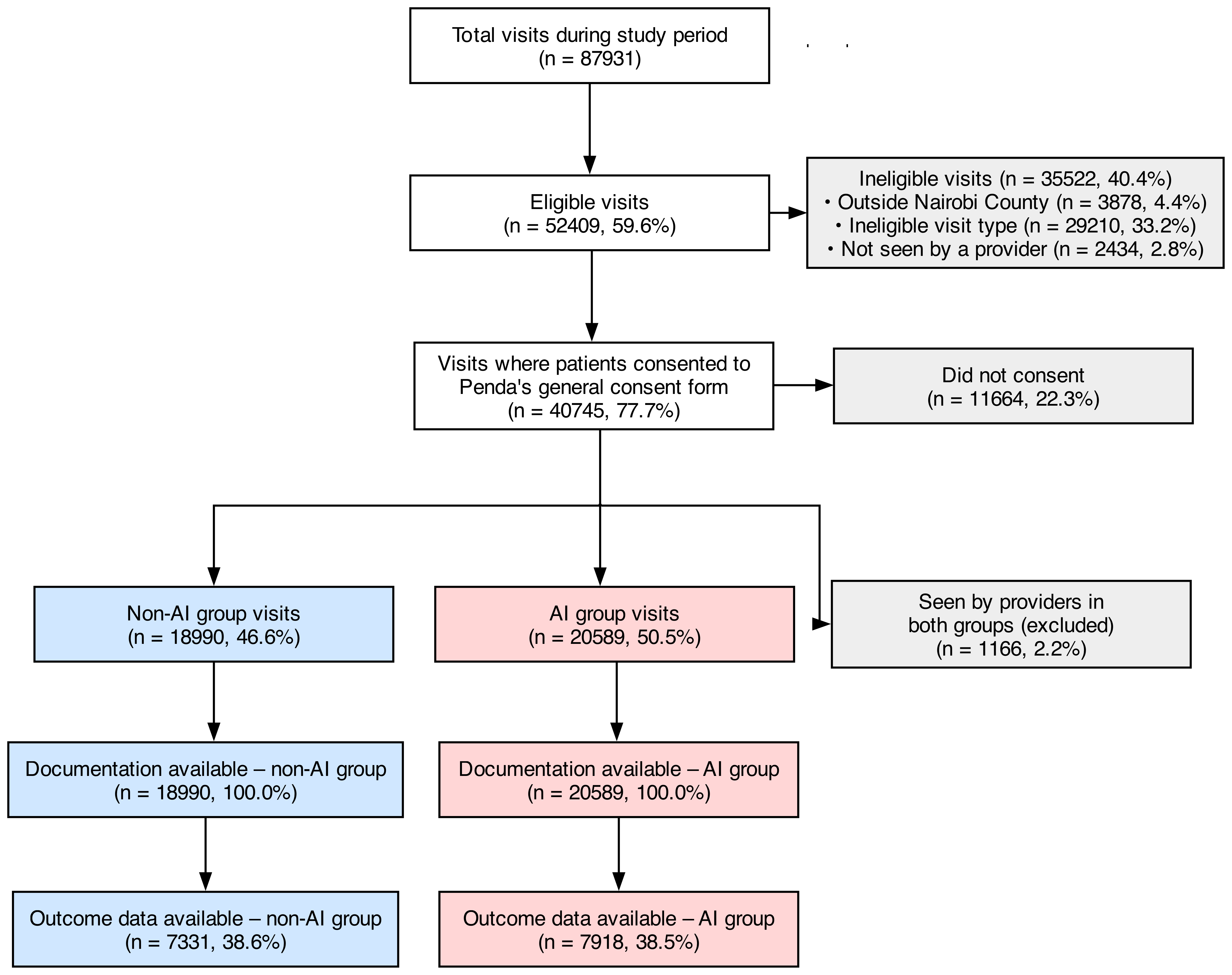}
  \caption{Flow diagram showing visit eligibility, consent, group assignment, and data availability.}
  \label{fig:visit_flow}
\end{figure}

From \rolloutStartDate to \rolloutEndDate, a total of \nTotalVisits patient visits were recorded at all of Penda's clinics (\cref{fig:visit_flow}). Of these, \nEligibleVisits visits (\pctEligible) met the study eligibility criteria (\cref{fig:visit_flow}). The remaining \nIneligibleVisits visits (\pctIneligible) were excluded either because (1) they occurred at Penda's single clinic outside Nairobi County (\nOutsideNairobiCounty visits, \pctOutsideNairobiCounty); (2) they had an ineligible visit category (\nIneligibleVisitType visits, \pctIneligibleVisitType); or (3) the patient was not seen by a clinical officer in the course of their visit (\nNotSeenByProvider visits, \pctNotSeenByProvider). Among the \nEligibleVisits{} eligible visits, \nConsentedVisits (\pctConsented) were visits where patients agreed to Penda's general consent form and so were included in the study. 

Across the study, \nActiveAI clinicians in the AI group had access to AI Consult, while the \nSilentAI clinical officers in the non-AI group did not. Each clinician contributed a median of \visitsPerProviderMedActive visits in the AI arm and  \visitsPerProviderMedSilent visits in the non-AI arm.

Patient visits were split into the ``AI'' group if all clinicians who saw them had access to AI Consult and into the ``non-AI'' group if no clinician who saw them had access to AI Consult. The AI group included \nActiveVisits visits (\pctActiveVisits of visits with general consent), with clinician documentation available for \nDocsActiveVisits (\pctDocsActiveVisits) and structured outcome data for  \nOutcomesActiveVisits (\pctOutcomesActiveVisits). The non-AI group included \nSilentVisits visits (\pctSilentVisits of visits with general consent), with patient documentation available for \nDocsSilentVisits (\pctDocsSilentVisits) and structured outcome data for \nOutcomesSilentVisits (\pctOutcomesSilentVisits). A small portion of visits (\nSeenByProviderInBothGroups{},  \pctSeenByProviderInBothGroups{}) were attended by clinicians in both groups, primarily due to handover at shift change, and were excluded from analysis.

Patient age, insurance vs cash-pay mix, and 8-day follow-up call response rates were generally well-balanced between the non-AI and AI arms (\cref{tab:demographics}).

Visits were distributed across Nairobi’s three service regions. Comparatively more patient visits in the AI group occurred in clinics in the Thika Road Corridor (\pctThikaActive AI vs \pctThikaSilent non-AI), while comparatively fewer were in Eastlands clinics (\pctEastlandsActive vs \pctEastlandsSilent) and Southwest clinics (\pctSouthwestActive vs \pctSouthwestSilent)  (\cref{tab:demographics}).

\begin{table}[htbp]
  \centering
  {\small
\begin{tabular}{lll}
\toprule
\textbf{Variable} & \textbf{Non-AI} & \textbf{AI} \\
\midrule
n & 18,990 & 20,589 \\
 &  &  \\
Induction period (before March 1 2025) & 7,773 (40.9\%) & 8,201 (39.8\%) \\
Main study period (March 1 2025 or later) & 11,217 (59.1\%) & 12,388 (60.2\%) \\
 &  &  \\
Visit location: Eastlands clinics & 8,277 (43.6\%) & 7,911 (38.4\%) \\
Visit location: Southwest clinics & 4,223 (22.2\%) & 3,867 (18.8\%) \\
Visit location: Thika Road Corridor clinics & 6,490 (34.2\%) & 8,811 (42.8\%) \\
 &  &  \\
Age (years), median [q25, q75] & 20.8 [4.0, 32.2] & 20.7 [3.9, 31.6] \\
 &  &  \\
Female & 10,505 (55.3\%) & 11,282 (54.8\%) \\
Male & 8,485 (44.7\%) & 9,307 (45.2\%) \\
 &  &  \\
Insurance visit & 10,501 (55.3\%) & 11,713 (56.9\%) \\
Cash visit & 8,489 (44.7\%) & 8,876 (43.1\%) \\
 &  &  \\
Did respond to 8-day follow-up call & 7,333 (38.6\%) & 7,919 (38.5\%) \\
Did not respond to 8-day follow-up call & 11,657 (61.4\%) & 12,670 (61.5\%) \\
\bottomrule
\end{tabular}
}
  \caption{Demographics of visits included in this study.}
  \label{tab:demographics}
\end{table}

\subsection{Data analysis}

Statistical analysis for this study was done using Python 3.12, using \texttt{scipy} for statistical testing, \texttt{statsmodels} for statistical modeling, and a threshold of $p = 0.05$ in determining statistical significance. We conducted an intent-to-treat analysis, comparing patient visits seen only by clinicians in the AI group with those seen only by clinicians in the non-AI group.

\subsection{Effects on quality of care}

We examined the effects of AI Consult on quality of care by having independent physicians rate visit documentation stripped of patient identifiers. 

We selected a random sample of \nUniqueVisitsHumanRated visits recorded during the study. We then presented these to a panel of \nPhysicianReviewers physicians for review of documentation and clinical decision-making quality, including diagnosis and treatment errors.\footnote{One consequence of this approach is that we have physician-assessed outcome data for a random sample of visits. We present an analysis of all complete cases for this outcome, as is valid for data missing at random \citep{Ross2020}.}

\paragraph{Physician rater panel.} The \nPhysicianReviewers physician raters included staff physicians and senior residents from around the world, including \nPhysicianReviewersKenya from Kenya. The vast majority of these were family physicians, emergency physicians, internists, pediatricians, or general practitioners. The remaining physicians had practice experience in other relevant specialties: obstetrics, preventative medicine, physical medicine and rehabilitation, general surgery, and public health. These physicians were selected by OpenAI using a multi-step process to ensure their quality and performance. For more details on selection and the physician panel, see \cite{arora2025healthbench}. 

\paragraph{Blinding.} Raters were blinded to whether the patient visit was in the AI or non-AI group. They also had no information about the quality improvement study, AI Consult, or study hypotheses. Raters were told these visits occurred in a primary/urgent-care setting in Kenya and the resources available in the setting, so they had enough information to rate visits.

\paragraph{The rating task.} Physician raters were presented with a form containing patient documentation stripped of patient identifiers, which included the patient history (age, gender, vital signs, chief complaint, and clinical note), any diagnostic investigations done with results, the clinician-assigned diagnosis, and management plan including medications, referrals, and any diagnostic investigations that could not be done in that clinic. 

\begin{table}[htbp]
\centering
\small
\renewcommand{\arraystretch}{1.3} 
\begin{tabularx}{\textwidth}{
  >{\centering\arraybackslash}m{3cm}  
  X                                  
}
\toprule
\textbf{Clinical Category} & \textbf{Description} \\
\midrule
\textbf{History \& Examination} & The patient’s presenting chief complaint, vital signs, past medical history, social and family history, and physical exam findings. A thorough history and physical examination is essential for high-quality clinical reasoning. \\\midrule
\textbf{Investigations} & Diagnostic tests ordered or performed, including laboratory investigations, imaging, and point-of-care tests. These investigations are critical to confirming or ruling out clinical hypotheses. \\\midrule
\textbf{Diagnosis} & Most likely clinical condition(s) given history and investigations. A high-quality diagnosis captures both primary and any clinically-relevant comorbid conditions with appropriate specificity. \\\midrule
\textbf{Treatment} & Clinical management plan, including medications prescribed, procedures performed, referrals made, patient education, and follow-up instructions. Treatment should be individualized and guideline-concordant. \\
\bottomrule
\end{tabularx}
\caption{Descriptions of the four core clinical categories used to evaluate visit quality.}
\label{tab:clinical_categories}
\end{table}

Physicians were asked to give a five-point Likert rating for (1) the depth and appropriateness of the history and physical exam; (2) whether appropriate investigations were done and inappropriate investigations were not; (3) whether the diagnosis assigned was likely correct, and whether relevant additional diagnoses were captured if present (e.g., anemia being captured if present on blood testing, even if the chief complaint was a respiratory illness); and (4) whether the management plan was correct and high quality. A score of 1 or 2 on any of these Likert scales was intended to correspond to a clinically meaningful error.\footnote{For cases after April 9, the structured chief complaint field was missing, so we omitted data from April 10 onward in analysis of History Likert and multiple-choice question data; sufficient information was still available about the history from the clinical notes and other history fields to enable assessment of the patient note, and so we still included the investigations, diagnosis, and treatment data for these cases.} For full Likert scale definitions for each category, see \cref{tab:Likert_definitions_history_examination,tab:Likert_definitions_investigations,tab:Likert_definitions_diagnosis,tab:Likert_definitions_treatment}. For the reference examples provided to physicians of each Likert value for each category, including examples of Likert 1 and 2 errors, see \cref{app:rater-form}.

For each of these categories, physicians were also asked to enumerate the failure modes present (i.e., the specific errors made in each category above), if any. For example, for the ``diagnosis'' category, physician raters were asked to choose as many options as applicable from the below. For all options across categories, see \cref{tab:MCQ_deficiencies}.
\begin{itemize}
    \item Primary diagnosis is likely incorrect
    \item Primary diagnosis is missing
    \item Primary diagnosis is too specific to be supported based on current documentation or investigations (e.g., using “allergic rhinitis” as the diagnosis where it’s clear that rhinitis is present but documentation does not support a specific etiology)
    \item Primary diagnosis is too broad when a more specific diagnosis is supported
    \item Additional diagnosis is likely incorrect
    \item Clinically relevant additional diagnosis is missing (e.g. anemia)
    \item None of the above
\end{itemize}

Finally, we asked physicians to rate the acuity of the clinical scenario as ``low'', ``medium'', or ``high'', to enable analysis stratified by severity. For the full form shown to physicians, including examples and full question text, please see \cref{app:rater-form}.

\paragraph{Rater agreement.} The physician rater panel was trained to reduce subjectivity and improve reliability. For each Likert scale, we established golden examples for each Likert value based on the consensus of three physician investigators, and shared these golden examples with the panel as reference points (see \cref{app:rater-form}). In addition to the multi-step onboarding and quality-filtering described in \cite{arora2025healthbench}, we also provided detailed training on how to review clinical documentation to evaluators. This included upfront training over video call, frequent ``office hours'', and ongoing clarifications when ambiguities arose.

We evaluate rater agreement by having a portion of unique tasks be completed by two independent raters. Of the \nUniqueVisitsHumanRated visits rated, \nVisitsDoubleRated (about 25\%) were rated by two physicians; the remaining \nVisitsSingleRated of these were rated by a single physician. We calculated inter-rater agreement for the Likert scales, defining ``agreement'' as cases where the Likert values that two physician raters chose were within one point of one another. We also calculate agreement on error (i.e., whether raters agreed on an instance being Likert 1/2 vs Likert 3/4/5) between two raters compared to the agreement that would be expected by chance using Fleiss' $\kappa$.

\paragraph{Statistical analysis.} Our primary outcome measure is the relative risk reduction (RRR) in clinically meaningful errors (i.e., Likert 1/2) for each category (history, investigations, diagnosis, and treatment) between the AI and non-AI groups. 

We report the proportion of clinically meaningful errors in each group and the corresponding 95\% Wilson confidence intervals, comparing this between the non-AI and AI groups using Fisher's exact test. We also compute the relative risk reduction for errors in the AI group compared to the non-AI group, with its 95\% confidence interval computed using the Katz method. For cases rated by two physicians, we assign each rating weight $0.5$ so that each visit ultimately has equal weight in the final analysis. We use the Benjamini-Hochberg procedure to control the false discovery rate between the four clinical domains that we measure. 

We also report covariate and clustering-adjusted measures of effect size. To do so, we fit a generalized linear model, using generalized estimating equations (GEE) to fit the model to account for within-clinician effects while yielding population-average effects. We fit a log-binomial GEE model, using a log link to estimate risk ratios and calculate the relative risk reduction as 1 minus the risk ratio. We fit this model with grouping at the clinician level, specifying an exchangeable covariance structure to account for clinician effects. The fixed effects we include in this model are AI vs non-AI (reference: non-AI), age (in years, continuous), gender (reference: male), and insurance vs cash visit (reference: cash). We also include clinic as a fixed effect in this model, both in order to estimate effects for specific clinics, and recognizing that we observe all of Penda's Nairobi County clinics in this study.\footnote{At Penda, clinicians operate across multiple clinics, so we cannot consider clinicians to be a level of grouping nested within a clinic.}  We use sum-to-zero coding for clinic, with Zimmerman as the necessary omitted clinic.

Finally, for robustness and to evaluate sensitivity to modeling assumptions, we also fit and report results from a modified Poisson regression model with cluster-robust standard errors. The modified Poisson approach has become a common method to estimate relative risks in binary outcome studies \citep{Zou2004,ZouDonner2013}. We specify this analysis with the fixed effects specified above and include clinicians as clustering variables.

\paragraph{Additional analyses.} Our primary analysis was of the main study period, after the induction phase (\restStudyStartDate to \rolloutEndDate).

To examine whether AI Consult signals correlate with clinical quality, we also study the physician-rated quality of cases that were ``left in red'' (i.e., where the final AI Consult response was red for at least one of the five AI Consult categories, or would have been red for cases in the non-AI group) vs cases that were not left in red. This measures how well the tool's responses match the clinical judgment of our physician rater panel.

We also do sensitivity analyses to examine the effect of AI Consult in visits where there was at least one red AI Consult response, and in visits during the induction period only.

\paragraph{LLM rater analysis.} We also conduct a version of this analysis where we have LLMs rate clinical documentation, enabling rating of all patient documentation and the evaluation of LLM ability to conduct such ratings. We conduct independent ratings using two different OpenAI models: o3, which is currently OpenAI's most capable model in health, and GPT-4.1, which the HealthBench paper established as a strong grader for health-related tasks \citep{arora2025healthbench}. 

For each rater model, we conduct analysis as described above, including computing risk ratios and fitting modified Poisson regression and GEE models. We also study the agreement of each model-based rater with physicians who rated the same visit, computing the rate of model agreement with physician ratings within one Likert point, and again evaluating whether models agreed with physician ratings on whether a given task contained errors (Likert 1/2 vs Likert 3/4/5) vs chance agreement with Fleiss' $\kappa$.

\subsection{Use and usability analysis}

\paragraph{Clinician survey.} At the end of the rollout period, we invited Penda clinicians in both the AI and non-AI groups to participate in an anonymous, consented survey.

We asked both groups about their experience with Penda's EMR (including AI Consult) and whether it changes the quality of care that they deliver on a five-point scale. We compare this between groups with a Mann-Whitney \textit{U}-test. We also asked the AI group about AI Consult: whether it changes the quality of the care they deliver (five-point scale), whether they'd recommend AI Consult to others (to compute net promoter score), and their satisfaction (five-point scale). Finally, we solicited qualitative feedback from both groups. The full text of the clinician survey is available in \cref{app:clinician-survey}.

\paragraph{Usage data.} We also examine differences in the median visit duration and median clinical documentation length between groups, as well as the fraction of AI Consult responses given ``thumbs up'' ratings vs ``thumbs down'' ratings by clinicians.

\subsection{Patient outcomes}
\label{sec:pro}

\paragraph{Patient-reported outcomes.}
As part of standard care, Penda Health makes calls to all eligible and consenting patients by telephone 8 days after an index visit to collect patient-reported outcomes. Patients who respond are asked whether they are feeling better on a five-point Likert scale (5 = ``much better'', 4 = ``a bit better'', 3 = ``about the same'', 2 = ``a bit worse'', 1 = ``much worse''), with patients who report 3 or less defined as ``not feeling better''. Patients are also asked whether they visited another pharmacy or went to another clinic themselves (without Penda's referral; see \cref{app:call-center-script} for the full script).

Penda also identifies patients with more clinically severe presentations as possible candidates for one-day follow-up calls, as described in \cref{app:one-day-eligibility}. Clinicians in Penda's call centers ultimately decide which patients to follow up with, and call to ask whether patients' conditions have worsened since their visit.

We compare these outcomes between the AI and non-AI groups with Fisher's exact test. Given the high rates of missingness in patient outcome data (about 60\%), we report a complete-case analysis that we treat as exploratory rather than confirmatory, following recommendations from \cite{Jakobsen2017}. 

\paragraph{Serious escalations.} 
We also monitored the frequency of serious escalations reported to Penda as part of routine patient care through Penda's patient safety reporting (PSR) system. This system, which has been in effect since 2020, allows any staff member at Penda to raise a quality or patient safety concern. PSRs are often raised by Penda's customer service team when a patient experiences harm or potential harm that could be related to their care at Penda. This includes harm that is deemed unavoidable (e.g., a new medication allergy) and serious errors that did not result in harm (``near miss'' events). 

Penda's clinical quality team reviews all PSRs to identify any safety or quality gaps that need to be addressed. We examined the frequency and severity of such reports in the non-AI and AI groups. For reports in the non-AI group, we examine whether AI Consult alerts that would have been raised during the patient's visit could have prevented harm or a near miss from occurring, if the clinician was able to see them and acted on them. For reports in the AI group, we examine whether the AI alerts raised during the patient's visit could have (i) been responsible for the harm experienced; (ii) failed to prevent harm; or (iii) have prevented harm but failed to because the clinician did not see them or chose not to act on them. 

\subsection{Data management and privacy}
We handled, stored, and processed all participant data following Kenya’s Data Protection Act. Patient data was fully stripped of patient identifiers (e.g., patient names, date of birth, phone numbers, national ID numbers, medical record numbers, specific geography) as well as clinician identifiers. This removal was accomplished in two steps. First, Penda clinician training includes specific instructions not to use patient identifiers in their free-text clinical notes. Secondly, the Penda data team reviewed the patient notes used in this study to ensure privacy. They identified a very low rate of cases where possible patient identifiers were used (approximately 5 per 10,000 notes). In these cases, any identifiers present were redacted. The research team had access only to this research dataset stripped of identifiers. 

Participants were able to request that we remove their data from this study until 15 days after the AI Consult rollout ended. After that period, the data from participants was fully processed and anonymized, meaning it was not possible to remove it. No participant requested that we remove their data at any time.

Study data will be retained for a 5-year period after publication of the results to enable research reproducibility. After this period, all study-related data will be securely destroyed to ensure privacy and compliance with data protection standards. 

\section{Results}
\label{sec:results}

We present results on quality of care (\cref{sec:results-quality-of-care}), use and usability (\cref{sec:results-use-and-usability}), and patient outcomes (\cref{sec:results-patient-outcomes}).

\subsection{Effects on quality of care}
\label{sec:results-quality-of-care}

\paragraph{AI Consult reduced clinical error rates.}

\begin{figure}
    \centering
    \includegraphics[width=0.9\linewidth]{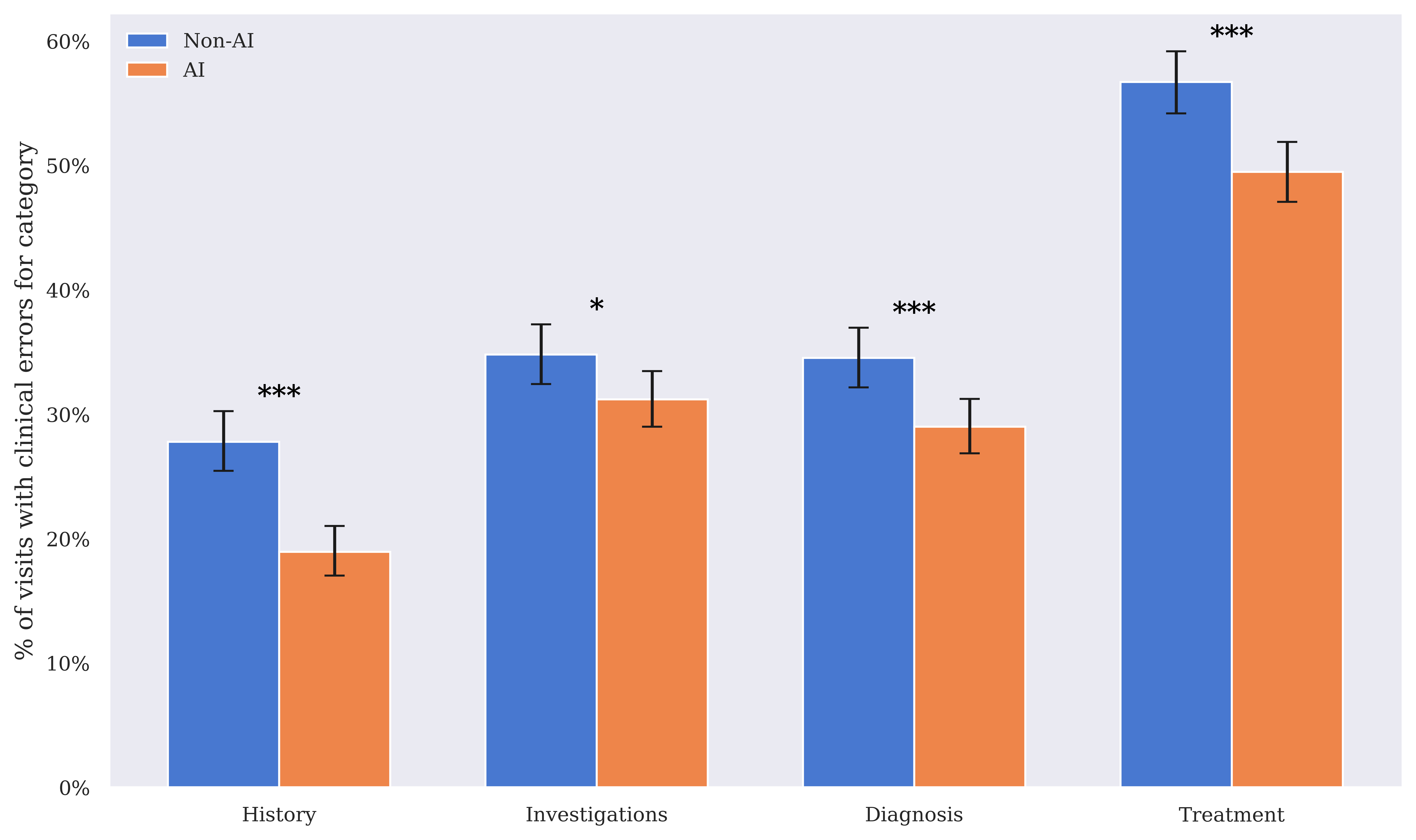}
    \caption{Clinical error rates for history-taking, investigations, diagnosis, and treatment, comparing the AI group to the non-AI group. Error bars show 95\% Wilson confidence intervals. * indicates $p < 0.05$, ** $p < 0.01$, *** $p < 0.001$.}
    \label{fig:likert_results_main_period}
\end{figure}

Error rates across each of the four clinical categories were significantly reduced in the AI group compared to the non-AI group. The relative risk reduction for AI compared to non-AI was \mainPeriodLikertErrorRatesRRRHistory (95\% CI \mainPeriodLikertErrorRatesRRRHistoryLowCI-\mainPeriodLikertErrorRatesRRRHistoryHighCI, $p=\mainPeriodLikertErrorRatesHistoryPVal$) for history-taking, \mainPeriodLikertErrorRatesRRRInvestigations (\mainPeriodLikertErrorRatesRRRInvestigationsLowCI-\mainPeriodLikertErrorRatesRRRInvestigationsHighCI, $p=\mainPeriodLikertErrorRatesInvestigationsPVal$) for investigations, \mainPeriodLikertErrorRatesRRRDiagnosis (\mainPeriodLikertErrorRatesRRRDiagnosisLowCI-\mainPeriodLikertErrorRatesRRRDiagnosisHighCI, $p=\mainPeriodLikertErrorRatesDiagnosisPVal$) for diagnostic errors, and \mainPeriodLikertErrorRatesRRRTreatment (\mainPeriodLikertErrorRatesRRRTreatmentLowCI-\mainPeriodLikertErrorRatesRRRTreatmentHighCI, $p=\mainPeriodLikertErrorRatesDiagnosisPVal$) for treatment errors (\cref{fig:likert_results_main_period}, \cref{tab:error_rate_reduction_main}). All four $p$-values remain significant after applying the Benjamini–Hochberg procedure with FDR 5\% across these tests.

Notably, the number needed to treat (NNT) for AI Consult was low, particularly for a tool with such broad effects: \mainPeriodLikertErrorRatesNNTDiagnosis for diagnostic errors and \mainPeriodLikertErrorRatesNNTTreatment for treatment errors. If Penda adopted AI Consult across its 400,000 annual visits, this would correspond to about \totalErrorReductionDiagnosis fewer diagnostic errors annually and \totalErrorReductionTreatment fewer treatment errors annually (\cref{tab:error_rate_reduction_main}).

\begin{table}
    \centering
    \begin{tabular}{llll}
\toprule
 & RRR: all visits & NNT & Yearly errors averted at Penda \\
\midrule
History & 31.8\% (21.9\%-40.5\%) & 11.3 & 35383 \\
Investigations & 10.3\% (1.0\%-18.8\%) & 27.8 & 14388 \\
Diagnosis & 16.0\% (6.9\%-24.2\%) & 18.1 & 22102 \\
Treatment & 12.7\% (6.8\%-18.3\%) & 13.9 & 28880 \\
\bottomrule
\end{tabular}

    \caption{Relative risk reduction in clinical errors. Includes overall effect size and number needed to treat for the main study period. Also includes the absolute number of errors we would expect to be averted if this tool were widely deployed in the 400,000 annual patient visits at Penda.}
    \label{tab:error_rate_reduction_main}
\end{table}

We also examined the effect size during the induction period. The error rate reduction for history, diagnosis, and treatment is much higher in the main study period compared to the induction period (e.g., for treatment, \mainPeriodLikertErrorRatesRRRTreatment, 95\% CI \mainPeriodLikertErrorRatesRRRTreatmentLowCI-\mainPeriodLikertErrorRatesRRRTreatmentHighCI during the main study period compared to \inductionRRRTreatment, 95\% CI \inductionRRRTreatmentLowCI-\inductionRRRTreatmentHighCI during the induction period), providing evidence for the value of active deployment (\cref{tab:induction_vs_risky_vs_main}). 

We also examined the effect size in visits where there was at least one red AI Consult response for each category. The effect sizes for diagnosis and treatment were considerably higher in such visits (diagnosis: RRR \mainPeriodRedOnlyRRRDiagnosis for visits with at least one red vs \mainPeriodLikertErrorRatesRRRDiagnosis for all cases; treatment: \mainPeriodRedOnlyRRRTreatment  vs \mainPeriodLikertErrorRatesRRRTreatment; \cref{tab:induction_vs_risky_vs_main}). There were no obvious trends in effect size by physician-rated acuity (\cref{tab:acuity_rrr}).  
\begin{table}
    \centering
    \begin{tabular}{llll}
\toprule
 & Main period, all visits & Induction period & Main period, only visits with reds \\
\midrule
History & 31.8\% (21.9\%-40.5\%) & 16.7\% (4.5\%-27.3\%) & 30.8\% (12.8\%-45.1\%) \\
Investigations & 10.3\% (1.0\%-18.8\%) & 13.8\% (2.7\%-23.6\%) & 17.9\% (-72.6\%-60.9\%) \\
Diagnosis & 16.0\% (6.9\%-24.2\%) & 6.4\% (-5.6\%-17.1\%) & 31.5\% (14.0\%-45.5\%) \\
Treatment & 12.7\% (6.8\%-18.3\%) & 4.3\% (-3.0\%-11.1\%) & 18.0\% (9.4\%-25.9\%) \\
\bottomrule
\end{tabular}

    \caption{Relative risk reduction in clinical errors across each category for all visits during the main study period, all visits during the induction period, and only visits with reds for the relevant category during the main study period.}
    \label{tab:induction_vs_risky_vs_main}
\end{table}

We also fit statistical models to account for clinician clustering, clinic effects, and patient covariates, which yielded similar results to the unadjusted analysis. For diagnosis and treatment, GEE model effect sizes were of the same magnitude and retained statistical significance: for diagnosis, \mainStudyGEEdiagnosisRRRPoint (GEE fit) vs \mainPeriodLikertErrorRatesRRRDiagnosis (unadjusted effect); and for treatment, \mainStudyGEEtreatmentRRRPoint vs \mainPeriodLikertErrorRatesRRRTreatment. For history, the effect size was somewhat smaller (\mainStudyGEEhistoryRRRPoint vs \mainPeriodLikertErrorRatesRRRHistory), but retained statistical significance. For investigations, the effect size was similar (\mainStudyGEEinvestigationsRRRPoint vs \mainPeriodLikertErrorRatesRRRInvestigations), but had $0.05 < p < 0.1$. Examining other model coefficients, there was notable variation in error rates across clinics for all categories. For the treatment category, we also observed a higher risk of errors in younger patients. Full GEE results are available in \cref{tab:history_gee,tab:investigations_gee,tab:diagnosis_gee,tab:treatment_gee}. Results from modified Poisson models (which are largely similar) are in \cref{tab:history_poisson,tab:investigations_poisson,tab:diagnosis_poisson,tab:treatment_poisson}. 

Lastly, we also repeated this analysis using a large language model rather than physician ratings. We report findings from this analysis in \cref{tab:physicians_ai_rrr_table}.

\paragraph{The rate of common clinical failure modes was lower in the AI group.}

\begin{figure}
    \centering
    \includegraphics[width=\linewidth]{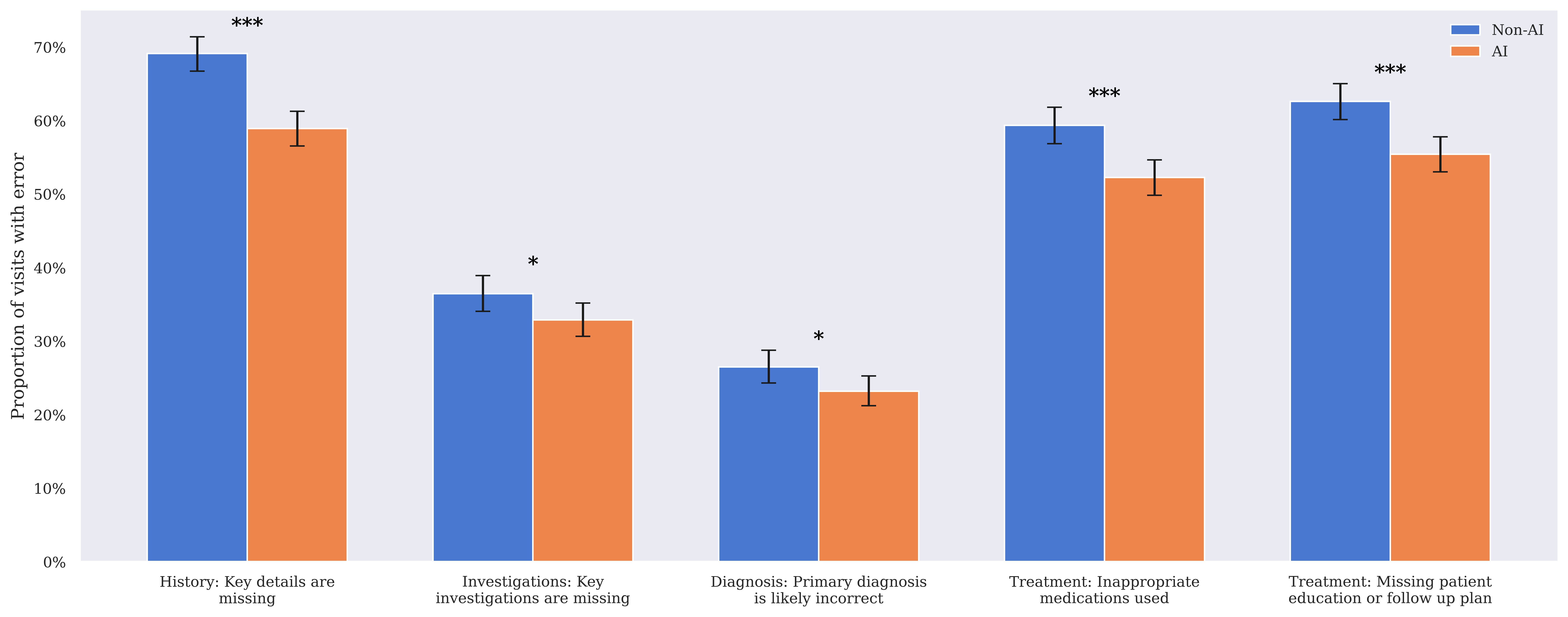}
    \caption{Rates of selected clinical failure modes in the AI group compared to the non-AI group. Error bars show 95\% Wilson confidence intervals. * indicates $p < 0.05$, ** $p < 0.01$, *** $p < 0.001$. For a full table of failure modes and their rates in both groups, see \cref{tab:mcq}.}
    \label{fig:mcq}
\end{figure}

We also had raters identify the specific clinical failure modes present in the visit documentation. We find that several error categories are less common in the AI group, including the rate of key details missed in the history, key investigations missed, or incorrect main diagnoses (\cref{fig:mcq}). We also find that AI group visits were less likely to have the wrong medications prescribed or important patient education omitted. No failure modes are more common in the AI group compared to the non-AI group. For a full table of failure modes and their frequency between groups, see \cref{tab:mcq}.

\paragraph{Fewer visits were left with red AI Consult responses in the AI group.}
To understand how AI Consult achieved this effect, we examined how many visits had any calls left in red–that is, where the final AI Consult call in the visit was red, for any of history, investigations, diagnosis, or treatment. 

\begin{figure}[ht]
    \centering
    \begin{subfigure}[t]{0.48\linewidth}
        \centering
        \includegraphics[width=\linewidth]{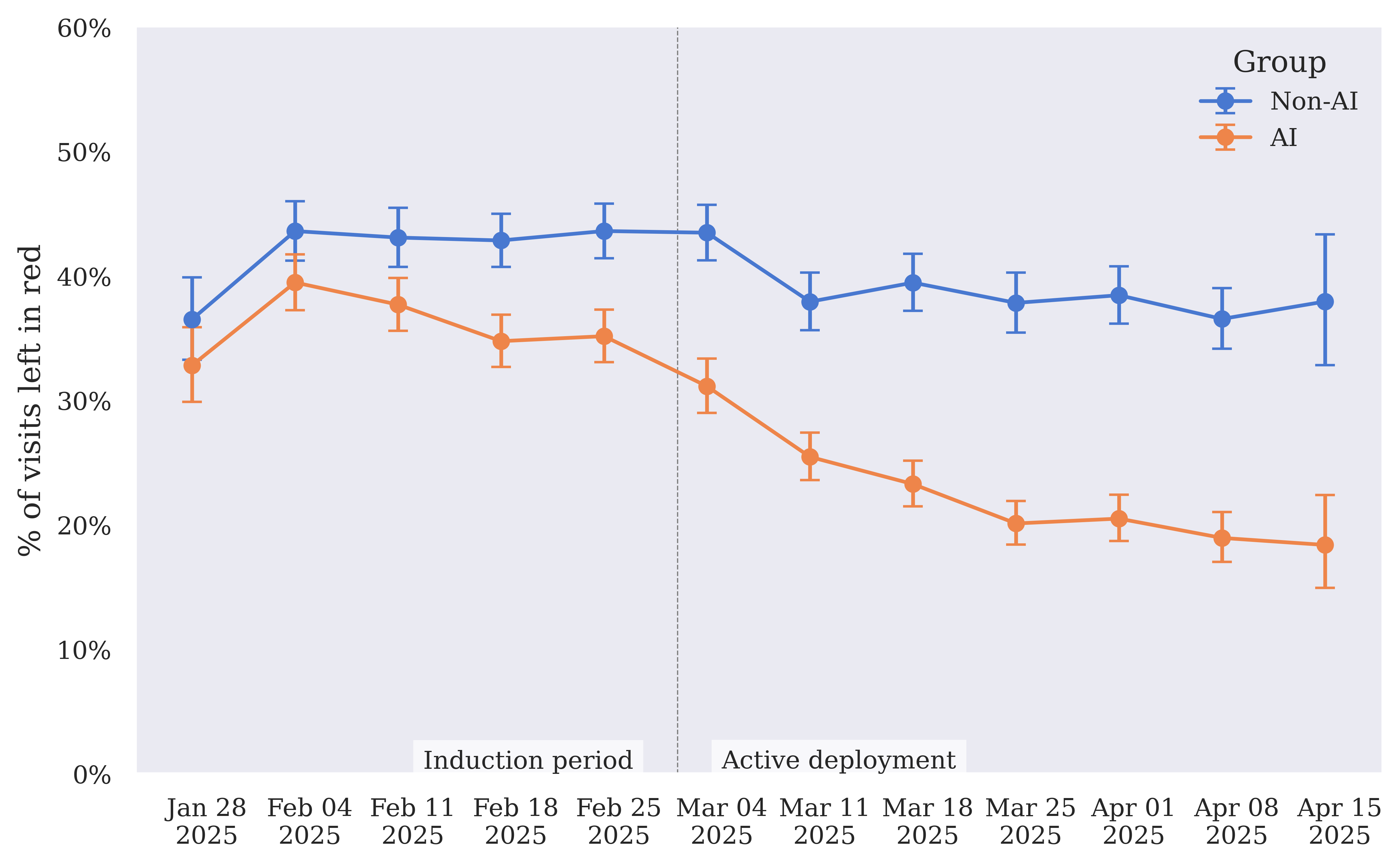}
        \caption{Left in red rate: rate of visits where the final call for any of the AI Consult categories is red, for AI and non-AI groups over time.}
        \label{fig:final_red_rate}
    \end{subfigure}
    \hfill
    \begin{subfigure}[t]{0.48\linewidth}
        \centering
        \includegraphics[width=\linewidth]{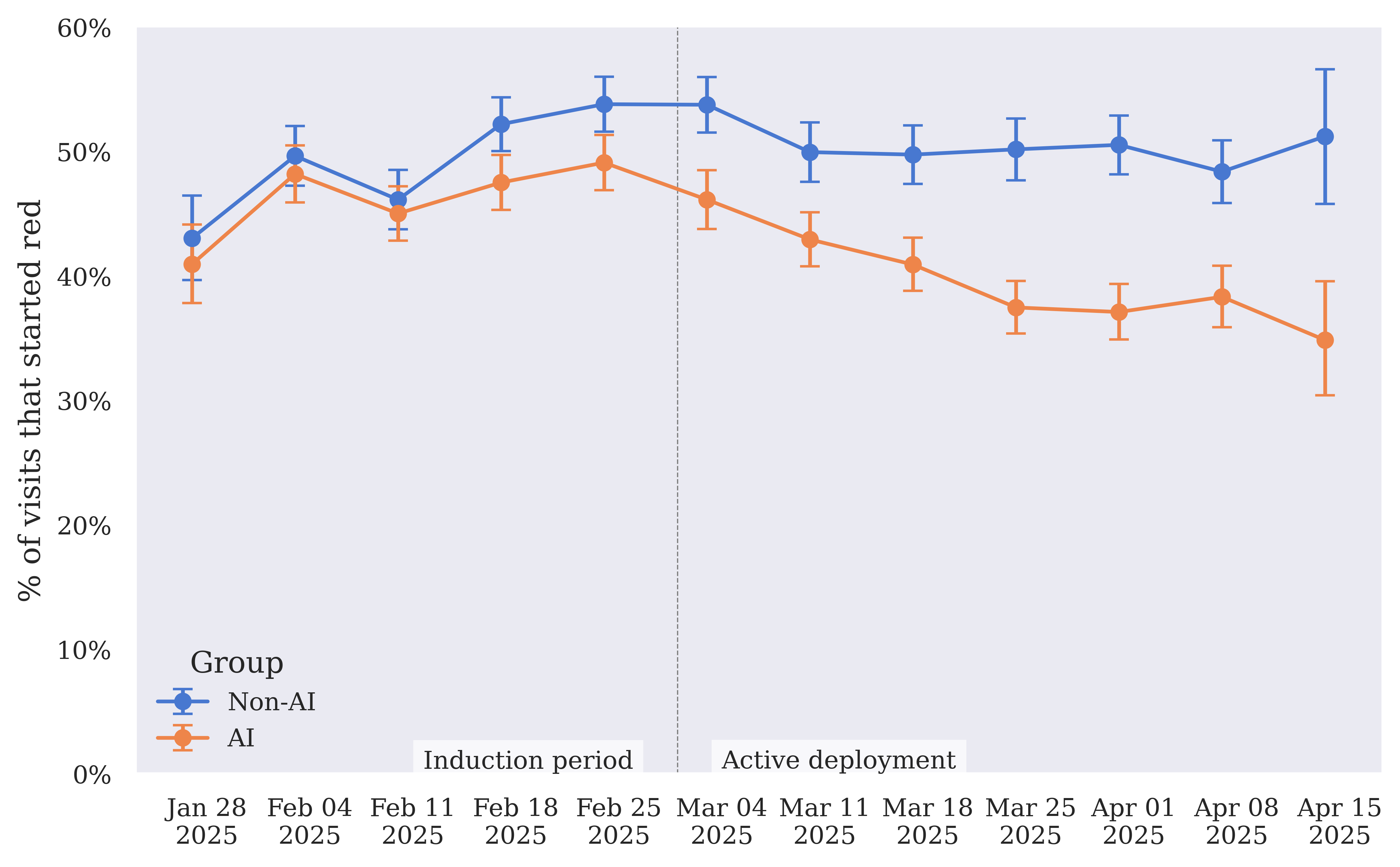}
        \caption{Started red rate: rate of visits where the first call for any of the AI Consult categories is red, for AI and non-AI groups over time.}
        \label{fig:first_red_rate}
    \end{subfigure}
    \caption{Rates of visits left in red and started in red over time for AI and non-AI groups.}
    \label{fig:reds_over_time_panels}
\end{figure}

At the start of the induction period, the left in red rate was similar between groups at 35-40\%, suggesting that clinicians in the AI group were only sometimes seeing or acting on the red alerts displayed to them. Once Penda iterated on AI Consult to improve reliability and started active deployment, the left in red rate in the AI group dropped to 20\% while the non-AI group rate stayed at 40\% (\cref{fig:final_red_rate}). This was also the case when looking at cases where the treatment specifically was left in red (\cref{fig:final_red_rate_treatment_only}). This difference helps explain AI Consult's effects and also emphasizes the importance of user testing and active deployment.

\paragraph{Clinicians in the AI group learned to avoid common mistakes over time.}
We also examine the proportion of visits where AI Consult started red–that is, where the first AI call for any category was red. In the AI group, this rate drops from 45\% at the start of the study to 35\% at the end of the study, while staying steady at 45-50\% in the non-AI group during the study (\cref{fig:first_red_rate}). This suggests that AI Consult is training clinicians to avoid common mistakes even prior to AI Consult alerts. We see this training effect even when only looking at the history-related AI Consult categories, indicating that this effect cannot just be explained by AI Consult history popups leading to better initial diagnoses and treatments (\cref{fig:first_red_rate_history_only}). We also see this training effect when looking at cases where the treatment specifically started red (\cref{fig:first_red_rate_treatment_only}).

To further interrogate this training effect, we examine the distribution of the left in red rate across clinicians over time. We find that the active deployment period led to a considerable drop in the left in red rate for the 10th percentile clinician from 20\% at the start of the study to 0\% at the end  (\cref{fig:final_red_quantiles}). The 25th, 50th, and 75th percentiles also improved considerably. In contrast, the 90th percentile (clinicians with the highest left in red rate) regressed towards the end of the active deployment period, suggesting that these clinicians may have been generally disengaged.

\textbf{AI Consult severity corresponds to clinician-graded severity.}
Our analysis of visits left in red raises the question of whether AI Consult responses (and left in red rates) correlate with clinical quality. To answer this question, we examined the clinical error rate in cases where AI Consult was left in red, yellow, or green. For history, diagnosis, and treatment, error rates were substantially higher in visits that were left in red vs yellow, and for visits that were left in yellow vs green (\cref{tab:ryg_corr}). For example, for the diagnosis category, the clinical error rate was \redVYellowFirstGroupDiagnosisRate (95\% CI \redVYellowFirstGroupDiagnosisLowerCI-\redVYellowFirstGroupDiagnosisUpperCI) for reds, \redVYellowSecondGroupDiagnosisRate (95\% CI \redVYellowSecondGroupDiagnosisLowerCI-\redVYellowSecondGroupDiagnosisUpperCI) for yellows, and \yellowVGreenSecondGroupDiagnosisRate (95\% CI \yellowVGreenSecondGroupDiagnosisLowerCI-\yellowVGreenSecondGroupDiagnosisUpperCI) for greens.

\begin{table}
    \small
    \centering
    \begin{tabular}{llllll}
\toprule
 & Left in red & Left in yellow & Left in green & p: R vs Y & p: Y vs G \\
\midrule
History & 33.3\% (28.5\%-38.6\%) & 22.7\% (21.0\%-24.5\%) & 14.5\% (11.0\%-18.9\%) & 0.000 & 0.000 \\
Investigations & 29.4\% (19.9\%-41.1\%) & 32.5\% (29.3\%-35.8\%) & 30.7\% (27.6\%-33.8\%) & 0.686 & 0.456 \\
Diagnosis & 46.2\% (40.7\%-51.7\%) & 35.3\% (32.8\%-37.8\%) & 20.5\% (18.4\%-22.9\%) & 0.000 & 0.000 \\
Treatment & 68.1\% (64.1\%-71.9\%) & 54.0\% (51.7\%-56.3\%) & 33.6\% (29.6\%-37.9\%) & 0.000 & 0.000 \\
\bottomrule
\end{tabular}

    \caption{Clinical error rates in cases where the final AI Consult response was red, yellow, or green for the relevant category. $p$-values calculated by Fisher's exact test.}
    \label{tab:ryg_corr}
\end{table}

\paragraph{Inter-rater reliability.} We examined inter-rater agreement using the \nVisitsDoubleRated cases where two physicians independently assigned ratings to the same case, for each of the four Likert types (history, investigations, diagnosis, and treatment). We first examined the within-1 agreement: the proportion of cases in which the two Likert ratings differed by no more than one point. Inter-rater agreement for the history Likert was \historyLikertOnePointProp (95\% CI: \historyLikertOnePointLow - \historyLikertOnePointHigh); for the investigations Likert, it was \investigationsLikertOnePointProp (\investigationsLikertOnePointLow - \investigationsLikertOnePointHigh); for the diagnosis Likert, it was \diagnosisLikertOnePointProp (\diagnosisLikertOnePointLow - \diagnosisLikertOnePointHigh); and for the treatment Likert, it was \treatmentLikertOnePointProp (\treatmentLikertOnePointLow - \treatmentLikertOnePointHigh). Full confusion matrices are in \cref{app:human_rater_study_agreement} (\cref{fig:human_rater_study_agreement}).

We also computed Fleiss' $\kappa$ to examine how much two physician raters agreed as to whether an error was present (i.e., whether a Likert was 1/2 vs 3/4/5) compared to the agreement expected by chance. Fleiss' $\kappa$ ranges from $-1$ to $1$, with negative values indicating less agreement than by chance, zero indicating chance levels of agreement, and positive values indicating more agreement than by chance. Fleiss' $\kappa$ indicated fair agreement between two human raters for each category: \fleissKappaHistory for history errors, \fleissKappaInvestigations for investigation errors, \fleissKappaDiagnosis for diagnosis errors, and \fleissKappaTreatment for treatment errors. 

\paragraph{Language model ratings agree with physician ratings and suggest a stronger AI Consult effect.} \label{sec:ai-analysis}
We were interested in whether our findings were robust to different raters and the quality of large language models' ratings compared to human expert ratings. To examine this, we provided \texttt{GPT-4.1} and \texttt{o3} with the same instructions as our physician raters, asked them to rate clinical documentation, and examined the resulting agreement with human raters and relative risk reduction.

We found that the agreement between model ratings and physician ratings exceeded the agreement between two physicians: for example, the within-1 agreement for history was \GPTFourOneHistoryWithinOneAgreement for \texttt{GPT-4.1} and physicians and \OThreeHistoryWithinOneAgreement for \texttt{o3} and physicians, compared to \historyLikertOnePointProp for physician-physician agreement (\cref{tab:ai_human_agreement}). This was also true for Fleiss' $\kappa$ on whether an error was present: \texttt{GPT-4.1} and physicians had $\kappa = \GPTFourOneHistoryFleissKappa$ and \texttt{o3} and physicians had $\kappa = \OThreeHistoryFleissKappa$, while the physician-physician $\kappa = \fleissKappaHistory $ (\cref{tab:ai_human_kappa}).

Both the \texttt{GPT-4.1} and \texttt{o3} analyses find that AI Consult significantly reduces clinical errors across categories, and generally find larger effect sizes compared to physician raters. For example, physician raters found a treatment error RRR of \mainPeriodLikertErrorRatesRRRTreatment (\mainPeriodLikertErrorRatesRRRTreatmentLowCI - \mainPeriodLikertErrorRatesRRRTreatmentHighCI). For \texttt{GPT-4.1}, the corresponding RRR was \GPTFourOneRRRTreatment (\GPTFourOneRRRTreatmentLowCI-\GPTFourOneRRRTreatmentHighCI) and for \texttt{o3}, it was \OThreeRRRTreatment (\OThreeRRRTreatmentLowCI-\OThreeRRRTreatmentHighCI; \cref{tab:physicians_ai_rrr_table,fig:likert_results_gpt41,fig:likert_results_o3}). 
\begin{table}
    \centering
    \begin{tabular}{llll}
\toprule
 & Physician raters & GPT-4.1 & o3 \\
\midrule
History & 31.8\% (21.9\%-40.5\%) & 46.5\% (42.5\%-50.2\%) & 46.4\% (42.5\%-49.9\%) \\
Investigations & 10.3\% (1.0\%-18.8\%) & 9.9\% (6.8\%-12.9\%) & 13.7\% (9.5\%-17.6\%) \\
Diagnosis & 16.0\% (6.9\%-24.2\%) & 19.4\% (15.5\%-23.2\%) & 16.4\% (12.7\%-19.9\%) \\
Treatment & 12.7\% (6.8\%-18.3\%) & 21.5\% (19.4\%-23.7\%) & 19.1\% (17.1\%-21.1\%) \\
\bottomrule
\end{tabular}

    \caption{Relative risk reductions based on clinical ratings provided by physicians, \texttt{GPT-4.1}, and \texttt{o3}.}
    \label{tab:physicians_ai_rrr_table}
\end{table}

We also ran modified Poisson regression and GEE regression models for both \texttt{GPT-4.1} and \texttt{o3} graders. These show a statistically significant and favorable effect from AI Consult. Modified Poisson models generally have similar effect sizes to the unadjusted analysis; the effect sizes for the GEE models are sometimes similar (e.g., GPT-4.1 diagnosis) and sometimes smaller (e.g., GPT-4.1 treatment), but retain statistical significance. Full regression tables are available in \cref{tab:gpt_history_poisson,tab:gpt_investigations_poisson,tab:gpt_diagnosis_poisson,tab:gpt_treatment_poisson,tab:o3_history_poisson,tab:o3_investigations_poisson,tab:o3_diagnosis_poisson,tab:o3_treatment_poisson,tab:gpt_history_gee,tab:gpt_investigations_gee,tab:gpt_diagnosis_gee,tab:gpt_treatment_gee,tab:o3_history_gee,tab:o3_investigations_gee,tab:o3_diagnosis_gee,tab:o3_treatment_gee}.

\subsection{Use and usability}
\label{sec:results-use-and-usability}

We surveyed clinicians in both groups to ask them how the EMR affects the quality of care that they deliver. We also asked clinicians in the AI group for their feedback on AI Consult. Note that response rates for this anonymous survey were relatively low, with \nSilentAISurveyResponses clinicians in the non-AI group (\pctSilentAISurveyResponses) and \nActiveAISurveyResponses clinicians in the AI group (\pctActiveAISurveyResponses) responding, meaning that these results should be interpreted with caution. 

\begin{figure}[ht]
    \centering
    \begin{subfigure}[t]{0.48\linewidth}
        \centering
        \includegraphics[width=\linewidth]{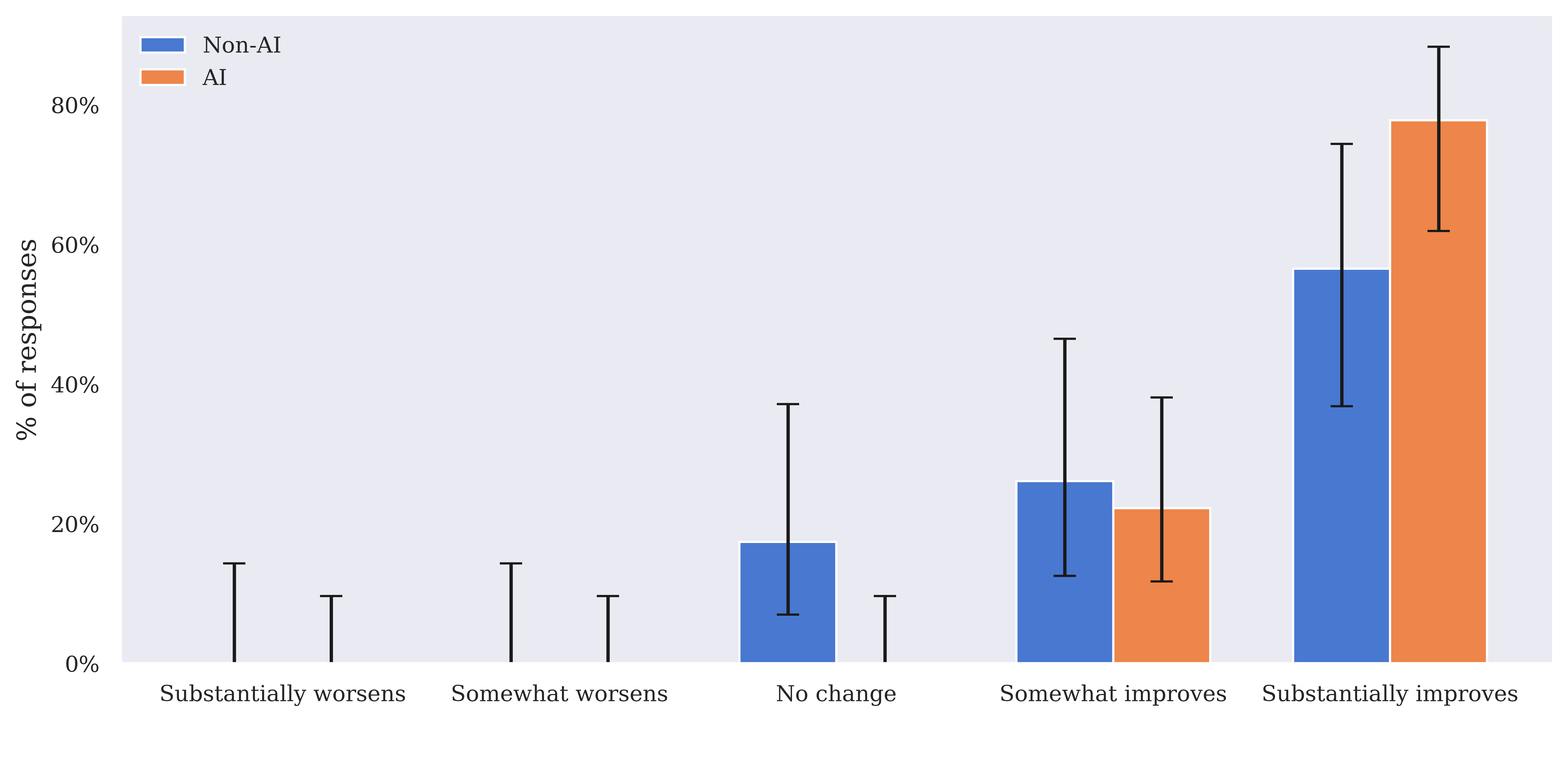}
        \caption{Impact of the EMR (including AI Consult, if present), on quality of care in both the AI and the non-AI group.}
    \end{subfigure}
    \hfill
    \begin{subfigure}[t]{0.48\linewidth}
        \centering
        \includegraphics[width=\linewidth]{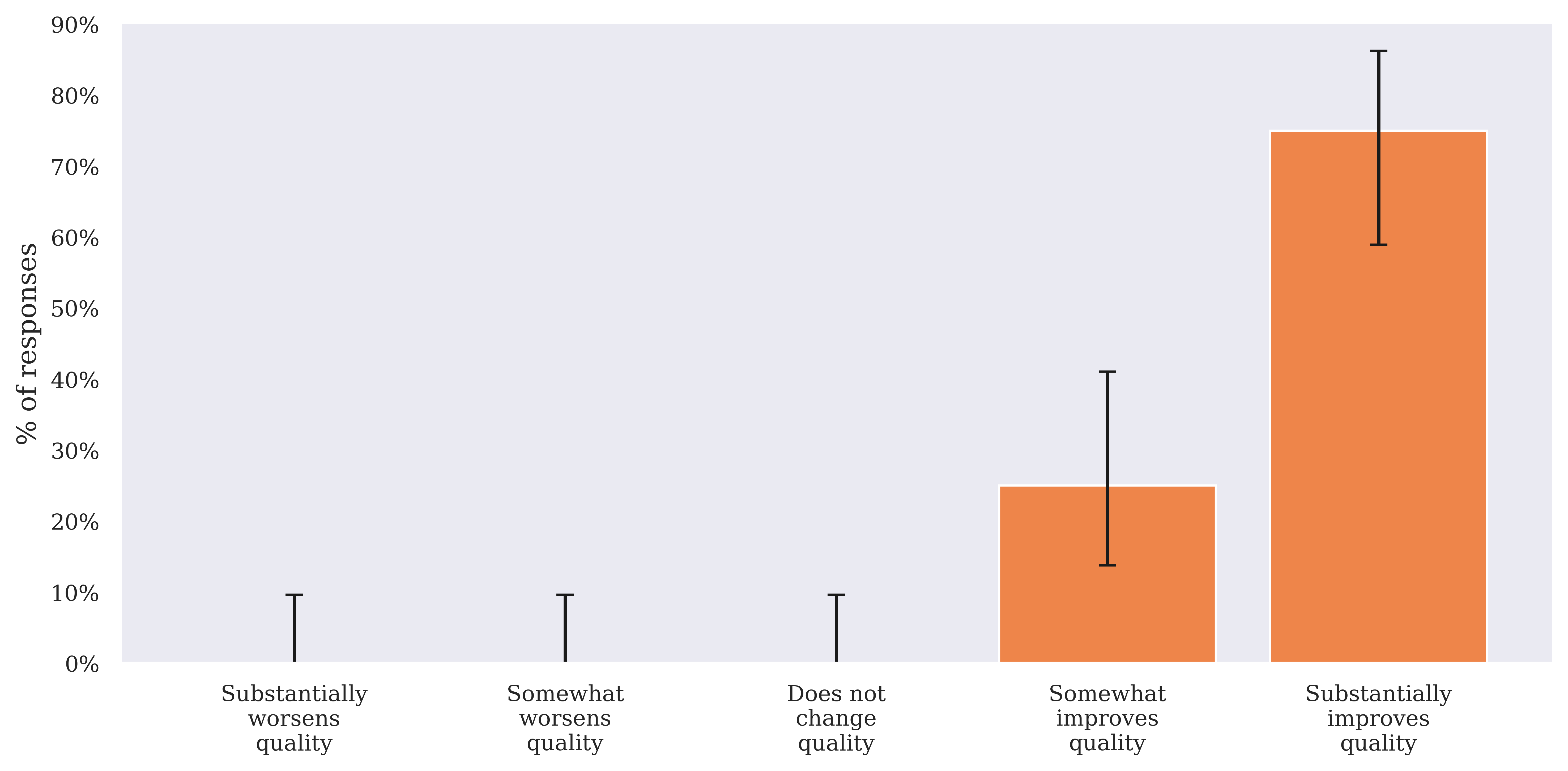}
        \caption{Impact of AI Consult specifically on quality of care.}
    \end{subfigure}
    \caption{Clinician survey results: impact of AI Consult on quality of care.}
    \label{fig:survey}
\end{figure}

\paragraph{Clinicians felt that AI Consult improved quality of care.} Significantly more respondents in the AI group than in the non-AI group noted that the EMR (including AI Consult) improved the quality of care they were able to deliver ($p = \emrQualityUPVal$, \cref{fig:survey}). 

In their qualitative EMR feedback, the \textbf{non-AI group} mostly emphasized operational improvements—speed, tidy documentation, easier stock checks—and in one case asserted that the \textit{EMR doesn’t give option in terms of treatment. The treatment depends on me.} In contrast, the \textbf{AI group} framed the EMR as an active clinical partner: \textit{“It has helped me in multiple occasions to make the correct clinical judgment,”} and highlighted support for \textit{“comprehensive management … from nutrition [to] pharmacological”} alongside provision of \textit{“real-time evidence-based practices”}. Both cohorts stressed time savings, workflow efficiencies, and improved documentation: \textit{“EMR is fast as compared to manual system.”}

\paragraph{Overall feedback on AI Consult was quite positive.} All clinicians in the AI group said that AI Consult improved quality of care, with \activeAISpecificQualityPctFive saying that it substantially improved care (\cref{fig:survey}). Clinician net promoter scores for AI Consult were also favorable, with an overall net promoter score of \activeAISpecificNPS (minimum possible $-100$, maximum possible $100$; for reference, the average net EHR experience score, a similar construct, was 33 in one study across multiple EMR implementations \citep{SuccessfulUserGuide}; \cref{fig:survey_nps}). While satisfaction was generally high, more clinicians noted that they were ``somewhat satisfied'' with AI Consult (\activeAISatisfactionPctFour) than ``very satisfied'' (\activeAISatisfactionPctFive), indicating that room for improvement remains (\cref{fig:survey_satisfaction}). In qualitative feedback, clinicians described AI Consult as “helpful, easy to use, and improves the quality of care.” One clinician “noted an improvement in our clinical notes, which has had a ripple effect on non users of AI” (\cref{tab:quotes}). 

\paragraph{Opportunities for improvement included localization, alert fatigue, and workflow integration.} Constructive feedback covered broader clinical refinement (\textit{“Although there are errors or AI hallucination cases, overall performance … has done tremendous improvement in service delivery”}), error-detection enhancement, and localization needs (e.g., \textit{“keep updating the software to include locally available drugs and management options available in a resource limited medical centers”}). Clinicians also cited alert fatigue and shifting recommendations (\textit{“At some point it keeps on changing the approach of management…”}), documentation burden, and workflow integration gaps (e.g., \textit{“In cases where you give a stat dose… it flags red saying the management is incomplete”}).

\renewcommand{\arraystretch}{1.2}

\begin{table}[htbp]
\centering
\small
\begin{tabularx}{\textwidth}{@{} >{\raggedright\arraybackslash}m{2.75cm} >{\raggedright\arraybackslash}m{0.38\textwidth} >{\raggedright\arraybackslash}m{0.38\textwidth} @{}}

\hline
\textbf{Theme} & \textbf{Representative positive quotes} & \textbf{Representative constructive quotes} \\
\hline

Patient Safety &
\begin{itemize}[leftmargin=*]
\item “It always alerts whenever there is a quality concern… can see the small things we overlook.”
\item “Good reminder in case I miss something.”
\item “I have been able to identify gaps in treatment and this improved treatment quality.”
\item “Acts like a consultant in the room.”
\end{itemize}
&
\begin{itemize}[leftmargin=*]
\item “It improves quality but also can mislead.”
\item “It is not 100\% detecting errors.”
\item “When am prescribing injectables to my patients the AI rates me red even after documenting that my patient vomits everything and can't retain any medication.”
\item “The Ai tool can work on not exaggerating certain conditions that require simple management”
\end{itemize}
\\
\hline

Knowledge and Professional Development &
\begin{itemize}[leftmargin=*]
\item “It’s very informative and broadens my knowledge.”
\item “It sharpens my skills.”
\item “Helps one know when they are on the right track, as it also guides on what next step to take or forgotten inputs.”
\item “It’s also a learning tool.”
\end{itemize}
&
\begin{itemize}[leftmargin=*]
\item “If possible you update it with the current guidelines of management for selective groups e.g. Pregnant mothers.”
\end{itemize}
\\
\hline

Guideline-Based Management and Stewardship &
\begin{itemize}[leftmargin=*]
\item “Has made me be thoughtful on prescriptions of medication that we unnecessarily administer for certain conditions.”
\item “It keeps one in line with the current guidelines.”
\end{itemize}
&
\begin{itemize}[leftmargin=*]
\item “Aligning it more to our protocol and guidelines managements published in Kenya would be amazing.”
\item “Needs to be updated with Kenyan guidelines on disease management… I encountered [issues] on meningitis, heart attack, hypertensive emergency.”
\end{itemize}
\\
\hline

Workflow and Efficiency &
\begin{itemize}[leftmargin=*]
\item “Helps... make better decisions and reduce errors.”
\end{itemize}
&
\begin{itemize}[leftmargin=*]
\item “It takes much time because it requires adequate documentation in history and examination bucket.”
\item “Would be nice if the speed is enhanced and red alerts come before the other alerts.”
\end{itemize}
\\
\hline

Overall Enthusiasm &
\begin{itemize}[leftmargin=*]
\item “It’s one of the best innovation to happen at Penda.”
\item “It should be provided to all health care providers.”
\item “AI is a good idea whose time has come.”
\end{itemize}
&
\begin{itemize}[leftmargin=*]
\item “AI is a good tool in clinicals because it provides thoughtful information… but key factor is to make diagnosis more broader and reducing prompts otherwise it is generally a good tool.”
\end{itemize}
\\
\hline

\end{tabularx}
\caption{Representative clinician user quotes on AI Consult, grouped by theme.}
\label{tab:quotes}
\end{table}

\paragraph{Clinicians in the AI group had longer attending times, which they used to resolve AI Consult triggers and improve quality.} EMR data reveal that the clinician attending time is higher for visits in the AI Consult group (median \AIMedianDuration minutes) compared to the non-AI Consult group (\nonAIMedianDuration minutes; $p = \MedianDurationP$). 

\begin{figure}[ht]
    \centering
    \begin{subfigure}[t]{0.48\linewidth}
        \centering
        \includegraphics[width=0.9\linewidth]{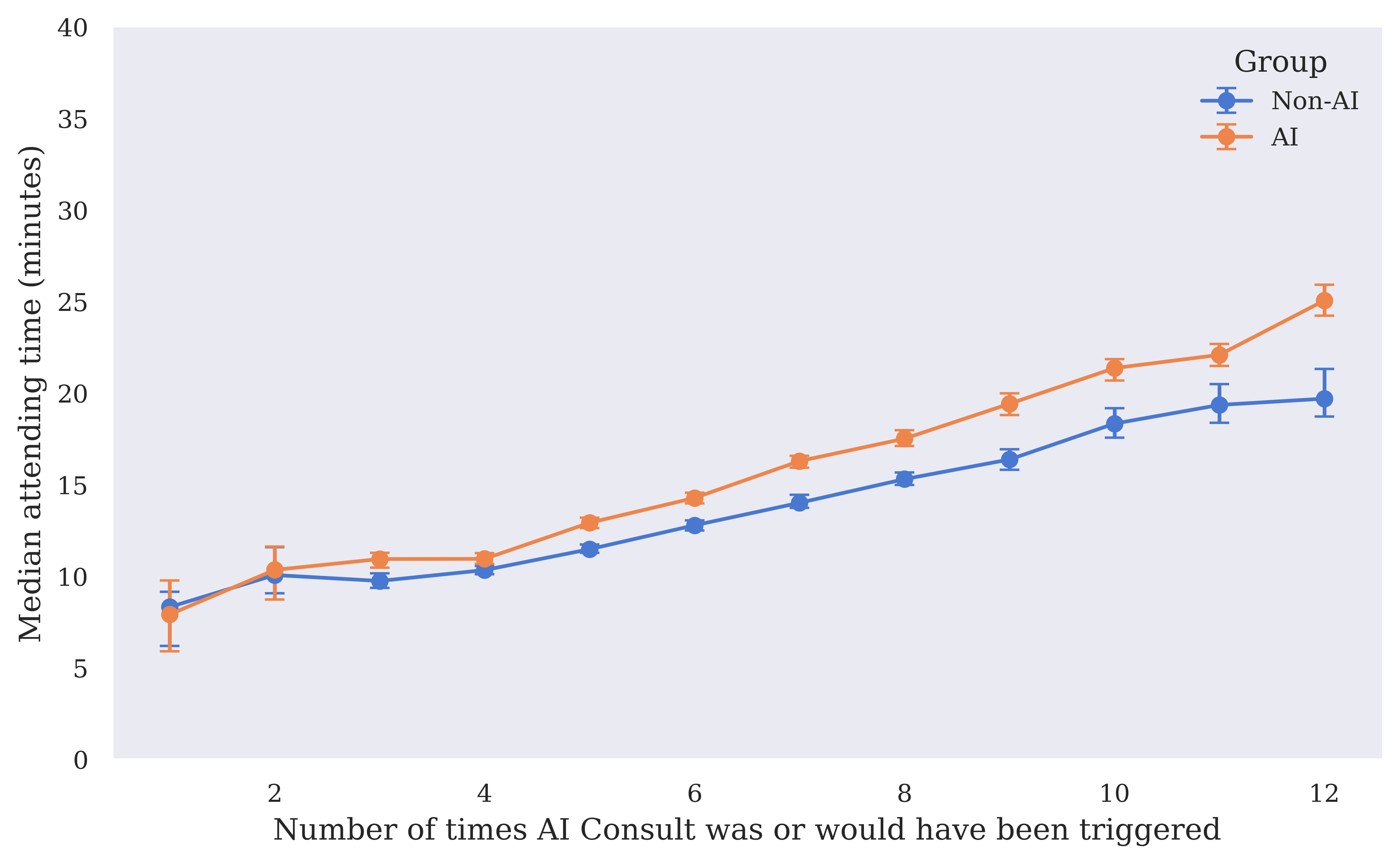}
        \caption{Median clinician attending time by number of AI Consult triggers in the non-AI and AI groups. 95\% CIs calculated with $1000$ bootstrap samples. Includes only visits with 12 or fewer AI Consult calls.}
        \label{fig:duration_by_n_ai_calls}
    \end{subfigure}
    \hfill
    \begin{subfigure}[t]{0.48\linewidth}
        \centering
        \includegraphics[width=0.9\linewidth]{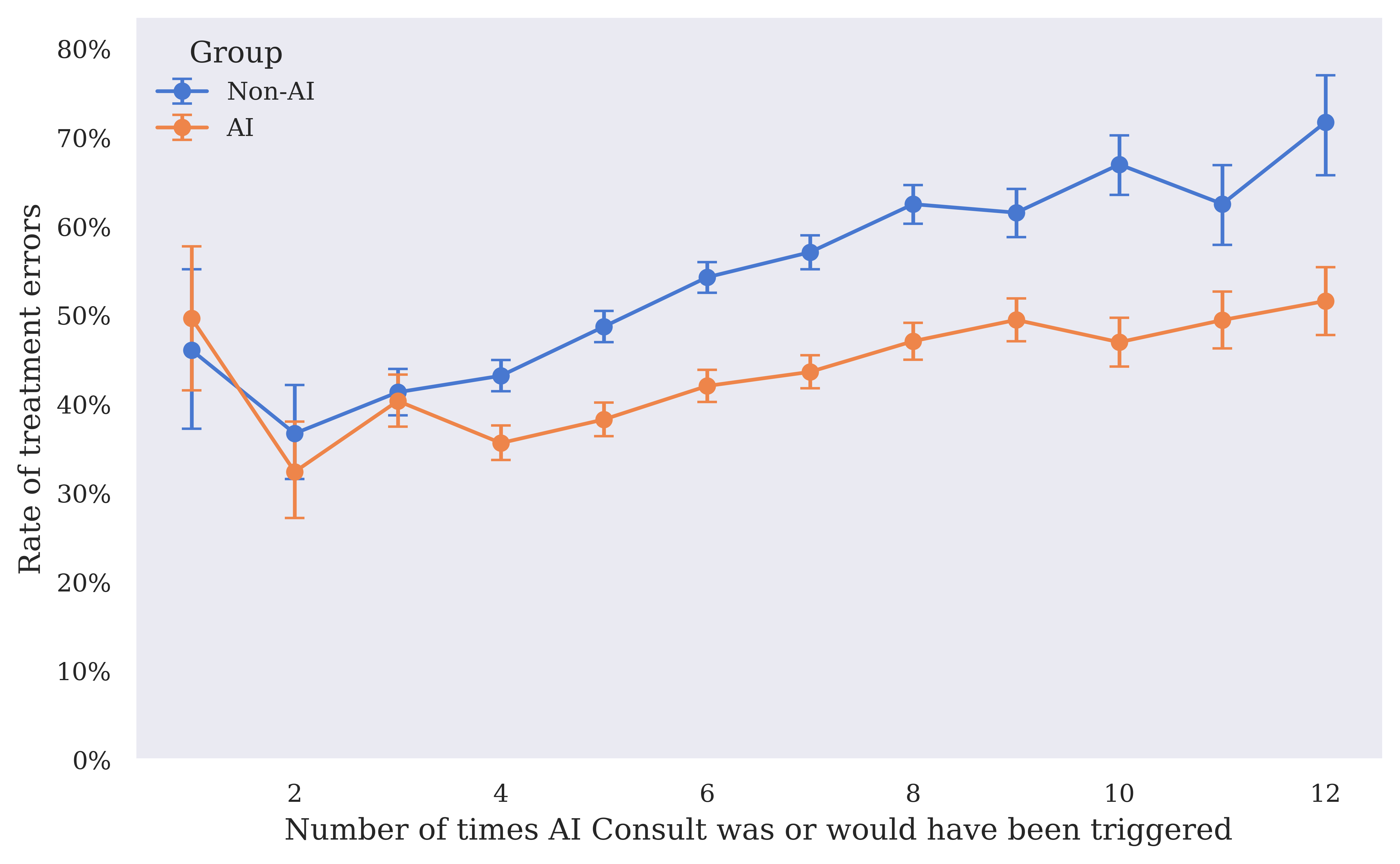}
        \caption{Rate of treatment errors from \texttt{GPT-4.1} by number of AI Consult triggers in the AI vs non-AI groups. 95\% CIs calculated with $1000$ bootstrap samples. Includes only visits with 12 or fewer AI Consult calls.}
        \label{fig:treatment_errors_by_n_ai_calls}
    \end{subfigure}
    \caption{Median clinician attending time and rate of treatment errors by number of AI Consult triggers in the non-AI and AI groups. Results suggest that in visits where there were more AI Consult triggers, clinicians in the AI group spent time responding to AI alerts and made fewer treatment errors as a result.}
    \label{fig:n_ai_calls_panels}
\end{figure}

To examine how AI Consult affects this, we plot median clinician attending time by number of AI Consult triggers in both the non-AI and AI groups (\cref{fig:duration_by_n_ai_calls}). This includes red, yellow, and green triggers, reflecting case complexity as clinicians revisit and change the documentation over time. We see very similar attending times between groups for a small number of AI Consult triggers. The AI group attending time increasingly exceeds the non-AI group attending time with more triggers, suggesting that this increased time is being spent by clinicians in responding to AI Consult feedback. Moreover, we also see improved clinical performance for cases with a higher number of AI calls: the rate of treatment errors (here based on \texttt{GPT-4.1} ratings, to increase sample size and reduce noise) is similar for low numbers of AI Consult triggers and increases more rapidly for the non-AI group compared to the AI group (\cref{fig:treatment_errors_by_n_ai_calls}). We also observe that the rate of treatment errors in the AI group is less than in the non-AI group in visits with the same attending time, suggesting that AI Consult reduces errors even when controlling for visit duration (\cref{fig:treatment_errors_by_visit_duration}).

\paragraph{Clinical notes were typically longer for clinicians in the AI group.}
AI Consult encouraged clinicians to provide more detail in their clinical notes. The median clinical note length was higher in the AI group than the non-AI group over the course of the study (initially, 500 vs 400 characters in the AI vs non-AI group), and this difference grew larger when the active deployment period started (during the week of March 10th, 600 vs 450 characters; at the end of the study, 600 vs 490 characters; \cref{fig:median_clinical_note_length_vs_week}). 

\paragraph{Clinicians generally gave positive feedback on AI Consult responses.}
When clinicians received an AI Consult response, they could give feedback by clicking thumbs up or thumbs down buttons. Among \nAICalls AI Consult responses in the AI group, raters gave feedback on \nAICallsThumbsAny (\pctAICallsThumbsAny). Of these, they gave thumbs up feedback on \nAICallsThumbsUp (\pctAICallsThumbsUp), and thumbs down feedback on the remaining \nAICallsThumbsDown (\pctAICallsThumbsDown). Much of the thumbs down feedback happened within the first two weeks of the induction period, while the prompts were still being iterated on (thumbs down rate of about 13\%); after that period, the thumbs down rate was between 4\% and 7\% in any given week (\cref{fig:thumbs_down_over_time}).

\subsection{Patient outcomes}
\label{sec:results-patient-outcomes}

\subsubsection{Patient-reported outcomes and unplanned follow-up visits}

\paragraph{There were no statistical differences in patient-reported outcomes.} The share of patients who still felt unwell seven days after the visit fell slightly from \notFeelingBetterFirstGroupRate (95\% CI \notFeelingBetterFirstGroupLowerCI–\notFeelingBetterFirstGroupUpperCI) in the non-AI arm to \notFeelingBetterSecondGroupRate (95\% CI \notFeelingBetterSecondGroupLowerCI–\notFeelingBetterSecondGroupUpperCI) in the AI arm (Fisher’s exact test $p = \notFeelingBetterPVal$; \cref{tab:patient_outcomes}), reflecting no statistical difference. The present study was not powered to detect an effect of this magnitude. Moreover, this analysis should be treated as exploratory rather than confirmatory given the high rate of missingness \citep{Jakobsen2017}.

\begin{table}
    \centering
    {\small
\begin{tabular}{llll}
\toprule
 & Rate in non-AI group & Rate in AI group & p \\
\midrule
Rate of patients not feeling better & 4.3\% (3.7\%-4.9\%) & 3.8\% (3.3\%-4.4\%) & 0.234 \\
Saw a pharmacist & 3.5\% (3.0\%-4.1\%) & 3.4\% (2.9\%-3.9\%) & 0.687 \\
Self-referred to another clinic or hospital & 2.9\% (2.4\%-3.4\%) & 3.0\% (2.5\%-3.5\%) & 0.803 \\
Unplanned visit at penda & 6.2\% (5.8\%-6.7\%) & 6.0\% (5.6\%-6.5\%) & 0.532 \\
Feeling worse on one-day follow-up & 2.2\% (0.8\%-6.3\%) & 3.3\% (1.5\%-7.1\%) & 0.737 \\
\bottomrule
\end{tabular}
}
    \caption{Rates of patient outcomes in the AI vs non-AI group.}
    \label{tab:patient_outcomes}
\end{table}

The rates of patients who sought unplanned, unreferred follow-up care was also quite similar between groups (\cref{tab:patient_outcomes}). The sample size of patients who responded to one-day follow-up calls at Penda was quite low (about 500) with low outcome rates, making it challenging to draw conclusions.

\paragraph{Patients were less likely to seek care outside Penda if inappropriate medications were given.} We conducted a post-hoc analysis to investigate how often patients sought care outside Penda depending on whether inappropriate medications were given, as rated by a physician. This rate was lower if inappropriate medications were given (\seekCareGivenInappropriateMedsRate, 95\% CI \seekCareGivenInappropriateMedsRateLowerCI-\seekCareGivenInappropriateMedsRateUpperCI) than if only appropriate medications were given (\seekCareGivenNotInappropriateMedsRate, 95\% CI \seekCareGivenNotInappropriateMedsRateLowerCI-\seekCareGivenNotInappropriateMedsRateUpperCI; $p = \seekCareVsAppropriateTableP$), suggesting that these patient-reported outcomes are tied to patient perception of the care they received and whether they feel their needs for medication were met, regardless of whether the needs were well-justified.

\subsubsection{Patient safety reports}
\label{sec:patient-safety-reports}

Across the 10-week study, 12 patient safety reports (PSRs) were documented: five in the non-AI group and seven in the AI group. Each event was independently reviewed for (i) whether a clinical quality lapse was present, (ii) severity of the event, and (iii) whether AI Consult contributed to harm (if present) or could have mitigated harm.

\paragraph{AI Consult advice could have prevented errors in some cases if available or heeded.} In the non-AI group, three events had AI Consult reds not visible to the clinician that, if visible and followed, might have prevented the lapse (missed anemia work-up, unsafe neonatal prescription, and unrecognized high-risk chest pain). This included one mortality event, in a young adult with chest pain and tachycardia, where AI Consult (which was silent to the clinician) flagged multiple red alerts regarding closer cardiopulmonary evaluation. 

In the AI group, there were similarly three cases where AI Consult issued red or yellow-alert guidance that, if seen by clinicians or heeded, would likely have averted or reduced harm. This also included one mortality event, during the induction period. This event was in an infant with vomiting, fever, and low oxygen saturation. AI Consult produced multiple red alerts recommending respiratory reassessment and oxygen administration. It is unclear whether these alerts were acknowledged or seen by the clinician, as this event occurred early in the induction period before acknowledgment was tracked and when AI Consult red alerts were not reliably visible.

\paragraph{While AI Consult failed to prevent harm in some cases, it did not actively cause harm in any cases.} Three AI group patient safety reports revealed limitations of AI Consult in which it did not prevent harm. In these three cases-pediatric peptic-ulcer misdiagnosis, use of a contraindicated medication in the first trimester of pregnancy, and a missed positive H. pylori test-the AI system failed to suggest a safer alternative. These were all cases where AI Consult did not change the course of the clinical encounter. In no patient safety report did AI Consult make suggestions that created new risk for patients.

\paragraph{AI Consult advice could not have prevented errors in other cases.} Two non-AI group events and one AI group event centered around limited history or documentation. In these visits, AI Consult could not have changed the outcome because the necessary clinical detail was never entered or the patient left before care could be completed. 

\section{Discussion}

Our findings demonstrate that a large language model–based clinical decision support tool can meaningfully reduce diagnostic and treatment errors when deployed in live outpatient care. This improvement occurred not in simulation or review of EMR data, but in the context of routine, real-world practice across nearly 40,000 patient visits in 15 clinics—supporting our view that AI systems, when carefully implemented in clinician workflows, can enhance care quality.

The scale and scope of AI Consult are also notable. Unlike prior decision support systems which target narrow conditions, specialties, or workflows—such as drug interactions or chronic disease screening—AI Consult operated continuously, across all patient visits and key decision points. 

One of the most important implications of this work is the potential for AI tools to further improve the quality of care delivered by primary care clinicians. By functioning as an asynchronous safety net and surfacing real-time feedback at decision points, AI Consult provides lightweight supervision that improved care without undermining clinician autonomy. In this sense, the system serves not only as a quality assurance mechanism but as an empowering tool for clinicians.

Beyond reducing errors in real-time, AI Consult appeared to foster substantial skill gains. During the study period, the proportion of visits that ``started red''—a proxy for clinicians missing a critical issue on first pass—for treatments specifically fell by about 10–15 absolute percentage points in the AI group while remaining flat in the non-AI group. Because these initial alerts precede AI feedback on treatments, the decline signals that clinicians internalized the system’s feedback and preemptively avoided common failure modes. The magnitude of the effect is notable; for every 7-10 patients AI group clinicians saw, they avoided one important initial treatment error. Such learning effects were evident not only for treatment decisions but also for history-taking, suggesting AI Consult facilitates broader learning rather than narrow protocol adherence. These findings, together with survey responses citing the tool as ``very informative,'' ``a learning tool,'' and helpful in ``sharpening my skills,'' support the view that well-designed copilots can function as continuous, case-based education—uplifting individual competence while simultaneously safeguarding patients.

Clinically-aligned implementation was a key factor in the effectiveness of AI Consult. Penda's previous iteration of AI Consult (\cref{sec:digital-infra-penda}) achieved limited uptake because it required clinicians to interrupt the flow of a patient visit to request AI feedback. The iteration we studied here provided a tiered, low-friction interface, enabling broad coverage with minimal disruption and alert fatigue. These changes reflect learnings from the implementation science literature, which has found that avoiding alert fatigue and surfacing CDS recommendations automatically instead of on demand improved clinician adherence \citep{kawamoto2005improving,van2018systematic,seidling2011factors}. Clinician feedback affirmed the utility of the tool–all AI group survey respondents reported that AI Consult improved the quality of care they could deliver–and indicated overall enthusiasm (``It should be provided to all health care providers'').

Active deployment was another key factor for the success of AI Consult. The tool had a significantly greater effect during the main study period (when active deployment strategies were employed) compared to the induction period (\cref{tab:induction_vs_risky_vs_main}), with a clear divergence between AI and non-AI groups for left in red rate and started red rate over the seven weeks of the main period (\cref{fig:reds_over_time_panels}). Based on these strong improvements over a short period, we would expect further gains from longer active deployment efforts. We expect the change management pillars introduced in \cref{sec:methods-active-deployment}–connection, measurement, and incentives–to be similarly important for future AI CDS tools.

Patient safety reports show that AI Consult has clear potential to reduce patient harm. In half of the reviewed reports, harm might have been prevented if AI Consult had been used and its guidance followed. Both deaths reviewed were judged to be potentially preventable with correct AI Consult use. The reports highlight the importance of adherence: AI group users ignored critical alerts in some cases, highlighting the need for improved clinician trust and responsiveness to AI recommendations as part of active deployment efforts. While there were no cases where AI Consult recommendations actively caused harm, in some reports, AI Consult failed to prevent minor or moderate harm, suggesting room for improvement. In other cases, AI Consult was unable to help due to inadequate clinician documentation, emphasizing the importance of clinician buy-in and training.

Localization to Penda's clinical context was important to clinicians.  A considerable amount of variation in medical care can be explained by different norms between facilities and geographies, and systems that are acceptable to end users need to be responsive to this variation. With today's capable and steerable models, localization may not require fine-tuning or specialized models–our experience was that prompting the model to share the Kenyan epidemiological context, provide details about Penda's setting and care protocols, and outline local clinical practice guidelines were all helpful steps towards localization.

\subsection{Limitations}
AI Consult represents an early, promising archetype of an AI-powered clinical copilot. While the results are encouraging, we emphasize that this is a first step. Continued iteration will be essential—to reduce documentation burden, improve contextual relevance, and align more closely with local practice norms. Future implementations may include voice-first interfaces, real-time charting assistants, or agents that execute clinician-confirmed actions in electronic medical records.

Although AI Consult was associated with reduced diagnostic and treatment errors, we did not observe statistically significant differences in patient-reported outcomes during the study period. This may reflect limitations in measurement sensitivity, response rates (response rates were 40\%), short follow-up period, or the relatively short duration of the study. Further work—particularly large studies powered for patient outcomes—will be needed to assess the downstream impact of AI-assisted care.

Physician panel inter-rater reliability was fair but not excellent, despite shared golden examples, multi-step onboarding and quality-filtering, task-specific training, and ongoing ``office hours''. Interestingly, when provided the same form as the physician panel to review visits (\cref{app:rater-form}), o3 and GPT-4.1 both displayed greater rater agreement with physicians than other physicians. Both models also found larger effect sizes for AI Consult than physicians did (\cref{tab:physicians_ai_rrr_table}). While greater effect sizes may be the result of Goodhart's Law (clinician documentation is assessed by an LLM in AI Consult as well), the greater model-physician agreement compared to physician-physician agreement suggests that LLM ratings, if validated via physician agreement on a subset of cases, may be a way to scale up both routine quality improvement and studies like this one.

Clinician survey response rates were somewhat low–63\% of clinicians in the AI group and 47\% in the non-AI group responded. While clinicians were broadly positive about the utility and usability of AI Consult in their responses, they also reported areas of improvement, particularly around response time and localization. We observed that clinicians with many AI Consult triggers had longer visit times but also a greater reduction in treatment errors (\cref{fig:n_ai_calls_panels}), suggesting a quality-time tradeoff in the design and deployment of AI-based CDS tools that needs additional study.

Broader generalizability also requires further research. Penda Health is a particularly strong setting for digital health implementation, given Penda's dedicated technical infrastructure investments and its focus on highly affordable care. Penda's implementation of AI Consult was tailored to its local context, and clinician uptake required active deployment work from Penda's team. Validating and deploying implementations of AI CDS tools in other clinical environments, care settings, and health systems remains an important area for future work.

\section{Related work}
\paragraph{Offline evaluation of LLMs for health.}
Advances in LLMs have spurred many works evaluating them for health applications. Prior works have evaluated health performance broadly \citep{arora2025healthbench,bedi2025medhelm} or for specific tasks, including differential diagnosis \citep{mcduff2025amie,nori2025sequential,goh2025gpt}, clinical summarization \citep{vanveenAdaptedLargeLanguage2024b,zaretskyGenerativeArtificialIntelligence2024}, radiology report generation \citep{tanno2025collaboration,tu2024towardsgeneralist}, and Q\&A \citep{ayers2023comparing,nori2023gpt4,pfohl2024toolbox}. Some works have focused on specialized models \citep{moor2023medflamingo,li2023llava,singhal2023large,singhal2025toward,mcduff2025amie,tu2025towardsconversational,tu2024towardsgeneralist,saab2024capabilities,yang2024advancing} and others on general models \citep{ayers2023comparing,nori2025sequential,nori2023gpt4,saabAdvancingConversationalDiagnostic2025,arora2025healthbench,johnsonAssessingAccuracyReliability2023}. Evaluation in many works relies heavily on narrow automated benchmarks that measure clinical knowledge \citep{nori2023gpt4,singhal2023large}. Some works have evaluated models across many benchmarks, offering more robust characterizations of model performance across tasks \citep{bedi2025medhelm,saab2024capabilities,tu2024towardsgeneralist}. Other works have employed human evaluation with physicians or patients, sometimes employing realistic clinical vignettes or electronic medical record data \citep{goh2025gpt,ong2024development,dash2023evaluation,ayers2023comparing,singhal2025toward,pfohl2024toolbox}. Some recent works have combined human and automated evaluation towards clinician-aligned evaluation at scale \citep{arora2025healthbench,fleming2024medalign}. All of these works involve ``offline'' evaluation of LLMs, which do not enable the study of the unique challenges of bringing model advances into clinical practice, including real-world patient diversity, designing for and learning from clinician workflows, and deployment towards successful clinician uptake. Unlike prior evaluations of LLMs, the present study examines outcomes of using an LLM-based tool live during patient care at scale, addressing the unique challenges of real-world implementation.

\paragraph{Clinical decision support.}
AI Consult is an example of a clinical decision support system. Such systems have been used in various forms since the 1970s \citep{suttonOverviewClinicalDecision2020,middletonClinicalDecisionSupport2016,shortliffeMycinKnowledgeBasedComputer1977,brightEffectClinicalDecisionSupport2012,musenClinicalDecisionSupportSystems2021}. These systems support clinicians with knowledge and tools at the point of care. CDS systems have traditionally drawn on explicit knowledge bases \citep{papadopoulos2022systematic,jing2023ontologies}, typically represented as rules / decision trees \citep{silva2023rule,rommers2013evaluation,gholamzadeh2023application} or cases \citep{althoff1998case,frize2000clinical,kumar2009hybrid}, rather than the distributed representations of LLMs. They have often been restricted in scope to particular conditions \citep{levraLargeLanguageModelbased2025,rajashekarHumanAlgorithmicInteractionUsing2024,onianiEnhancingLargeLanguage2024,kaiserAccuracyConsistencyPublicly2024}, specialties \citep{ong2024development,lammertExpertGuidedLargeLanguage2024,millerComparativeEvaluationLarge2024,benaryLeveragingLargeLanguage2023}, or workflows \citep{slight2013we,kublanov2019development,cheng2013icuarm,rommers2013evaluation}. There are several works that evaluate LLMs ``offline'' for decision support tasks \citep{ong2024development,gaberEvaluatingLargeLanguage2025,lammertExpertGuidedLargeLanguage2024,bhimaniRealWorldEvaluationLarge2025,levraLargeLanguageModelbased2025,rajashekarHumanAlgorithmicInteractionUsing2024,onianiEnhancingLargeLanguage2024,millerComparativeEvaluationLarge2024,kaiserAccuracyConsistencyPublicly2024,benaryLeveragingLargeLanguage2023}; these studies measure model performance on tasks that could support clinicians, using electronic medical record datasets, clinical vignettes, or fictional patients. Similar to other studies evaluating LLMs for health, these do not capture the unique challenges of real-world implementation and deployment. To the best of our knowledge, this is the first study of an LLM-based CDS used in live patient care. Additionally, unlike other CDS systems that provide assistance for targeted workflows or specialties, AI Consult serves to broadly assist primary care clinicians with all major aspects of their patient care workflow, including history-taking, investigations, diagnosis, and treatment.

\paragraph{Implementation science.}
Implementation science examines methods to promote uptake of evidence-based findings into routine care practice and policy \citep{ecclesWelcomeImplementationScience2006,bauer2020implementation,grimshawKnowledgeTranslationResearch2012,olswangBridgingGapResearch2015}. This literature often produces structured frameworks \citep{damschroderFosteringImplementationHealth2009,greenhalghAdoptionNewFramework2017} which identify key factors influencing adoption and sustainability of health interventions. Several works have studied factors specific to clinical decision support \citep{kawamoto2005improving,van2018systematic,castillo2013considerations,murphy2014clinical,kilsdonk2017factors,sittig2006survey,seidling2011factors}. A recurring theme is that technology alone is insufficient to change clinician behavior; effective uptake requires attention to usability, workflow fit, leadership buy-in, iterative training, contextual adaptation, and other factors \citep{rossFactorsThatInfluence2016,smithCodeBedsideImplementing2021,greenhalghAdoptionNewFramework2017,ojoRoleImplementationScience2021}. These insights are particularly critical in low-resource contexts, where health systems face workforce shortages, limited infrastructure, and competing priorities \citep{ojoRoleImplementationScience2021,yapaImplementationScienceResourcepoor2018}. In works studying implementation of clinical decision support, avoiding alert fatigue and surfacing computerized CDS recommendations automatically at relevant points (instead of on demand) have been shown to improve clinician adherence \citep{kawamoto2005improving,van2018systematic,seidling2011factors}. Our implementation design and deployment approach—embedding AI into clinical workflows, iterative user-centered development, automatically surfacing targeted AI responses via a traffic light system, and pairing the tool with change management strategies—takes inspiration from these prior efforts to study the translation of research into clinical impact.

\section{Conclusion}

We have presented a real-world evaluation of a large language model-based clinical decision support tool deployed in live patient care, with a meaningful reduction in diagnostic and treatment errors. Our findings underscore three critical components: (1) capable models, which are now widely available; (2) clinically-aligned implementation, which supports the user rather than distracting them; and (3) active deployment, including building clinician connection, measurement, and incentives. Clinical impact does not emerge solely from model performance, but from a confluence of technical, human, and organizational factors.

With advancements in model capabilities, closing the model-implementation gap has become the most important challenge for the health AI ecosystem. This study provides a template for how AI systems can be safely and effectively embedded into clinical workflows. Further progress requires coordinated efforts across the ecosystem, including policymakers developing regulatory frameworks, engineers designing better implementations, and healthcare systems driving thoughtful deployments. Ultimately, we hope that systems like AI Consult will become the standard of care, supporting clinicians in delivering safer, more consistent, and more accessible care worldwide.

\section*{Code}
We have released code used for analysis and plotting to foster transparency of the results in this study. Raw study data cannot be released due to privacy and data protection requirements. Code can be found at: \url{https://github.com/openai/penda_code}.

\section*{Acknowledgements}
Amelia Glaese, Benjamin Kinsella, Dorothy Cheboi, Lilian Weng, Magdalene Kaisa, Nino Jananashvili, Phoebe Thacker, Rachel Ndiema, Spruce Campbell, Stephanie Koczela, Wyatt Thompson

We would like to thank the following reviewers for generously providing feedback: Ethan Goh, Fred Mutisiya, Isaac Kohane, Nigam Shah, and Steven Wanyee. Any errors are our own.

We would also like to thank the physician reviewers who graded clinical documentation quality for this study.

\clearpage
\bibliography{refs}

\clearpage
\appendix

\FloatBarrier
\section{Images of AI Consult in use}

\begin{figure}[!ht]
    \centering
    \includegraphics[width=\linewidth]{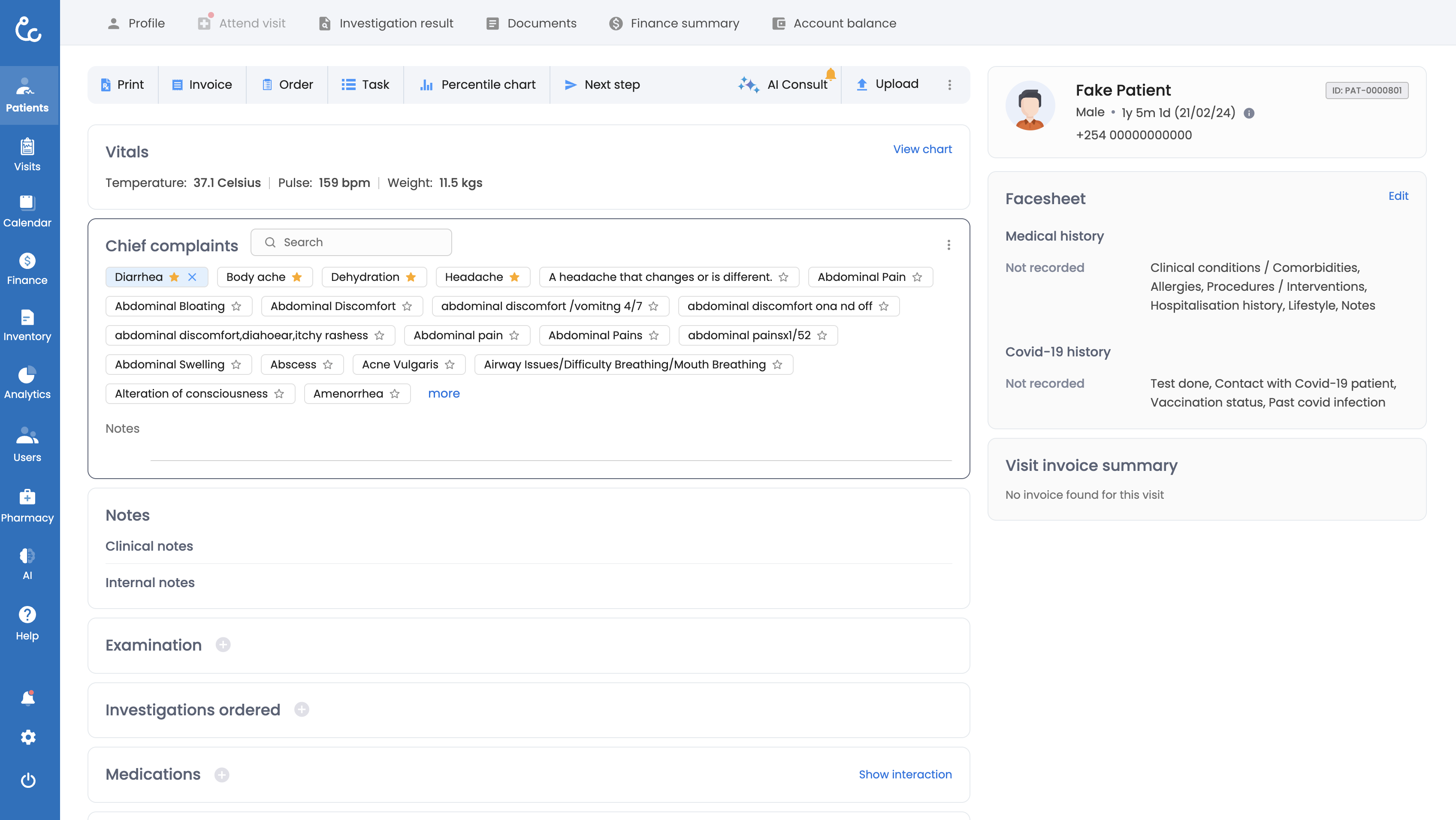}
    \caption{Image of AI Consult yellow notification.}
    \label{fig:ss_yellow_main}
\end{figure}

\begin{figure}[!ht]
    \centering
    \includegraphics[width=\linewidth]{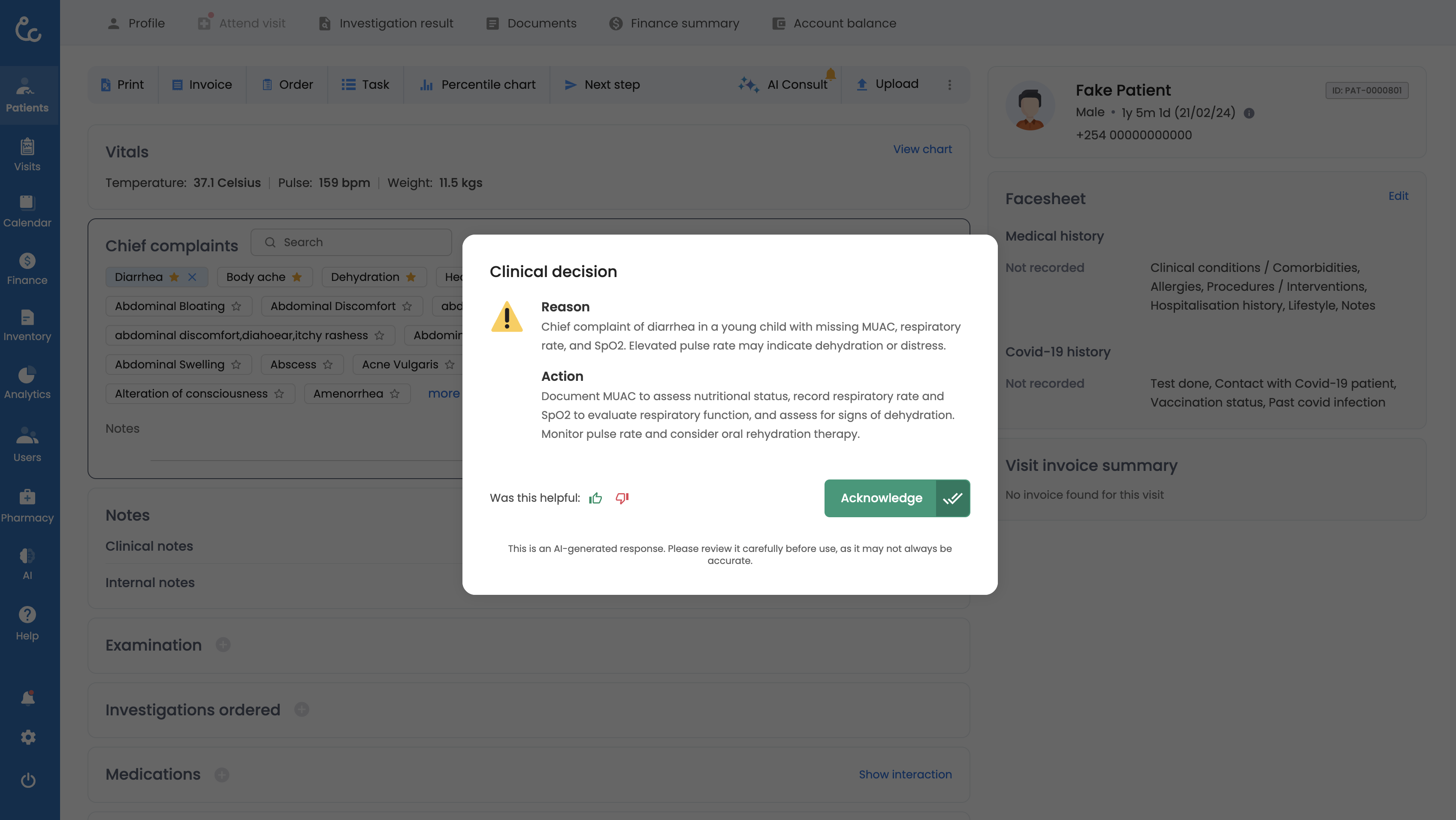}
    \caption{Image of AI Consult yellow popup, after clicking on the notification bell.}
    \label{fig:ss_yellow_click}
\end{figure}

\begin{figure}[!ht]
    \centering
    \includegraphics[width=\linewidth]{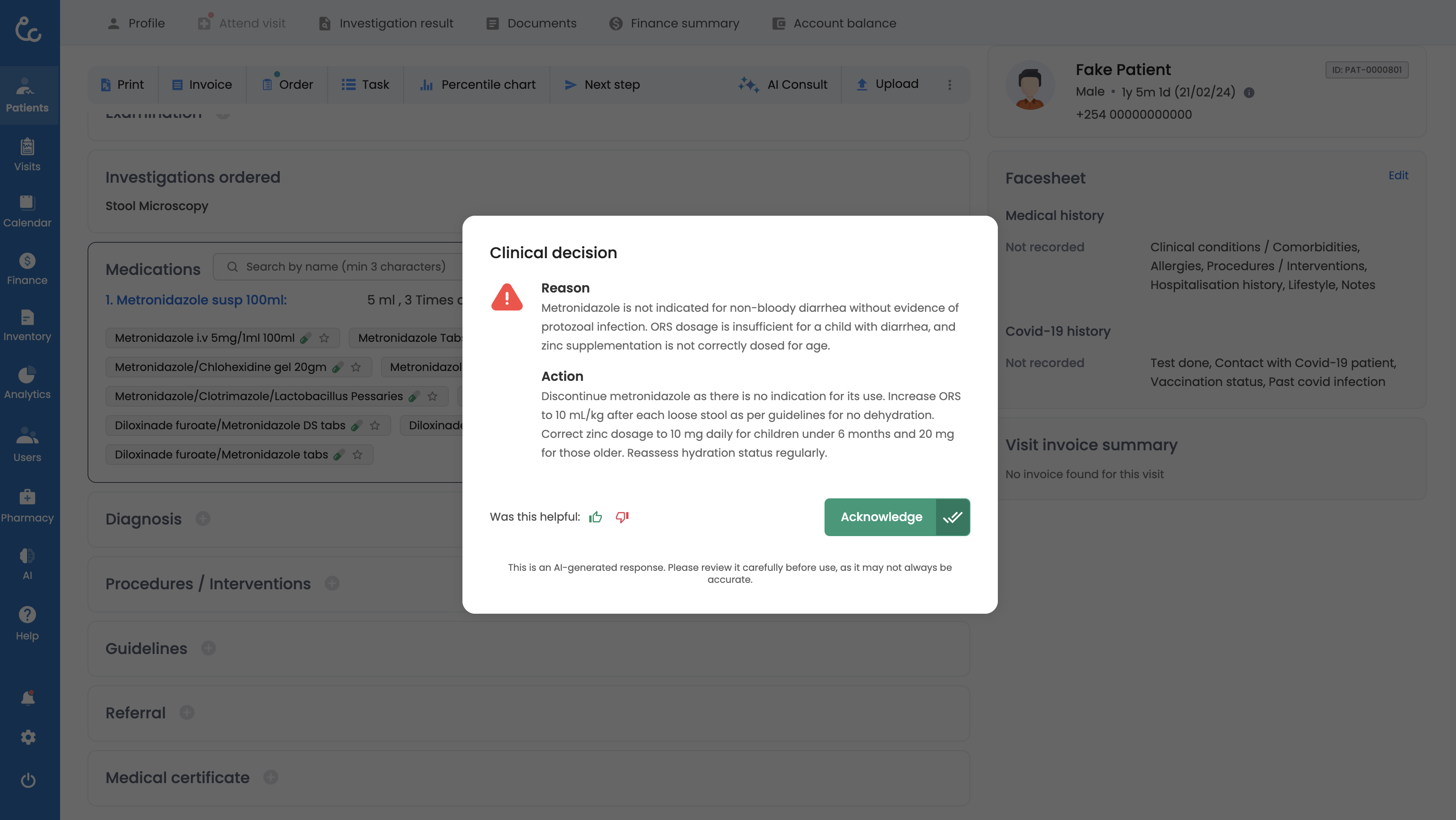}
    \caption{Image of AI Consult red popup.}
    \label{fig:ss_red}
\end{figure}

\begin{figure}[!ht]
    \centering
    \includegraphics[width=\linewidth]{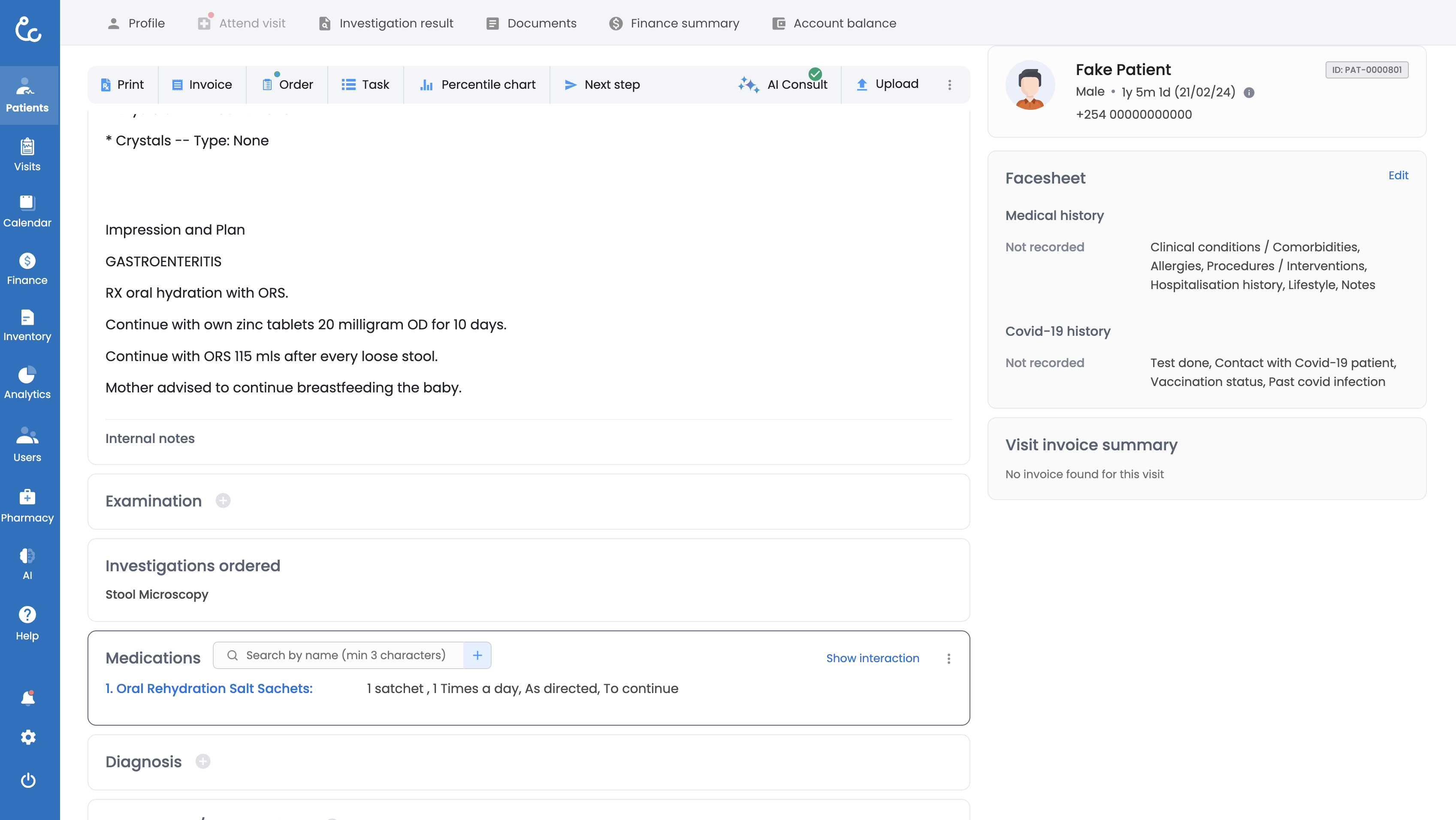}
    \caption{Image of AI Consult green notification.}
    \label{fig:ss_green_main}
\end{figure}

\begin{figure}[!ht]
    \centering
    \includegraphics[width=\linewidth]{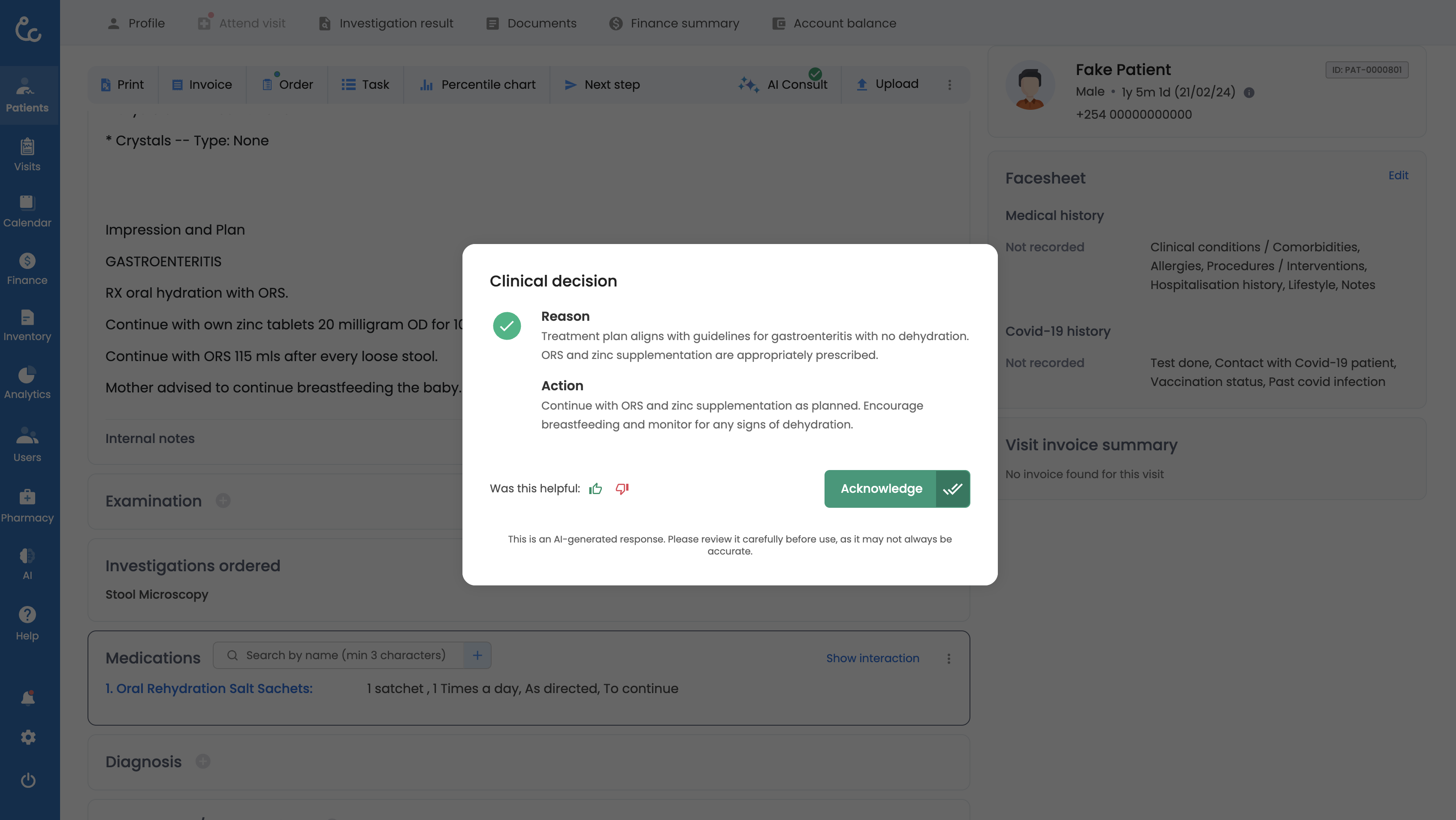}
    \caption{Image of AI Consult green popup, after clicking on the notification bell.}
    \label{fig:ss_green_click}
\end{figure}

\FloatBarrier
\section{Full examples of AI Consult}

\begin{table}[!ht]
\centering
\scriptsize
\setlength{\tabcolsep}{4pt}
\renewcommand{\arraystretch}{1.1}
\begin{tabular}{p{0.31\textwidth}p{0.31\textwidth}p{0.31\textwidth}}
\toprule
\textbf{Initial documentation} &
\textbf{AI Consult response} &
\textbf{Final documentation} \\[4pt]
\midrule
\raggedright
\textbf{Vitals}\\
Age 2 y 6 m, Wt 12.5 kg, HR 127 bpm, MUAC green, Temp 36.3\,$^\circ$C.\\[6pt]
\textbf{Clinical note}\\
mother report th child has been vomiting, 4 episode, non projectile , vomitous contained lasgt eaten meal , no history of fever, no history of cough ,no history of diarrhea, no history of refusal to feed , no history of crying while passing urine , no history of travelling to malaria endemic zone .
\\[4pt]
\textbf{on examination}:
in fair general condition
not pale ,no jaundice ,no cyanosis .
skin pinch return is immediate , capillary refill return is less than 2 sec, no dryness of mucus membrane\\
Per abdomen-normal fullness, no palpable mass, no tenderness\\
Respiratory-chest clear, no respiratory distress, no crackles, no rhonchi\\
Ear, nose, and throat-inflamed, non exudative ,non erythematous throat , no adenoids inflammation , no inflammation in the ear\\
Cardiovascular- s1s2 heard no murmur\\
Neurologic-AVPU@A , neck soft , no sign of meningeal irritation\\[6pt]
\textbf{Investigations}\\
Full Hemogram:\\
WBC 12.38\\ \sethlcolor{blue!25}\textbf{\hl{HGB 9.90}}\\
HCT 30.70\\
Plt 248\\
RBC 5.26\\ \textbf{\hl{MCV 58.30}}\\MCH 18.8\\ MCHC 32.2.\\
Strep-A Ag negative.\\[6pt]
\textbf{Diagnosis}\\
Tonsillitis, acute bacterial.\\
\sethlcolor{blue!25}\textbf{\hl{CLINICIAN DOES NOT INCLUDE A DIAGNOSIS RELATED TO THE ABNORMAL FINDING OF MICROCYTIC ANEMIA}}\\[4pt] &
\raggedright
\textbf{\textcolor{red}{RED FLAG}}\\[4pt]
\textbf{Responses}:
The clinical documentation shows the presence of 
\sethlcolor{blue!25}
\hl{microcytic anemia with a hemoglobin level of 9.90 g/dL and an MCV of 58.30 fL}, indicative of \hl{anemia that is not addressed in the diagnosis}. The presence of microcytic anemia should be evaluated alongside the current symptom of vomiting.\\[4pt]

\textbf{Recommendations}:
\sethlcolor{blue!25}
\hl{Reevaluate the diagnosis considering the significant finding of microcytic anemia, which is not addressed}. Investigate underlying causes of anemia, such as iron deficiency or chronic disease. Consider additional testing for iron studies or further dietary investigation. Ensure anemia is included in the management plan to address this critical finding.

&
\raggedright
\textbf{Vitals}\\
Age 2 y 6 m, Wt 12.5 kg, HR 127 bpm, MUAC green, Temp 36.3\,$^\circ$C.\\[6pt]
\textbf{Clinical note}\\
mother report th child has been vomiting, 4 episode, non projectile , vomitous contained lasgt eaten meal , no history of fever, no history of cough ,no history of diarrhea, no history of refusal to feed , no history of crying while passing urine , no history of travelling to malaria endemic zone .
\\[4pt]
\textbf{on examination}:
in fair general condition
not pale ,no jaundice ,no cyanosis .
skin pinch return is immediate , capillary refill return is less than 2 sec, no dryness of mucus membrane\\
Per abdomen-normal fullness, no palpable mass, no tenderness\\
Respiratory-chest clear, no respiratory distress, no crackles, no rhonchi\\
Ear, nose, and throat-inflamed, non exudative ,non erythematous throat , no adenoids inflammation , no inflammation in the ear\\
Cardiovascular- s1s2 heard no murmur\\
Neurologic-AVPU@A , neck soft , no sign of meningeal irritation\\[6pt]
\textbf{Investigations}\\
Full Hemogram:\\
WBC 12.38\\ {HGB 9.90}\\
HCT 30.70\\
Plt 248\\
RBC 5.26\\ MCV 58.30\\MCH 18.8\\ MCHC 32.2.\\
Strep-A Ag negative.\\[6pt]
\textbf{Diagnoses}\\
Tonsillitis, acute bacterial; Emesis, unspecified; 
\sethlcolor{blue!25}\textbf{\hl{Iron deficiency anemia}}\\[4pt]
\end{tabular}

\caption{Example of AI Consult in action: AI Consult flags an important missing diagnosis of iron deficiency anemia, leading the clinician to add this diagnosis so it can be appropriately treated. Highlighted portions show the initial text that triggered the tool, important messages in the AI Consult response and changes the clinician made after reviewing the flag.}
\label{tab:ai_consult_example}
\end{table}

\begin{table}[!ht]
\centering
\scriptsize
\setlength{\tabcolsep}{4pt}
\renewcommand{\arraystretch}{1.1}
\begin{tabular}{p{0.31\textwidth}p{0.31\textwidth}p{0.31\textwidth}}
\toprule
\textbf{Initial documentation} &
\textbf{AI Consult response} &
\textbf{Final documentation} \\[4pt]
\midrule
\raggedright
\textbf{Vitals}\\
Age 1 y 5 m, Wt 11.5 kg, HR 159 bpm, MUAC green, Temp 37.1\,$^\circ$C.\\[6pt]
\textbf{Clinical note}\\
CHild was treated yesterday with vomiting now subsided.\\
Mother reports developed diarrhea today morning reports several episodes since morning\\
Stool is waterly mucoid stained but not bloody though smelly\\ normal urine outpu.
No associated fevers\\
No history of travel to malaria zone\\
Baby breastfeeding well but has reduced appetite today
\\[4pt]
\textbf{on examination}:
no dehydration nornal skin turgor normal cap refill no pallor\\ 
Per abdomen: normal abdominal fullness, no distension no tenderness\\[6pt]
\textbf{Investigations}\\
Stool Microscopy:\\ 
* Consistency: Liquid/Loose\\
\sethlcolor{blue!25}\textbf{\hl{* Blood: Absent}}\\
* Mucous: Present\\
* RBC's (Microscopy): None\\
\sethlcolor{blue!25}\textbf{\hl{* Parasites: No Ova/Cyst\\
* Crystals -- Amount: None\\
* Crystals -- Type: None}}\\[6pt]
\textbf{Impression and Plan}\\
Gastroenteritis\\
RX oral hydration with ORS\\ 
Contine with  own zinc tablets 20 milligram DO for 10 days\\
\sethlcolor{blue!25}\textbf{\hl{Metronidazole susp 100ml dosage: 5 ml, 3 Times a day, After food for 5 Days, Quantity 1 }}\\
Oral Rehydration Salt Sachets dosage: 1 satchet, 1 Times a day, As directed for 3 Days, Quantity 3 \\[4pt] &
\raggedright
\textbf{\textcolor{red}{RED FLAG}}\\[4pt]
\textbf{Responses}:
\sethlcolor{blue!25}
\hl{The treatment involves the use of metronidazole, which is not indicated in uncomplicated gastroenteritis with no identified protozoal infection. This could lead to unnecessary exposure to antibiotics.}\\[4pt]

\textbf{Recommendations}:
\sethlcolor{blue!25}
\hl{Discontinue metronidazole as there is no evidence of protozoan infection.}\\
Focus on continued oral rehydration with ORS and administer zinc supplements as planned. Monitor the child's hydration and general condition, and educate the caregiver on signs of dehydration requiring medical attention.\\

&
\raggedright
\textbf{Vitals}\\
Age 1 y 5 m, Wt 11.5 kg, HR 159 bpm, MUAC green, Temp 37.1\,$^\circ$C.\\[6pt]
\textbf{Clinical note}\\
CHild was treated yesterday with vomiting now subsided.\\
Mother reports developed diarrhea today morning reports several episodes since morning\\
Stool is waterly mucoid stained but not bloody though smelly\\ normal urine outpu.
No associated fevers\\
No history of travel to malaria zone\\
Baby breastfeeding well but has reduced appetite today
\\[4pt]
\textbf{on examination}:
no dehydration nornal skin turgor normal cap refill no pallor\\ 
Per abdomen: normal abdominal fullness, no distension no tenderness\\[6pt]
\textbf{Investigations}\\
Stool Microscopy:\\ 
* Consistency: Liquid/Loose\\
* Blood: Absent\\
* Mucous: Present\\
* RBC's (Microscopy): None\\
* Parasites: No Ova/Cyst\\
* Crystals -- Amount: None\\
* Crystals -- Type: None\\[6pt]
\textbf{Impression and Plan}\\
Gastroenteritis\\
RX oral hydration with ORS
Contine with own zinc tablets 20 milligram OD for 10 days\\
\sethlcolor{blue!25}\textbf{\hl{CLINICIAN REMOVED THE METRONIDAZOLE PRESCRIPTION}}\\
Continue with ORS 115 mls after every loose stool.
Mother  advised continue breastfeeding the baby\\[4pt]
\end{tabular}
\\[2pt]
\caption{Example of AI Consult in action: AI Consult suggestions help the clinician identify and remove an unnecessary antibiotic prescription.}
\label{tab:ai_consult_example_2}
\end{table}

\begin{table}[!ht]
\centering
\scriptsize
\setlength{\tabcolsep}{4pt}
\renewcommand{\arraystretch}{1.1}
\begin{tabular}{p{0.31\textwidth}p{0.31\textwidth}p{0.31\textwidth}}
\toprule
\textbf{Initial documentation} &
\textbf{AI Consult response} &
\textbf{Final documentation} \\[4pt]
\midrule
\raggedright
\textbf{Vitals}\\
Age 26 y 10 m, Wt 76.5 kg, HR 77 bpm, RR 18 bpm, SpO2 98\%, Temp 36.3\,$^\circ$C.\\[6pt]
\textbf{Clinical note}\\
report history of dry cough ,which is on and off occurs mostly in the evening this is associated with dizziness for 2/7 now, this was preceded with headache which resolved after taking pain killer, 3/7 ago . above is accompanied with fatigue , he associates this with overworking since hes been working from 4am - midnight for the past 2/52 without rest . Has history of nasal blockage , no history of runny nose, no history of fever. no history of dib . no history of travelling to malaria endemic zone .
\\[4pt]
\textbf{on examination}:
in fair general condition
not pale ,no jaundice ,no cyanosis ,no deho\\
Ear, nose and throat-no inflamed tonsils, no exudation , no inflamed sinus, no ear infection\\
Respiratory-chest clear , no crackles, no rhonchi\\
Neurologic-GCS15/15 , No sign of meningeal irritation , no sign of neurologic deficit\\
Cardiovascular-s1s2heard no murmur\\[6pt]
\textbf{Diagnosis}\\
Upper respiratory tract infection (URTI).\\[6pt]
\textbf{Medications}\\
\sethlcolor{blue!25}\textbf{\hl{Betamethasone/Dexchlorpheniramine maleate tabs dosage: 1 tablet, 2 Times a day, After food for 5 Days,  dosage: 1 application}}\\[4pt] &
\raggedright
\textbf{\textcolor{red}{RED FLAG}}\\[4pt]
\textbf{Responses}:
The treatment plan includes \sethlcolor{blue!25}
\hl{Betamethasone/Dexchlorpheniramine, which is not recommended} for URTI due to potential side effects and lack of efficacy. Additionally, the \sethlcolor{blue!25}
\hl{use of sedating antihistamines like Dexchlorpheniramine is not advisable} for treating URTI symptoms.
.\\[4pt]

\textbf{Recommendations}:
\sethlcolor{blue!25}
\hl{Discontinue Betamethasone/Dexchlorpheniramine}. Consider symptomatic \sethlcolor{blue!25}
\hl{treatment with non-sedating antihistamines} if necessary, and advise on rest, hydration, and steam inhalation as supportive care.

&
\raggedright
\textbf{Vitals}\\
Age 26 y 10 m, Wt 76.5 kg, HR 77 bpm, RR 18 bpm, SpO2 98\%, Temp 36.3\,$^\circ$C.\\[6pt]
\textbf{Clinical note}\\
report history of dry cough ,which is on and off occurs mostly in the evening this is associated with dizziness for 2/7 now, this was preceded with headache which resolved after taking pain killer, 3/7 ago . above is accompanied with fatigue , he associates this with overworking since hes been working from 4am - midnight for the past 2/52 without rest . Has history of nasal blockage , no history of runny nose, no history of fever. no history of dib . no history of travelling to malaria endemic zone .
\\[4pt]
\textbf{on examination}:
in fair general condition
not pale ,no jaundice ,no cyanosis ,no deho\\
Ear, nose and throat-no inflamed tonsils, no exudation , no inflamed sinus, no ear infection\\
Respiratory-chest clear , no crackles, no rhonchi\\
Neurologic-GCS15/15 , No sign of meningeal irritation , no sign of neurologic deficit\\
Cardiovascular-s1s2heard no murmur\\[6pt]
\textbf{Diagnosis}\\
Upper respiratory tract infection (URTI).\\[6pt]
\textbf{Medications}\\
\sethlcolor{blue!25}\textbf{\hl{CLINICIAN REMOVED THE BETAMETHASONE/DEXCHLORPHENIRAMINE PRESCRIPTION}}\\
Cetrizine tabs 10mg: 1 application, 1 Times a day, After food for 5 Days\\
Xylometazoline Adult Drops 10ml: 2 drop, 3 Times a day, Morning, Afternoon, Night for 3 Days\\[4pt]
\end{tabular}
\\[2pt]
\caption{Example of AI Consult in action: AI Consult identified potentially harmful use of a combination steroid / sedating antihistamine, leading the clinician to replace this with more appropriate, non-sedating medications.}
\label{tab:ai_consult_example_3}
\end{table}

\begin{table}[!ht]
\centering
\scriptsize
\setlength{\tabcolsep}{4pt}
\renewcommand{\arraystretch}{1.1}
\begin{tabular}{p{0.31\textwidth}p{0.31\textwidth}p{0.31\textwidth}}
\toprule
\textbf{Initial documentation} &
\textbf{AI Consult response} &
\textbf{Final documentation} \\[4pt]
\midrule
\raggedright
\textbf{Vitals}\\
Age 0 y 10 m, Wt 8.9 kg, HR 118 bpm, MUAC green, RR 26 bpm, Temp 36.6\,$^\circ$C.\\[6pt]
\textbf{Clinical note}\\
child has been accompanied with 2 day history of mouth sores\\
mother also reports child with 3 day history of loose stool , started yesterday gradually mother reports stool is soft, mucoid , non blood stained\\
\sethlcolor{blue!25}\textbf{\hl{CLINICIAN DOES NOT INCLUDE THE FREQUENCY OF LOOSE STOOLS}}\\
no vomiting \\
also reports to have noticed signs of teething\\
pre  med: nystatin used for 2 days now\\
no known food or drug allergy\\[4pt]
\textbf{on examination}:
stable ,not pale, no jaundice, not dehydrated 
able to drink adequately, no sunken eyes\\ capillary refill: immediately\\
oral exam has oral thrush in the mouth\\
other systems essentially normal\\[6pt]
 &
\raggedright
\textbf{\textcolor{yellow!60!red}{YELLOW FLAG}}\\[4pt]
\textbf{Responses}:
Documentation is generally sufficient but \sethlcolor{blue!25}\hl{lacks details on the frequency and quantity of loose stools, which are important for assessing dehydration risk in a child with gastroenteritis.}
\\[4pt]

\textbf{Recommendations}:
\sethlcolor{blue!25}
\hl{Document the frequency and quantity of loose stools to better assess the risk of dehydration and guide management.}

&
\raggedright
\textbf{Vitals}\\
Age 0 y 10 m, Wt 8.9 kg, HR 118 bpm, MUAC green, RR 26 bpm, Temp 36.6\,$^\circ$C.\\[6pt]
\textbf{Clinical note}\\
child has been accompanied with 2 day history of mouth sores\\
mother also reports child with 3 day history of loose stool , started yesterday gradually mother reports stool is soft, mucoid , non blood stained\\
\sethlcolor{blue!25}\textbf{\hl{reports child with loose stool after 4 to 6hrs , less amount , not watery.
}}\\
no vomiting \\
also reports to have noticed signs of teething\\
pre  med: nystatin used for 2 days now\\
no known food or drug allergy\\[4pt]
\textbf{on examination}:
stable ,not pale, no jaundice, not dehydrated 
able to drink adequately, no sunken eyes\\ capillary refill: immediately\\
oral exam has oral thrush in the mouth\\
other systems essentially normal\\[6pt]
\end{tabular}

\caption{Example of AI Consult in action: AI Consult suggestions help the clinician seek important missing history which will help them know what treatment to give.}
\label{tab:ai_consult_example_4}
\end{table}

\begin{table}[!ht]
\centering
\scriptsize
\setlength{\tabcolsep}{4pt}
\renewcommand{\arraystretch}{1.1}
\begin{tabular}{p{0.31\textwidth}p{0.31\textwidth}p{0.31\textwidth}}
\toprule
\textbf{Initial documentation} &
\textbf{AI Consult response} &
\textbf{Final documentation} \\[4pt]
\midrule
\raggedright
\textbf{Vitals}\\
Age 30 y 2 m, Wt 69.5 kg, HR 102 bpm, BP 105/72, Temp 36.3\,$^\circ$C.\\[6pt]
\textbf{Clinical note}\\
presented with history of per vaginal itchiness on and off for the past 3 days\\ 
accompanied by whitish per vaginal discharge non smelly non bloody\\
later back pains,no per vaginal spotting,nopainful urination,no urine frequency,no lower abdominal pains\\[4pt]
\textbf{on examination}:
stable,afebrile,not pale,not dehydrated,not jaundiced\\
per abdomen-soft none tender,no guarding,no rebound tenderness\\
respiratory-clear chest,no creps,no rhonchi\\
cardiovascular-s1 and s2 heard,no murmur\\
central nervous system-gcs 15/15,no neck stiffness,soft neck\\
other systems-essentially normal\\[6pt]
\textbf {Investigations}\\
Urine Analysis\\
\sethlcolor{blue!25}\textbf{\hl{CLINICIAN DOES NOT ORDER A TEST TO INVESTIGATE THE VAGINAL DISCHARGE}}\\
 &
\raggedright
\textbf{\textcolor{yellow!60!red}{YELLOW FLAG}}\\[4pt]
\textbf{Responses}:
The urine analysis is a common preliminary test that can help rule out urinary tract causes of symptoms. However, given the primary symptoms are related to vaginal discharge and itchiness,\sethlcolor{blue!25}\hl{a vaginal swab for microscopy, culture, and sensitivity would be more directly relevant to identify possible causes} like yeast infection or bacterial vaginosis.
\\[4pt]

\textbf{Recommendations}:
\sethlcolor{blue!25}
\hl{Consider ordering a vaginal swab for microscopy, culture, and sensitivity to better identify the cause of the vaginal discharge and itchiness.}

&
\raggedright
\textbf{Vitals}\\
Age 30 y 2 m, Wt 69.5 kg, HR 102 bpm, BP 105/72, Temp 36.3\,$^\circ$C.\\[6pt]
\textbf{Clinical note}\\
presented with history of per vaginal itchiness on and off for the past 3 days\\ 
accompanied by whitish per vaginal discharge non smelly non bloody\\
later back pains,no per vaginal spotting,nopainful urination,no urine frequency,no lower abdominal pains\\[4pt]
\textbf{on examination}:
stable,afebrile,not pale,not dehydrated,not jaundiced\\
per abdomen-soft none tender,no guarding,no rebound tenderness\\
respiratory-clear chest,no creps,no rhonchi\\
cardiovascular-s1 and s2 heard,no murmur\\
central nervous system-gcs 15/15,no neck stiffness,soft neck\\
other systems-essentially normal\\[6pt]
\textbf {Investigations}\\
Urine Analysis\\
\sethlcolor{blue!25}\textbf{\hl{High Vaginal Swab (HVS Gram Stain)}}
\end{tabular}

\caption{Example of AI Consult in action: AI Consult flags a missing test to evaluate the cause of vaginal itchiness and discharge, resulting in the clinician ordering this test.}
\label{tab:ai_consult_example_5}
\end{table}

\begin{table}[!ht]
\centering
\scriptsize
\setlength{\tabcolsep}{4pt}
\renewcommand{\arraystretch}{1.1}
\begin{tabular}{p{0.31\textwidth}p{0.31\textwidth}p{0.31\textwidth}}
\toprule
\textbf{Initial documentation} &
\textbf{AI Consult response} &
\textbf{Final documentation} \\[4pt]
\midrule
\raggedright
\textbf{Vitals}\\
Age 16 y 4 m, Wt 45 kg, HR 112 bpm, BP 98/61, RR 17 bpm, Temp 38.3\,$^\circ$C.\\[6pt]
\textbf{Clinical note}\\
high fevers, exposed\\
parafast 1g Per oral administered\\    
Brought in by the mother with 1 day history of painful throat, chills, fever and left sided headache.\\
no cough or runny nose.\\
no gastrointestinal complainr.\\
no bladder disturbances. no travel history\\[4pt]
\textbf{on examination}:
fair condition, febrile, no pallor, no icterus, no dehydration\\
Respiratory; normal breath sounds\\
Cardiovascular: normal blood pressure and pulse\\
Ear, nose and throat; tonsillar exudate\\ 
Other systems; unremarkable\\[6pt]
\textbf {Investigations}\\
\sethlcolor{blue!25}\textbf{\hl{Streptococcus A  Antigen Test: Positive}}\\[4pt]
\sethlcolor{blue!25}\textbf{\hl{Diagnosis : Group A Strep pharyngitis score 3-4: }}
 &
\raggedright
\textbf{\textcolor{green!80!black}{GREEN FLAG}}\\[4pt]
\textbf{Responses}:
\sethlcolor{blue!25}\hl{The diagnosis of Group A Strep pharyngitis aligns with the clinical presentation of throat pain, fever, and positive Streptococcus A antigen test.}
\\[4pt]

\textbf{Recommendations}:
Proceed with appropriate antibiotic therapy for Group A Strep pharyngitis, such as penicillin or amoxicillin, and provide supportive care for symptoms.\\

&
\raggedright
\textbf{Vitals}\\
Age 16 y 4 m, Wt 45 kg, HR 112 bpm, BP 98/61, RR 17 bpm, Temp 38.3\,$^\circ$C.\\[6pt]
\textbf{Clinical note}\\
high fevers, exposed\\
parafast 1g Per oral administered\\    
Brought in by the mother with 1 day history of painful throat, chills, fever and left sided headache.\\
no cough or runny nose.\\
no gastrointestinal complainr.\\
no bladder disturbances. no travel history\\[4pt]
\textbf{on examination}:
fair condition, febrile, no pallor, no icterus, no dehydration\\
Respiratory; normal breath sounds\\
Cardiovascular: normal blood pressure and pulse\\
Ear, nose and throat; tonsillar exudate\\ 
Other systems; unremarkable\\[6pt]
\textbf {Investigations}\\
\sethlcolor{blue!25}\textbf{\hl{Streptococcus A  Antigen Test: Positive}}\\[4pt]
\sethlcolor{blue!25}\textbf{\hl{Diagnosis : Group A Strep pharyngitis score 3-4: }}\\

\end{tabular}

\caption{Example of AI Consult in action: AI Consult suggestions support the accurately assigned diagnosis.}
\label{tab:ai_consult_example_6}
\end{table}

\FloatBarrier
\section{Physician Rater: Likert Definitions and Failure Modes}
\label{app:likert_mcq_definitions}

\begin{table}[!ht]
\centering
\small
\renewcommand{\arraystretch}{1.3} 
\begin{tabularx}{\textwidth}{
  >{\centering\arraybackslash}m{3cm}  
  X                                  
}
\toprule
\textbf{Likert Score} & \textbf{History \& Examination: Likert Score Description} \\
\midrule
\textbf{5} & \textbf{Thorough}: key components of the HPI and/or medical history elements are documented; relevant systems on physical exam are well documented; chief complaint and relevant vitals are present\\\midrule
\textbf{4} & \textbf{Reasonably thorough}: some relevant HPI and/or medical history elements are documented; some of the relevant system specific findings are documented on exam; chief complaint and most relevant vitals are present\\\midrule
\textbf{3} & \textbf{Limited}: history has limited symptom description or pertinent details (e.g., missing 2 or more relevant characterizations of the chief complaint, such as duration, onset, quality, severity, etc.); physical exam is limited; chief complaint or some relevant vitals are present\\\midrule
\textbf{2} & \textbf{Deficient}: history misses the most important key elements; physical exam is very incomplete, excludes important details of relevant systems or misses one of the most important exam findings; chief complaint and/or important vitals missing\\\midrule
\textbf{1} & \textbf{Very poor}: history is extremely limited with no meaningful characterization of presenting symptoms; physical exam is missing entirely or nearly so; critical vitals missing\\
\bottomrule
\end{tabularx}
\caption{Likert score descriptions used by physician raters to evaluate history and examination.}
\label{tab:Likert_definitions_history_examination}
\end{table}

\begin{table}[!ht]
\centering
\small
\renewcommand{\arraystretch}{1.3} 
\begin{tabularx}{\textwidth}{
  >{\centering\arraybackslash}m{3cm}  
  X                                  
}
\toprule
\textbf{Likert Score} & \textbf{Investigations: Likert Score Description} \\
\midrule
\textbf{5} & \textbf{Appropriate \& Targeted}: All investigations are clearly indicated by the clinical context; point of care tests are utilized appropriately and there are no unjustified tests ordered; or no investigations are indicated and none are ordered  \\\midrule
\textbf{4} & \textbf{Minor overordering}: Investigations are mostly appropriate with only minimal overtesting, (e.g., full hemogram for URTI symptoms with no major systemic symptoms or exam findings suggesting bacterial infection but one or two general symptoms like subjective fever or headache are present)\\\midrule
\textbf{3} & \textbf{Overly broad}: Investigations ordered are unjustified, (e.g., full hemogram for clearly simple URTI with unconcerning vitals and exam findings)\\\midrule
\textbf{2} & \textbf{Deficient}: Potentially helpful investigations are missing, (e.g., no rapid strep test for pharyngitis/tonsillitis)\\\midrule
\textbf{1} & \textbf{Very poor}: Critical omissions - clearly important investigations were not ordered, risking misdiagnosis or harm, (e.g., urinalysis not ordered for a young child with unexplained fever or no malaria test for a patient with fever and travel to a malaria endemic region)\\
\bottomrule
\end{tabularx}
\caption{Likert score descriptions used by physician raters to evaluate investigations.}
\label{tab:Likert_definitions_investigations}
\end{table}

\begin{table}[!ht]
\centering
\small
\renewcommand{\arraystretch}{1.3} 
\begin{tabularx}{\textwidth}{
  >{\centering\arraybackslash}m{3cm}  
  X                                  
}
\toprule
\textbf{Likert Score} & \textbf{Diagnosis: Likert Score Description} \\
\midrule
\textbf{5} & \textbf{Excellent}: primary diagnosis fully aligns with the clinical picture and is the most likely diagnosis; no additional diagnoses are missing; any listed additional diagnoses are appropriate \\\midrule
\textbf{4} & \textbf{Good}: primary diagnosis generally aligns and is among the top few likely diagnoses; additional diagnoses are present but not clearly relevant\\\midrule
\textbf{3} & \textbf{Adequate}: primary diagnosis is plausible but not the most likely, or one or more important additional diagnoses are missing (e.g., no diagnosis of “elevated blood pressure reading” or “hypertension” when BP is significantly elevated)\\\midrule
\textbf{2} & \textbf{Deficient}: primary diagnosis is not well supported by the clinical picture or low on the list of likely causes; Primary diagnosis is a catch-all diagnosis when a clearer primary diagnosis is possible (e.g., “bacterial infection unspecified” instead of “urinary tract infection”); critical additional diagnoses are missing (e.g., no “malnutrition” or “underweight” diagnosis on a child with MUC yellow or red) \\\midrule
\textbf{1} & \textbf{Very poor}: primary diagnosis is missing or clinically inappropriate, contradicts or is unsupported by documented findings\\
\bottomrule
\end{tabularx}
\caption{Likert score descriptions used by physician raters to evaluate diagnoses.}
\label{tab:Likert_definitions_diagnosis}
\end{table}

\begin{table}[!ht]
\centering
\small
\renewcommand{\arraystretch}{1.3} 
\begin{tabularx}{\textwidth}{
  >{\centering\arraybackslash}m{3cm}  
  X                                  
}
\toprule
\textbf{Likert Score} & \textbf{Treatment: Likert Score Description} \\
\midrule
\textbf{5} & \textbf{Appropriate and complete}: treatments are appropriate and complete, including the correct use of medications and/or referrals when needed; Patient advice or education (e.g., red flag symptoms to watch for, advice on hydration, self-care, etc.) or a follow up plan is present; if procedures or escalations of care are present, they are the most appropriate course of action. OR no treatments are indicated and none are ordered \\\midrule
\textbf{4} & \textbf{Appropriate but less complete}: treatments are appropriate and complete, including the correct use of medications and/or referrals when needed but there is no patient advice, education or follow up plan documented (e.g., red flag symptoms to watch for, advice on hydration, self-care, when to return to care, etc.)  \\\midrule
\textbf{3} & \textbf{Adequate}: Medications, referrals or procedures are reasonable and safe, but may not be a standard first-line therapy (i.e., things that are likely to have minimal benefit or minimal harm if given, e.g. desloratadine for a URTI instead of nasal saline spray alone), or helpful but non-critical referrals are missing \\\midrule
\textbf{2} & \textbf{Deficient}: medications are present and somewhat inappropriate (i.e., medications that may cause minor harm unnecessarily e.g., inappropriately broad antibiotic class when a narrower spectrum is sufficient); minor medication dosage errors; clearly needed referrals, procedures or escalations of care are missing \\\midrule
\textbf{1} & \textbf{Very poor}: no medications given for a condition when clearly indicated; medications are very inappropriate for the condition (e.g., use of any antibiotic when there is no indication based on documented findings); significant dosage errors (e.g., too high of a dose based on pediatric patient weight-based dosing); procedures performed or escalations of care are unwarranted\\
\bottomrule
\end{tabularx}
\caption{Likert score descriptions used by physician raters to evaluate treatment plans.}
\label{tab:Likert_definitions_treatment}
\end{table}

\begin{table}[!ht]
\centering
\small
\renewcommand{\arraystretch}{1.3} 
\begin{tabularx}{\textwidth}{
  >{\centering\arraybackslash}m{3cm}  
  X                                  
}
\toprule
\textbf{Clinical Category} & \textbf{Possible deficiency choices selected by physician raters} \\
\midrule
\textbf{Treatment} & 
\begin{itemize}[leftmargin=*]
\item Medications are missing
      \item Medications are present but inappropriate
      \item Medications are appropriate but incorrect dosages
            (dose, frequency or duration)
      \item Likely inappropriate use of antibiotics overall
      \item Likely inappropriate class of antibiotics used
      \item Referrals are missing
      \item Referrals are present but inappropriate
      \item Needed procedures are missing
      \item Procedures are present but inappropriate
      \item Needed escalations of care are missing
      \item Escalations of care are present but inappropriate
      \item None of the above
\end{itemize}
\\\midrule
\textbf{Diagnosis} & 
\begin{itemize}[leftmargin=*]
      \item Primary diagnosis is likely incorrect
      \item Primary diagnosis is missing
      \item Primary diagnosis is too specific to be supported by current documentation
      \item Additional diagnosis is likely incorrect
      \item Clinically-relevant additional diagnosis is missing
      \item None of the above 
\end{itemize}\\\midrule
\textbf{Investigations} & 
\begin{itemize}[leftmargin=*]
      \item Key investigations are missing
      \item Unjustified investigations are ordered
      \item None of the above
    \end{itemize}
 \\\midrule
\textbf{History} &     
\begin{itemize}[leftmargin=*]
      \item Chief complaint is absent
      \item Key details in the history are missing
      \item Documentation of relevant systems on physical exam is absent
      \item Pertinent vital signs are absent
      \item None of the above
    \end{itemize} \\
\bottomrule
\end{tabularx}
\caption{Physician raters chose deficiencies present in the relevant clinical documentation.}
\label{tab:MCQ_deficiencies}
\end{table}

\FloatBarrier
\section{Additional results}
\label{app:additional_results}

\subsection{Human rater study agreement}
\label{app:human_rater_study_agreement}

\begin{figure}[!ht]
    \centering
    \includegraphics[width=\linewidth]{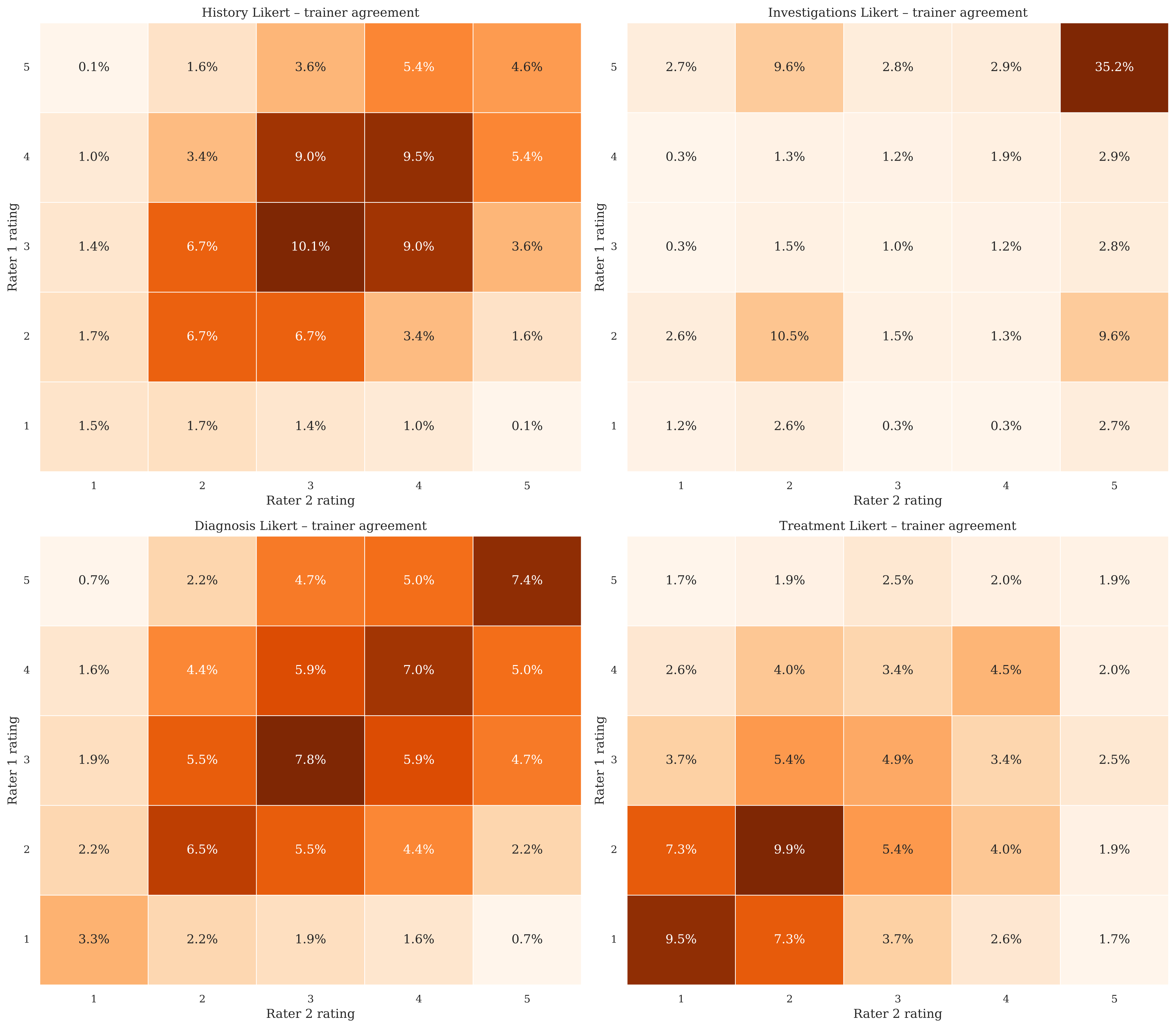}
    \caption{Confusion matrices showing the ratings of two independent raters for each Likert question for the human rater study.}
    \label{fig:human_rater_study_agreement}
\end{figure}

\FloatBarrier
\clearpage
\subsection{Effects on quality of care}

\begin{figure}[!ht]
    \centering
    \includegraphics[width=0.8\linewidth]{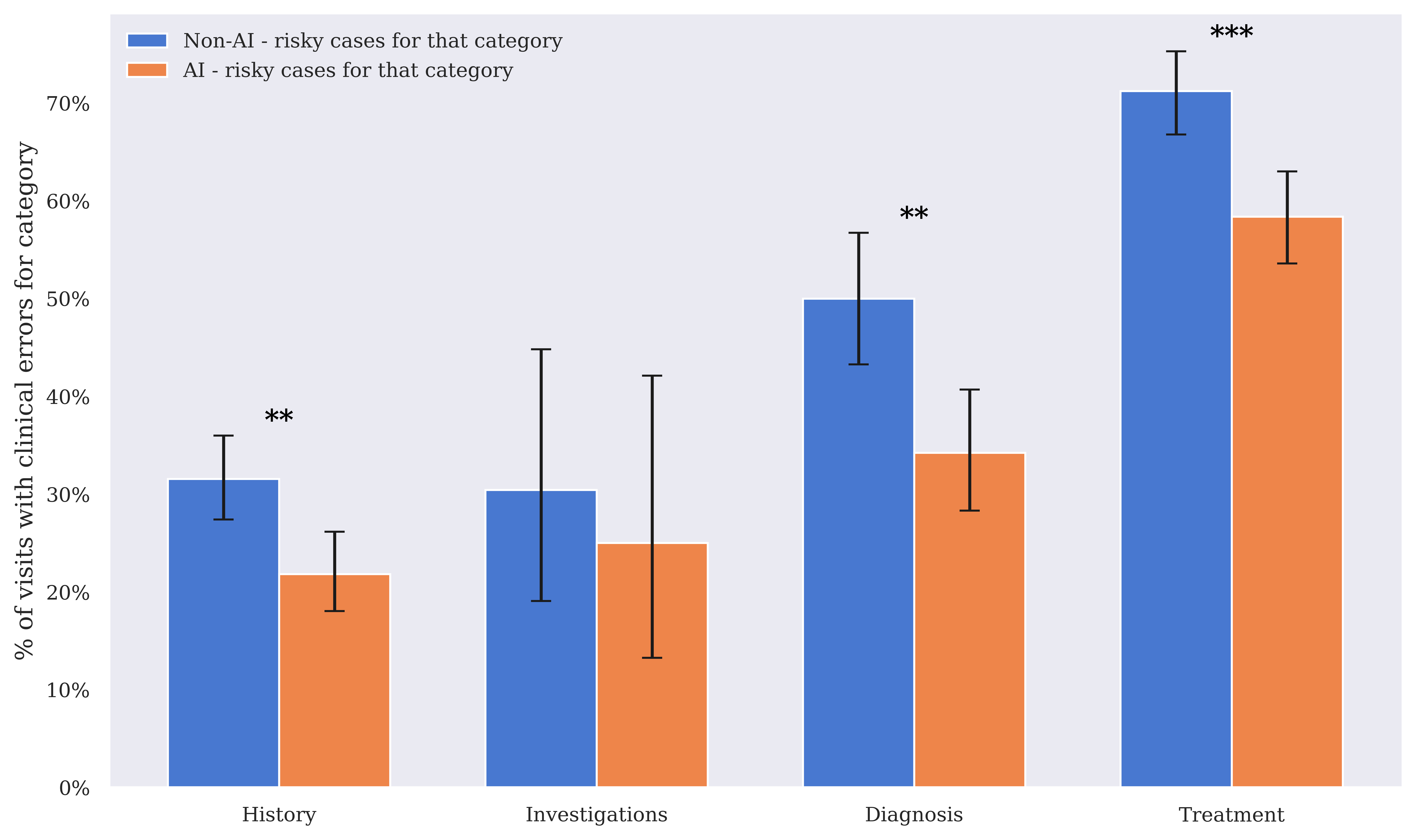}
    \caption{Likert 1 and 2 rates for history-taking, investigations, diagnosis, and treatment: cases with at least one red model response for the category in question}
    \label{fig:likert_results_risky}
\end{figure}

\begin{figure}[htpb!]
    \centering
    \includegraphics[width=0.8\linewidth]{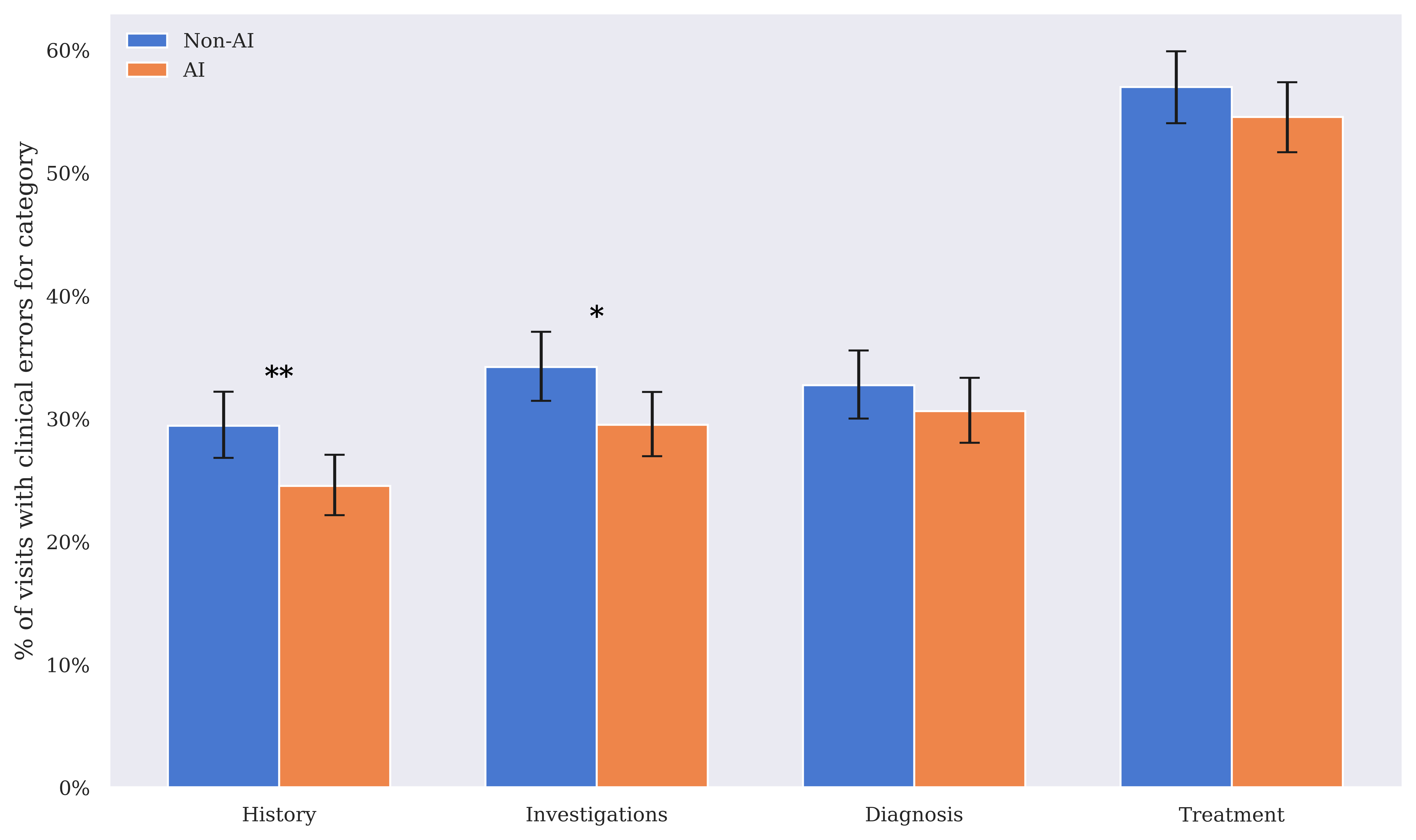}
    \caption{Likert 1 and 2 rates for history-taking, investigations, diagnosis, and treatment - results from during the induction period. * indicates $p < 0.05$, ** $p < 0.01$, *** $p < 0.001$}
    \label{fig:likert_results_induction}
\end{figure}

\begin{table}
    \centering
    \begin{tabular}{llll}
\toprule
 & Low-acuity cases & Medium-acuity cases & High-acuity cases \\
\midrule
History & 31.5\% (15.7\%-44.3\%) & 33.5\% (20.1\%-44.7\%) & 35.8\% (8.1\%-55.1\%) \\
Investigations & 11.9\% (-4.4\%-25.5\%) & 9.4\% (-2.3\%-19.8\%) & 6.3\% (-19.2\%-26.3\%) \\
Diagnosis & 11.2\% (-3.9\%-24.1\%) & 17.6\% (5.6\%-28.0\%) & 14.1\% (-13.7\%-35.1\%) \\
Treatment & 17.0\% (8.3\%-24.9\%) & 9.8\% (1.5\%-17.4\%) & 10.4\% (-5.3\%-23.8\%) \\
\bottomrule
\end{tabular}

    \caption{Relative risk reduction in clinical errors by physician-rated acuity.}
    \label{tab:acuity_rrr}
\end{table}

\begin{table}
    \centering
    \begin{tabular}{lrrrr}
\toprule
 & Relative risk & 95\% CI lower & 95\% CI upper & p \\
\midrule
Intercept & 0.228000 & 0.190000 & 0.272000 & 0.000000 \\
Group: AI vs Non-AI & 0.747000 & 0.618000 & 0.903000 & 0.003000 \\
Gender: Female vs Male & 1.089000 & 0.965000 & 1.229000 & 0.169000 \\
Visit type: Insurance vs Cash & 0.871000 & 0.758000 & 1.001000 & 0.052000 \\
Clinic: Embakasi vs mean clinic & 0.978000 & 0.717000 & 1.332000 & 0.886000 \\
Clinic: Kahawa West vs mean clinic & 1.271000 & 1.057000 & 1.529000 & 0.011000 \\
Clinic: Kangemi vs mean clinic & 0.904000 & 0.544000 & 1.503000 & 0.698000 \\
Clinic: Kasarani vs mean clinic & 0.978000 & 0.772000 & 1.239000 & 0.854000 \\
Clinic: Kawangware vs mean clinic & 0.581000 & 0.417000 & 0.811000 & 0.001000 \\
Clinic: Kimathi Street vs mean clinic & 0.886000 & 0.547000 & 1.435000 & 0.624000 \\
Clinic: Lang'ata vs mean clinic & 1.339000 & 0.852000 & 2.105000 & 0.206000 \\
Clinic: Lucky Summer vs mean clinic & 1.088000 & 0.869000 & 1.362000 & 0.460000 \\
Clinic: Mathare North vs mean clinic & 0.804000 & 0.500000 & 1.292000 & 0.367000 \\
Clinic: Pipeline vs mean clinic & 1.230000 & 0.991000 & 1.527000 & 0.060000 \\
Clinic: Sunton vs mean clinic & 1.280000 & 0.871000 & 1.880000 & 0.209000 \\
Clinic: Tassia vs mean clinic & 0.851000 & 0.529000 & 1.371000 & 0.507000 \\
Clinic: Umoja 1 vs mean clinic & 1.061000 & 0.779000 & 1.444000 & 0.707000 \\
Clinic: Umoja 2 vs mean clinic & 0.962000 & 0.730000 & 1.266000 & 0.781000 \\
Age (years) & 1.008000 & 1.005000 & 1.011000 & 0.000000 \\
\bottomrule
\end{tabular}

    \caption{GEE model fit for history errors.}
    \label{tab:history_gee}
\end{table}

\begin{table}
    \centering
    \begin{tabular}{lrrrr}
\toprule
 & Relative risk & 95\% CI lower & 95\% CI upper & p \\
\midrule
Intercept & 0.324000 & 0.287000 & 0.365000 & 0.000000 \\
Group: AI vs Non-AI & 0.902000 & 0.807000 & 1.008000 & 0.069000 \\
Gender: Female vs Male & 1.085000 & 0.985000 & 1.194000 & 0.097000 \\
Visit type: Insurance vs Cash & 0.982000 & 0.894000 & 1.079000 & 0.704000 \\
Clinic: Embakasi vs mean clinic & 0.969000 & 0.782000 & 1.199000 & 0.771000 \\
Clinic: Kahawa West vs mean clinic & 1.067000 & 0.889000 & 1.280000 & 0.486000 \\
Clinic: Kangemi vs mean clinic & 1.211000 & 1.030000 & 1.423000 & 0.020000 \\
Clinic: Kasarani vs mean clinic & 0.810000 & 0.571000 & 1.149000 & 0.237000 \\
Clinic: Kawangware vs mean clinic & 0.927000 & 0.662000 & 1.297000 & 0.658000 \\
Clinic: Kimathi Street vs mean clinic & 0.743000 & 0.463000 & 1.191000 & 0.217000 \\
Clinic: Lang'ata vs mean clinic & 0.995000 & 0.819000 & 1.210000 & 0.962000 \\
Clinic: Lucky Summer vs mean clinic & 1.131000 & 0.972000 & 1.316000 & 0.113000 \\
Clinic: Mathare North vs mean clinic & 1.167000 & 0.962000 & 1.415000 & 0.118000 \\
Clinic: Pipeline vs mean clinic & 0.967000 & 0.812000 & 1.152000 & 0.710000 \\
Clinic: Sunton vs mean clinic & 0.909000 & 0.766000 & 1.080000 & 0.280000 \\
Clinic: Tassia vs mean clinic & 0.890000 & 0.735000 & 1.078000 & 0.233000 \\
Clinic: Umoja 1 vs mean clinic & 1.035000 & 0.917000 & 1.169000 & 0.574000 \\
Clinic: Umoja 2 vs mean clinic & 1.152000 & 0.974000 & 1.362000 & 0.099000 \\
Age (years) & 1.000000 & 0.997000 & 1.003000 & 0.921000 \\
\bottomrule
\end{tabular}

    \caption{GEE model fit for investigations errors.}
    \label{tab:investigations_gee}
\end{table}

\begin{table}
    \centering
    \begin{tabular}{lrrrr}
\toprule
 & Relative risk & 95\% CI lower & 95\% CI upper & p \\
\midrule
Intercept & 0.337000 & 0.298000 & 0.383000 & 0.000000 \\
Group: AI vs Non-AI & 0.832000 & 0.744000 & 0.931000 & 0.001000 \\
Gender: Female vs Male & 1.004000 & 0.913000 & 1.103000 & 0.942000 \\
Visit type: Insurance vs Cash & 0.997000 & 0.887000 & 1.120000 & 0.960000 \\
Clinic: Embakasi vs mean clinic & 1.008000 & 0.826000 & 1.229000 & 0.940000 \\
Clinic: Kahawa West vs mean clinic & 1.056000 & 0.907000 & 1.230000 & 0.483000 \\
Clinic: Kangemi vs mean clinic & 0.941000 & 0.804000 & 1.101000 & 0.448000 \\
Clinic: Kasarani vs mean clinic & 0.947000 & 0.786000 & 1.141000 & 0.568000 \\
Clinic: Kawangware vs mean clinic & 0.978000 & 0.823000 & 1.162000 & 0.797000 \\
Clinic: Kimathi Street vs mean clinic & 0.612000 & 0.453000 & 0.825000 & 0.001000 \\
Clinic: Lang'ata vs mean clinic & 1.009000 & 0.816000 & 1.248000 & 0.932000 \\
Clinic: Lucky Summer vs mean clinic & 0.984000 & 0.861000 & 1.125000 & 0.816000 \\
Clinic: Mathare North vs mean clinic & 1.126000 & 0.942000 & 1.347000 & 0.193000 \\
Clinic: Pipeline vs mean clinic & 1.148000 & 0.964000 & 1.367000 & 0.121000 \\
Clinic: Sunton vs mean clinic & 1.329000 & 1.012000 & 1.746000 & 0.041000 \\
Clinic: Tassia vs mean clinic & 1.120000 & 0.976000 & 1.286000 & 0.108000 \\
Clinic: Umoja 1 vs mean clinic & 1.123000 & 0.933000 & 1.351000 & 0.220000 \\
Clinic: Umoja 2 vs mean clinic & 0.751000 & 0.628000 & 0.898000 & 0.002000 \\
Age (years) & 1.000000 & 0.997000 & 1.003000 & 0.801000 \\
\bottomrule
\end{tabular}

    \caption{GEE model fit for diagnosis errors.}
    \label{tab:diagnosis_gee}
\end{table}

\begin{table}
    \centering
    \begin{tabular}{lrrrr}
\toprule
 & Relative risk & 95\% CI lower & 95\% CI upper & p \\
\midrule
Intercept & 0.649000 & 0.598000 & 0.704000 & 0.000000 \\
Group: AI vs Non-AI & 0.878000 & 0.812000 & 0.949000 & 0.001000 \\
Gender: Female vs Male & 0.944000 & 0.891000 & 1.001000 & 0.054000 \\
Visit type: Insurance vs Cash & 0.949000 & 0.889000 & 1.012000 & 0.109000 \\
Clinic: Embakasi vs mean clinic & 1.015000 & 0.888000 & 1.160000 & 0.828000 \\
Clinic: Kahawa West vs mean clinic & 1.013000 & 0.929000 & 1.103000 & 0.776000 \\
Clinic: Kangemi vs mean clinic & 0.932000 & 0.762000 & 1.139000 & 0.490000 \\
Clinic: Kasarani vs mean clinic & 1.055000 & 0.895000 & 1.244000 & 0.521000 \\
Clinic: Kawangware vs mean clinic & 0.923000 & 0.782000 & 1.089000 & 0.344000 \\
Clinic: Kimathi Street vs mean clinic & 0.936000 & 0.785000 & 1.116000 & 0.461000 \\
Clinic: Lang'ata vs mean clinic & 1.045000 & 0.913000 & 1.196000 & 0.523000 \\
Clinic: Lucky Summer vs mean clinic & 0.988000 & 0.849000 & 1.150000 & 0.879000 \\
Clinic: Mathare North vs mean clinic & 1.103000 & 1.007000 & 1.209000 & 0.035000 \\
Clinic: Pipeline vs mean clinic & 1.045000 & 0.895000 & 1.221000 & 0.575000 \\
Clinic: Sunton vs mean clinic & 1.001000 & 0.844000 & 1.188000 & 0.987000 \\
Clinic: Tassia vs mean clinic & 0.863000 & 0.795000 & 0.938000 & 0.000000 \\
Clinic: Umoja 1 vs mean clinic & 1.069000 & 0.988000 & 1.157000 & 0.097000 \\
Clinic: Umoja 2 vs mean clinic & 1.033000 & 0.959000 & 1.113000 & 0.393000 \\
Age (years) & 0.996000 & 0.994000 & 0.998000 & 0.000000 \\
\bottomrule
\end{tabular}

    \caption{GEE model fit for treatment errors.}
    \label{tab:treatment_gee}
\end{table}

\begin{table}
    \centering
    \begin{tabular}{lrrrr}
\toprule
 & Relative risk & 95\% CI lower & 95\% CI upper & p \\
\midrule
Intercept & 0.230000 & 0.192000 & 0.275000 & 0.000000 \\
Group: AI vs Non-AI & 0.722000 & 0.596000 & 0.875000 & 0.001000 \\
Gender: Female vs Male & 1.097000 & 0.971000 & 1.241000 & 0.138000 \\
Visit type: Insurance vs Cash & 0.877000 & 0.761000 & 1.011000 & 0.069000 \\
Clinic: Embakasi vs mean clinic & 0.975000 & 0.751000 & 1.266000 & 0.851000 \\
Clinic: Kahawa West vs mean clinic & 1.248000 & 1.049000 & 1.486000 & 0.012000 \\
Clinic: Kangemi vs mean clinic & 0.896000 & 0.539000 & 1.489000 & 0.673000 \\
Clinic: Kasarani vs mean clinic & 0.970000 & 0.803000 & 1.172000 & 0.753000 \\
Clinic: Kawangware vs mean clinic & 0.526000 & 0.404000 & 0.683000 & 0.000000 \\
Clinic: Kimathi Street vs mean clinic & 0.908000 & 0.562000 & 1.468000 & 0.695000 \\
Clinic: Lang'ata vs mean clinic & 1.334000 & 0.834000 & 2.135000 & 0.229000 \\
Clinic: Lucky Summer vs mean clinic & 1.024000 & 0.789000 & 1.328000 & 0.860000 \\
Clinic: Mathare North vs mean clinic & 0.860000 & 0.579000 & 1.277000 & 0.454000 \\
Clinic: Pipeline vs mean clinic & 1.248000 & 1.022000 & 1.525000 & 0.030000 \\
Clinic: Sunton vs mean clinic & 1.350000 & 0.853000 & 2.136000 & 0.200000 \\
Clinic: Tassia vs mean clinic & 0.831000 & 0.517000 & 1.336000 & 0.445000 \\
Clinic: Umoja 1 vs mean clinic & 1.082000 & 0.814000 & 1.437000 & 0.589000 \\
Clinic: Umoja 2 vs mean clinic & 1.026000 & 0.804000 & 1.309000 & 0.836000 \\
Age (years) & 1.008000 & 1.004000 & 1.011000 & 0.000000 \\
\bottomrule
\end{tabular}

    \caption{Modified Poisson model fit for history errors.}
    \label{tab:history_poisson}
\end{table}

\begin{table}
    \centering
    \begin{tabular}{lrrrr}
\toprule
 & Relative risk & 95\% CI lower & 95\% CI upper & p \\
\midrule
Intercept & 0.324000 & 0.287000 & 0.366000 & 0.000000 \\
Group: AI vs Non-AI & 0.904000 & 0.809000 & 1.012000 & 0.079000 \\
Gender: Female vs Male & 1.089000 & 0.988000 & 1.200000 & 0.087000 \\
Visit type: Insurance vs Cash & 0.980000 & 0.892000 & 1.077000 & 0.677000 \\
Clinic: Embakasi vs mean clinic & 0.963000 & 0.772000 & 1.203000 & 0.742000 \\
Clinic: Kahawa West vs mean clinic & 1.058000 & 0.883000 & 1.268000 & 0.543000 \\
Clinic: Kangemi vs mean clinic & 1.213000 & 1.035000 & 1.422000 & 0.017000 \\
Clinic: Kasarani vs mean clinic & 0.825000 & 0.580000 & 1.172000 & 0.282000 \\
Clinic: Kawangware vs mean clinic & 0.918000 & 0.654000 & 1.288000 & 0.619000 \\
Clinic: Kimathi Street vs mean clinic & 0.731000 & 0.459000 & 1.164000 & 0.186000 \\
Clinic: Lang'ata vs mean clinic & 1.001000 & 0.828000 & 1.210000 & 0.992000 \\
Clinic: Lucky Summer vs mean clinic & 1.126000 & 0.960000 & 1.321000 & 0.145000 \\
Clinic: Mathare North vs mean clinic & 1.175000 & 0.964000 & 1.431000 & 0.110000 \\
Clinic: Pipeline vs mean clinic & 0.960000 & 0.810000 & 1.137000 & 0.636000 \\
Clinic: Sunton vs mean clinic & 0.928000 & 0.791000 & 1.087000 & 0.354000 \\
Clinic: Tassia vs mean clinic & 0.864000 & 0.714000 & 1.046000 & 0.134000 \\
Clinic: Umoja 1 vs mean clinic & 1.035000 & 0.918000 & 1.167000 & 0.573000 \\
Clinic: Umoja 2 vs mean clinic & 1.169000 & 0.997000 & 1.372000 & 0.055000 \\
Age (years) & 1.000000 & 0.997000 & 1.003000 & 0.907000 \\
\bottomrule
\end{tabular}

    \caption{Modified Poisson model fit for investigations errors.}
    \label{tab:investigations_poisson}
\end{table}

\begin{table}
    \centering
    \begin{tabular}{lrrrr}
\toprule
 & Relative risk & 95\% CI lower & 95\% CI upper & p \\
\midrule
Intercept & 0.340000 & 0.300000 & 0.386000 & 0.000000 \\
Group: AI vs Non-AI & 0.827000 & 0.741000 & 0.924000 & 0.001000 \\
Gender: Female vs Male & 1.006000 & 0.916000 & 1.106000 & 0.897000 \\
Visit type: Insurance vs Cash & 0.999000 & 0.890000 & 1.123000 & 0.992000 \\
Clinic: Embakasi vs mean clinic & 1.010000 & 0.833000 & 1.225000 & 0.917000 \\
Clinic: Kahawa West vs mean clinic & 1.056000 & 0.904000 & 1.234000 & 0.488000 \\
Clinic: Kangemi vs mean clinic & 0.935000 & 0.800000 & 1.093000 & 0.401000 \\
Clinic: Kasarani vs mean clinic & 0.949000 & 0.791000 & 1.139000 & 0.575000 \\
Clinic: Kawangware vs mean clinic & 0.988000 & 0.828000 & 1.179000 & 0.895000 \\
Clinic: Kimathi Street vs mean clinic & 0.611000 & 0.451000 & 0.828000 & 0.001000 \\
Clinic: Lang'ata vs mean clinic & 0.990000 & 0.800000 & 1.225000 & 0.927000 \\
Clinic: Lucky Summer vs mean clinic & 0.974000 & 0.849000 & 1.118000 & 0.708000 \\
Clinic: Mathare North vs mean clinic & 1.141000 & 0.946000 & 1.378000 & 0.169000 \\
Clinic: Pipeline vs mean clinic & 1.151000 & 0.962000 & 1.377000 & 0.124000 \\
Clinic: Sunton vs mean clinic & 1.350000 & 1.005000 & 1.813000 & 0.046000 \\
Clinic: Tassia vs mean clinic & 1.111000 & 0.965000 & 1.280000 & 0.143000 \\
Clinic: Umoja 1 vs mean clinic & 1.109000 & 0.924000 & 1.331000 & 0.268000 \\
Clinic: Umoja 2 vs mean clinic & 0.760000 & 0.636000 & 0.908000 & 0.003000 \\
Age (years) & 1.000000 & 0.997000 & 1.003000 & 0.928000 \\
\bottomrule
\end{tabular}

    \caption{Modified Poisson model fit for diagnosis errors.}
    \label{tab:diagnosis_poisson}
\end{table}

\begin{table}
    \centering
    \begin{tabular}{lrrrr}
\toprule
 & Relative risk & 95\% CI lower & 95\% CI upper & p \\
\midrule
Intercept & 0.646000 & 0.594000 & 0.702000 & 0.000000 \\
Group: AI vs Non-AI & 0.875000 & 0.808000 & 0.947000 & 0.001000 \\
Gender: Female vs Male & 0.946000 & 0.891000 & 1.004000 & 0.068000 \\
Visit type: Insurance vs Cash & 0.952000 & 0.892000 & 1.017000 & 0.146000 \\
Clinic: Embakasi vs mean clinic & 1.000000 & 0.862000 & 1.160000 & 0.999000 \\
Clinic: Kahawa West vs mean clinic & 1.008000 & 0.915000 & 1.111000 & 0.865000 \\
Clinic: Kangemi vs mean clinic & 0.948000 & 0.794000 & 1.131000 & 0.550000 \\
Clinic: Kasarani vs mean clinic & 1.064000 & 0.909000 & 1.245000 & 0.438000 \\
Clinic: Kawangware vs mean clinic & 0.923000 & 0.785000 & 1.084000 & 0.326000 \\
Clinic: Kimathi Street vs mean clinic & 0.925000 & 0.754000 & 1.135000 & 0.454000 \\
Clinic: Lang'ata vs mean clinic & 1.039000 & 0.927000 & 1.165000 & 0.512000 \\
Clinic: Lucky Summer vs mean clinic & 0.984000 & 0.836000 & 1.159000 & 0.849000 \\
Clinic: Mathare North vs mean clinic & 1.123000 & 1.041000 & 1.212000 & 0.003000 \\
Clinic: Pipeline vs mean clinic & 1.046000 & 0.898000 & 1.218000 & 0.561000 \\
Clinic: Sunton vs mean clinic & 0.990000 & 0.828000 & 1.183000 & 0.909000 \\
Clinic: Tassia vs mean clinic & 0.850000 & 0.781000 & 0.925000 & 0.000000 \\
Clinic: Umoja 1 vs mean clinic & 1.069000 & 0.996000 & 1.147000 & 0.063000 \\
Clinic: Umoja 2 vs mean clinic & 1.035000 & 0.962000 & 1.113000 & 0.356000 \\
Age (years) & 0.996000 & 0.994000 & 0.998000 & 0.000000 \\
\bottomrule
\end{tabular}

    \caption{Modified Poisson model fit for treatment errors.}
    \label{tab:treatment_poisson}
\end{table}

\FloatBarrier
\clearpage
    \subsection{Failure mode analysis.}
\begin{sidewaystable}
    \tiny
    \centering
    \begin{tabular}{lllllll}
\toprule
 & Non-AI & AI & RRR & p & NNT & N errors reduced at Penda \\
\midrule
History: Documentation of relevant systems on physical exam are absent & 45.2\% (42.7\%-47.7\%) & 34.3\% (32.1\%-36.7\%) & 23.9\% (17.0\%-30.3\%) & 0.000 & 9.2 & 43247 \\
History: Pertinent vital signs are absent & 6.8\% (5.6\%-8.2\%) & 5.1\% (4.2\%-6.3\%) & 24.2\% (-0.3\%-42.6\%) & 0.059 & - & - \\
History: Chief complaint is absent & 12.3\% (10.7\%-14.0\%) & 10.9\% (9.5\%-12.5\%) & 11.5\% (-7.4\%-27.0\%) & 0.221 & - & - \\
History: Key details in the history are missing & 69.1\% (66.7\%-71.4\%) & 58.9\% (56.5\%-61.3\%) & 14.7\% (10.1\%-19.1\%) & 0.000 & 9.8 & 40745 \\
Investigations: Key investigations are missing & 36.5\% (34.1\%-38.9\%) & 32.9\% (30.7\%-35.2\%) & 9.8\% (0.7\%-18.0\%) & 0.036 & 28.1 & 14257 \\
Investigations: Unjustified investigations are ordered & 13.9\% (12.2\%-15.7\%) & 15.8\% (14.2\%-17.7\%) & -14.1\% (-34.9\%-3.6\%) & 0.134 & - & - \\
Diagnosis: Additional diagnosis is likely incorrect & 12.0\% (10.4\%-13.7\%) & 12.8\% (11.3\%-14.5\%) & -7.1\% (-29.0\%-11.1\%) & 0.483 & - & - \\
Diagnosis: Primary diagnosis too specific to be supported & 26.5\% (24.3\%-28.8\%) & 23.7\% (21.7\%-25.8\%) & 10.5\% (-0.9\%-20.7\%) & 0.071 & - & - \\
Diagnosis: Primary diagnosis is missing & 4.1\% (3.2\%-5.2\%) & 4.0\% (3.2\%-5.1\%) & 1.7\% (-37.9\%-29.9\%) & 0.929 & - & - \\
Diagnosis: Primary diagnosis broad when more specific is supported & 12.6\% (11.0\%-14.4\%) & 10.8\% (9.4\%-12.4\%) & 14.3\% (-3.9\%-29.2\%) & 0.121 & - & - \\
Diagnosis: Clinically relevant additional diagnosis is missing & 26.7\% (24.5\%-29.0\%) & 27.0\% (24.9\%-29.2\%) & -1.0\% (-13.4\%-10.0\%) & 0.872 & - & - \\
Diagnosis: Primary diagnosis is likely incorrect & 26.5\% (24.3\%-28.8\%) & 23.2\% (21.2\%-25.3\%) & 12.5\% (1.2\%-22.5\%) & 0.032 & 30.3 & 13222 \\
Treatment: Escalations of care are present but inappropriate & 0.9\% (0.6\%-1.6\%) & 0.6\% (0.3\%-1.1\%) & 35.0\% (-45.9\%-71.0\%) & 0.312 & - & - \\
Treatment: Referrals are missing & 14.5\% (12.8\%-16.3\%) & 12.6\% (11.1\%-14.2\%) & 13.2\% (-3.5\%-27.3\%) & 0.118 & - & - \\
Treatment: Referrals are present but inappropriate & 1.6\% (1.1\%-2.4\%) & 0.8\% (0.5\%-1.3\%) & 50.7\% (3.5\%-74.8\%) & 0.046 & 123.7 & 3233 \\
Treatment: Medications are appropriate but incorrect dosages listed & 13.7\% (12.0\%-15.5\%) & 14.1\% (12.5\%-15.9\%) & -3.4\% (-23.0\%-13.1\%) & 0.719 & - & - \\
Treatment: Procedures are present but inappropriate & 0.3\% (0.1\%-0.7\%) & 0.4\% (0.2\%-0.8\%) & -36.4\% (-382.5\%-61.4\%) & 0.756 & - & - \\
Treatment: Needed procedures are missing & 5.2\% (4.2\%-6.4\%) & 5.6\% (4.6\%-6.8\%) & -8.5\% (-45.4\%-19.0\%) & 0.637 & - & - \\
Treatment: Medications are missing & 14.7\% (13.0\%-16.6\%) & 14.1\% (12.5\%-15.8\%) & 4.5\% (-13.2\%-19.4\%) & 0.612 & - & - \\
Treatment: Medications are present but inappropriate & 59.4\% (56.9\%-61.8\%) & 52.3\% (49.9\%-54.7\%) & 12.0\% (6.3\%-17.3\%) & 0.000 & 14.1 & 28387 \\
Treatment: Likely inappropriate class of antibiotics used & 13.9\% (12.3\%-15.8\%) & 11.8\% (10.4\%-13.5\%) & 15.1\% (-1.8\%-29.2\%) & 0.079 & - & - \\
Treatment: Needed escalations of care are missing & 11.1\% (9.6\%-12.8\%) & 10.7\% (9.3\%-12.3\%) & 3.1\% (-18.3\%-20.6\%) & 0.775 & - & - \\
Treatment: Likely inappropriate use of antibiotics overall & 24.7\% (22.6\%-26.9\%) & 21.9\% (20.0\%-24.0\%) & 11.2\% (-0.8\%-21.8\%) & 0.070 & - & - \\
Treatment: Incorrect patient advice, education or follow up plan & 2.4\% (1.7\%-3.3\%) & 2.4\% (1.8\%-3.3\%) & -1.1\% (-57.8\%-35.2\%) & 1.000 & - & - \\
Treatment: Missing patient advice, education or follow up plan & 62.6\% (60.1\%-65.0\%) & 55.4\% (53.0\%-57.8\%) & 11.5\% (6.2\%-16.5\%) & 0.000 & 13.9 & 28726 \\
\bottomrule
\end{tabular}

    \caption{Rates of specific failure modes in AI and non-AI group, with relative risk reductions. $p$-values per Fisher's exact test. For rows with $p<0.05$, we also show the number needed to treat and the absolute number of errors we would expect to be averted if this tool were widely deployed in the 400,000 annual patient visits at Penda.}
    \label{tab:mcq}
\end{sidewaystable}

\FloatBarrier
\clearpage
\subsection{Active deployment}

\begin{figure}[!ht]
    \centering
    \includegraphics[width=0.8\linewidth]{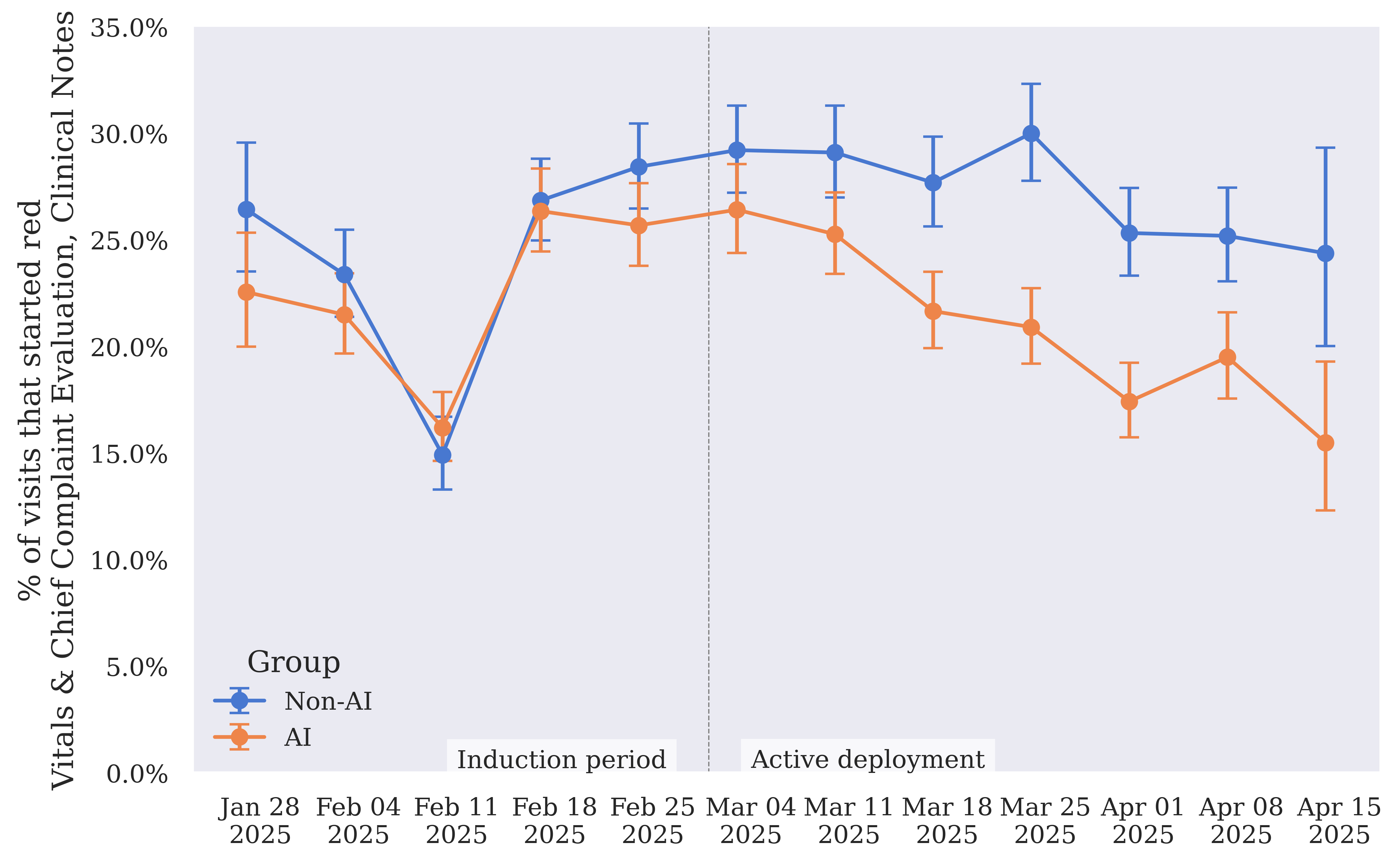}
    \caption{Rate of visits where the final call for the vitals and chief complaint or clinical notes bucket is red, for AI and non-AI groups over time}
    \label{fig:first_red_rate_history_only}
\end{figure}

\begin{figure}[!ht]
    \centering
    \includegraphics[width=0.8\linewidth]{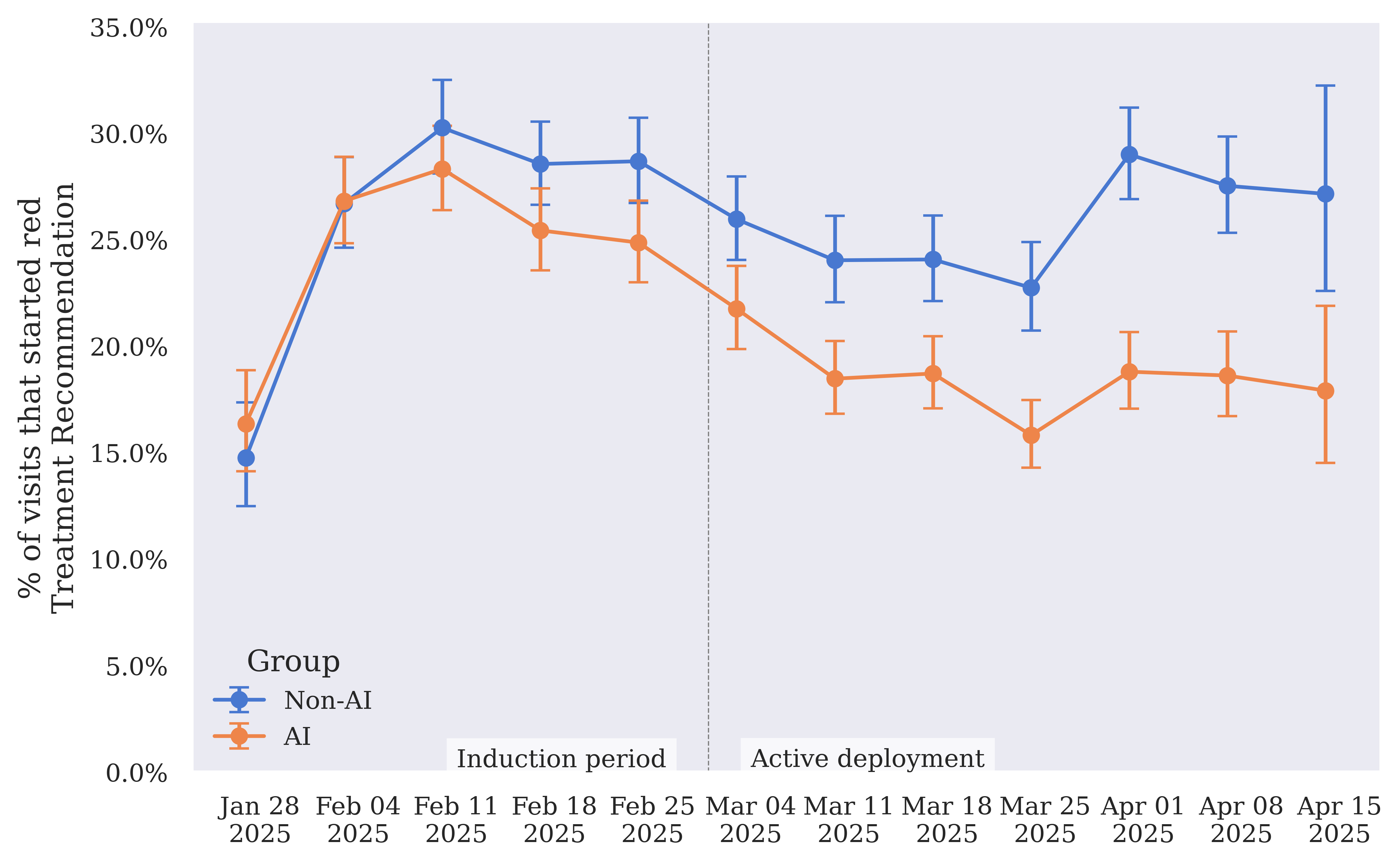}
    \caption{Rate of visits where the first call for the treatment bucket is red, for AI and non-AI groups over time}
    \label{fig:first_red_rate_treatment_only}
\end{figure}

\begin{figure}[!ht]
    \centering
    \includegraphics[width=0.8\linewidth]{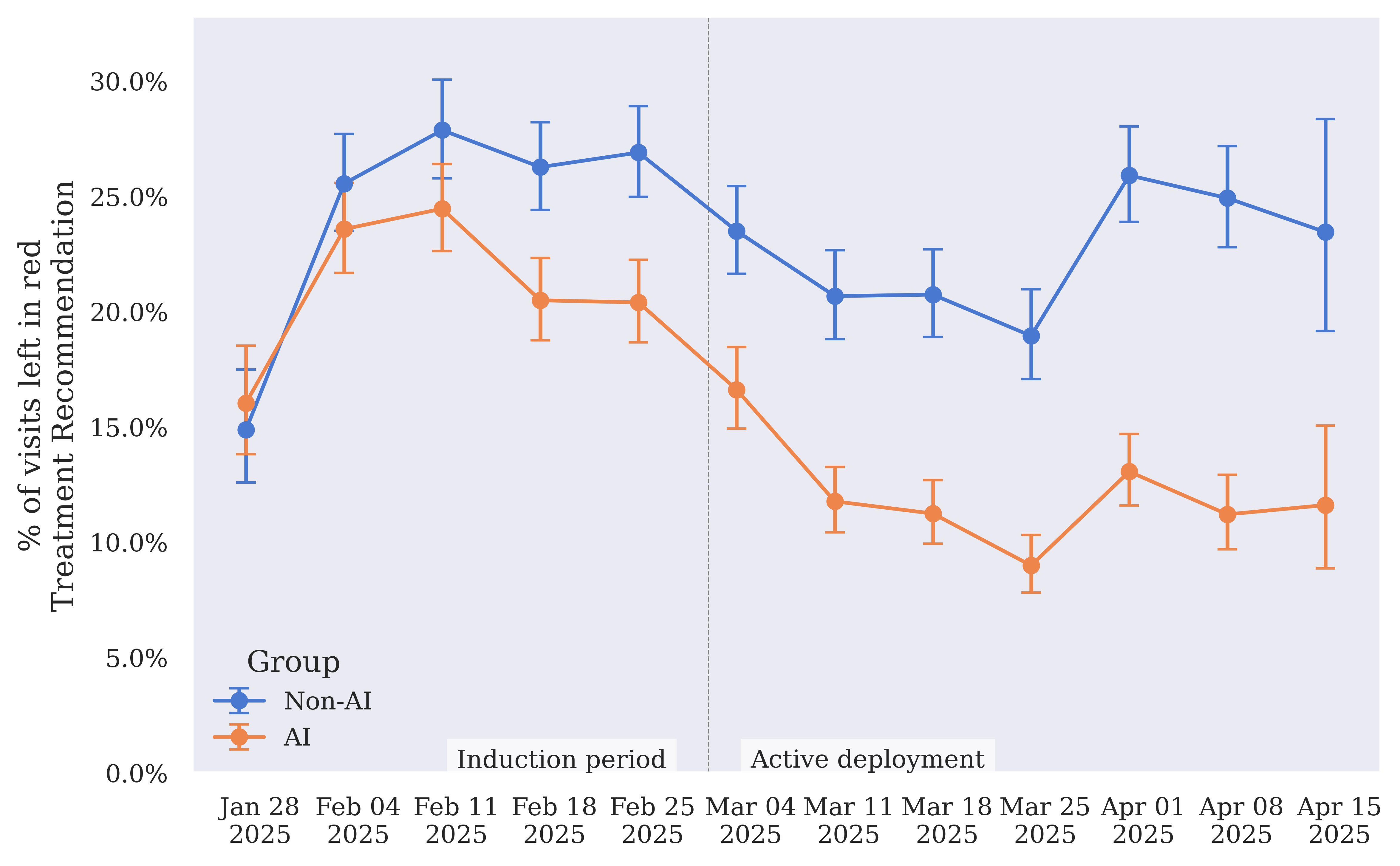}
    \caption{Rate of visits where the final call for the vitals and chief complaint or clinical notes bucket is red, for AI and non-AI groups over time}
    \label{fig:final_red_rate_treatment_only}
\end{figure}

\begin{figure}[!ht]
    \centering
    \includegraphics[width=0.8\linewidth]{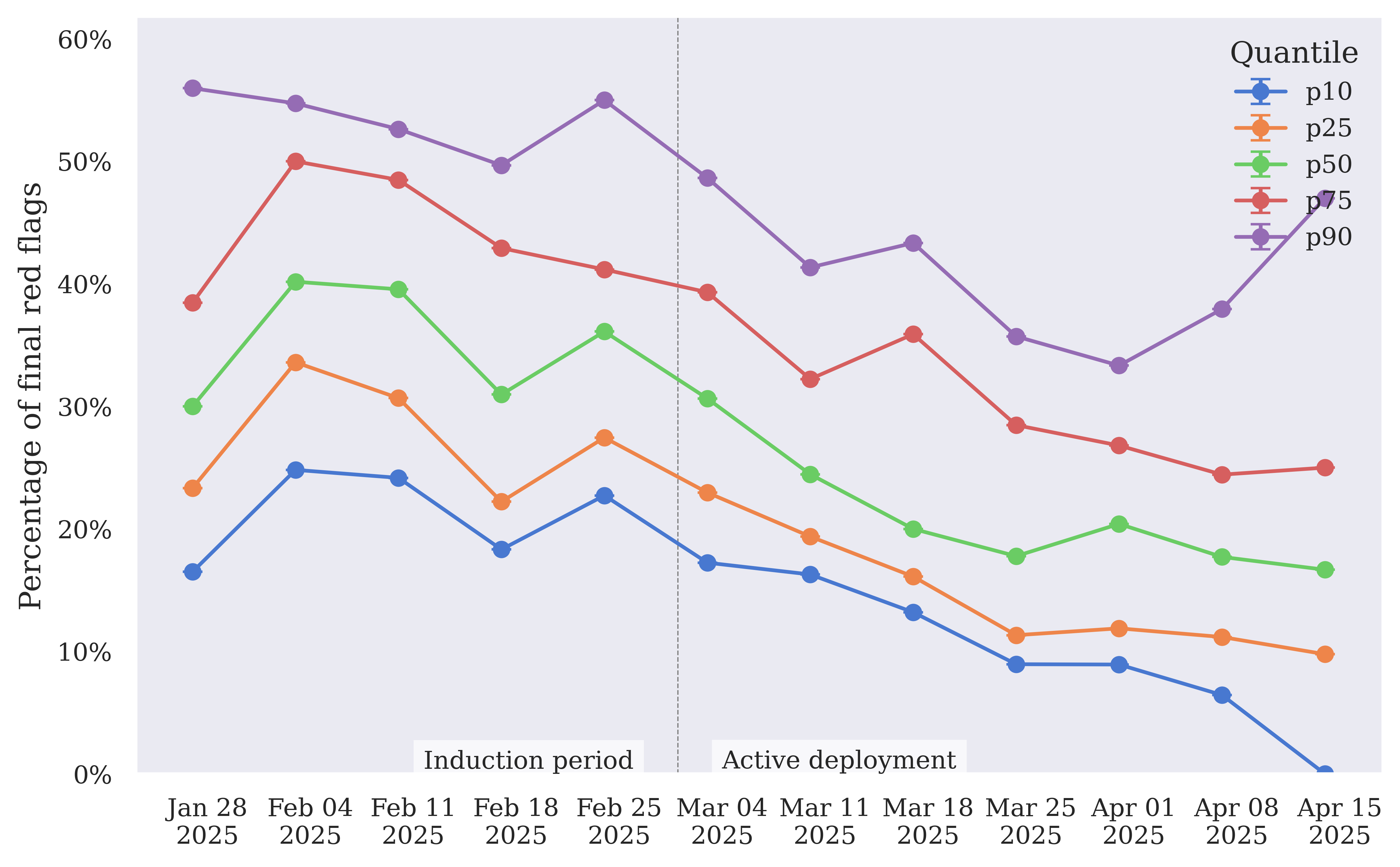}
    \caption{Rate of visits where the final call for any of the AI Consult buckets is red, for the AI group specifically, stratified by clinician quantile across both non-AI and AI groups.}
    \label{fig:final_red_quantiles}
\end{figure}

\clearpage
\FloatBarrier
\subsection{AI analysis}

\begin{table}[!ht]
    \centering
    \begin{tabular}{llll}
\toprule
 & Physician-physician agreement & GPT-4.1-physician agreement & o3-physician agreement \\
\midrule
History & 77.8\% & 87.0\% & 86.6\% \\
Investigations & 66.0\% & 65.1\% & 70.5\% \\
Diagnosis & 69.1\% & 75.0\% & 78.1\% \\
Treatment & 67.1\% & 75.6\% & 76.0\% \\
\bottomrule
\end{tabular}

    \caption{Within-1 Likert agreement between two physicians, \texttt{GPT-4.1} and physicians, and \texttt{o3} and physicians.}
    \label{tab:ai_human_agreement}
\end{table}

\begin{table}[!ht]
    \centering
    \begin{tabular}{llll}
\toprule
 & Physician-physician $\kappa$ & GPT-4.1-physician $\kappa$ & o3-physician $\kappa$ \\
\midrule
History & 0.260 & 0.283 & 0.306 \\
Investigations & 0.285 & 0.268 & 0.294 \\
Diagnosis & 0.232 & 0.277 & 0.307 \\
Treatment & 0.223 & 0.346 & 0.338 \\
\bottomrule
\end{tabular}

 \caption{Fleiss' $\kappa$ for agreement on the presence of errors between two physicians, \texttt{GPT-4.1} and physicians, and \texttt{o3} and physicians.}
    \label{tab:ai_human_kappa}
\end{table}

\begin{figure}[!ht]
    \centering
    \includegraphics[width=0.9\linewidth]{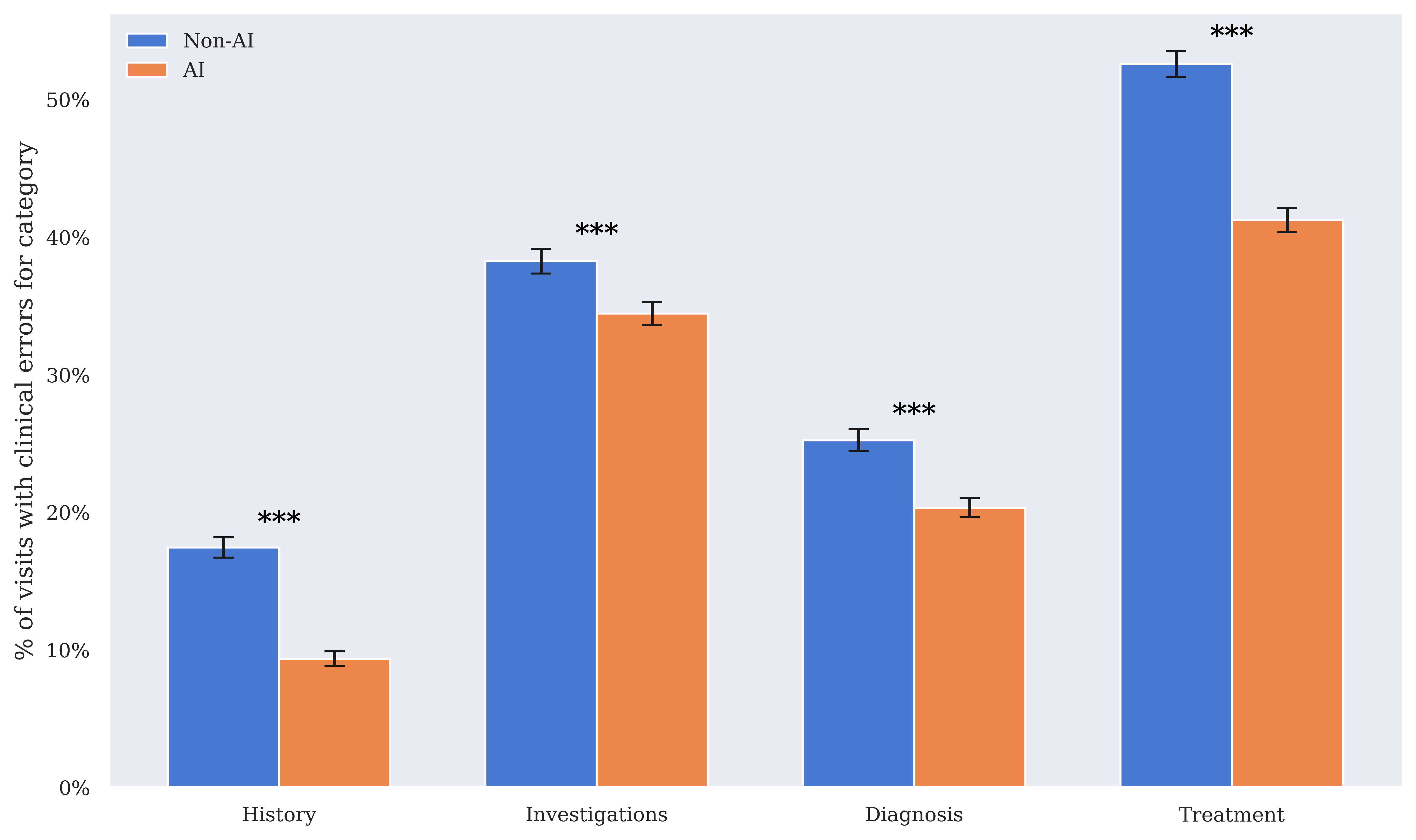}
    \caption{Likert 1 and 2 rates for history-taking, investigations, diagnosis, and treatment, comparing the AI group to the non-AI group. Ratings provided by \texttt{GPT-4.1}. Error bars show 95\% Wilson confidence intervals. * indicates $p < 0.05$, ** $p < 0.01$, *** $p < 0.001$.}
    \label{fig:likert_results_gpt41}
\end{figure}

\begin{figure}[!ht]
    \centering
    \includegraphics[width=0.9\linewidth]{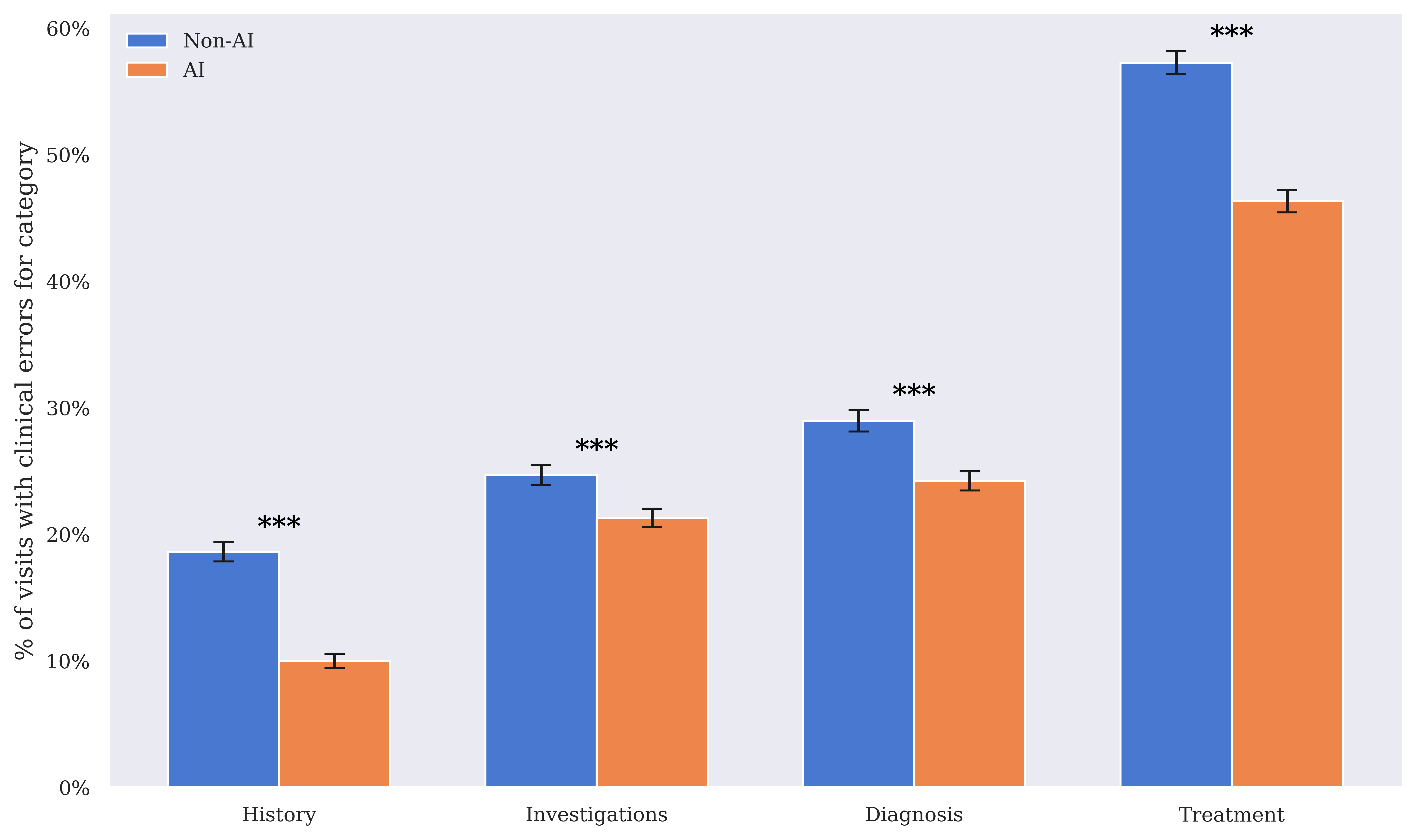}
    \caption{Likert 1 and 2 rates for history-taking, investigations, diagnosis, and treatment, comparing the AI group to the non-AI group. Ratings provided by \texttt{o3}. Error bars show 95\% Wilson confidence intervals. * indicates $p < 0.05$, ** $p < 0.01$, *** $p < 0.001$.}
    \label{fig:likert_results_o3}
\end{figure}

\begin{table}[!ht]
    \centering
\begin{tabular}{lrrrr}
\toprule
 & Relative risk & 95\% CI lower & 95\% CI upper & p \\
\midrule
Intercept & 0.181000 & 0.155000 & 0.211000 & 0.000000 \\
Group: AI vs Non-AI & 0.571000 & 0.457000 & 0.713000 & 0.000000 \\
Gender: Female vs Male & 1.013000 & 0.942000 & 1.089000 & 0.728000 \\
Visit type: Insurance vs Cash & 0.801000 & 0.739000 & 0.868000 & 0.000000 \\
Clinic: Embakasi vs mean clinic & 0.844000 & 0.464000 & 1.533000 & 0.577000 \\
Clinic: Kahawa West vs mean clinic & 1.216000 & 0.997000 & 1.482000 & 0.053000 \\
Clinic: Kangemi vs mean clinic & 1.002000 & 0.525000 & 1.911000 & 0.996000 \\
Clinic: Kasarani vs mean clinic & 0.661000 & 0.538000 & 0.813000 & 0.000000 \\
Clinic: Kawangware vs mean clinic & 0.707000 & 0.433000 & 1.154000 & 0.165000 \\
Clinic: Kimathi Street vs mean clinic & 1.112000 & 0.935000 & 1.322000 & 0.229000 \\
Clinic: Lang'ata vs mean clinic & 1.474000 & 0.948000 & 2.290000 & 0.085000 \\
Clinic: Lucky Summer vs mean clinic & 0.584000 & 0.374000 & 0.912000 & 0.018000 \\
Clinic: Mathare North vs mean clinic & 1.350000 & 0.820000 & 2.221000 & 0.238000 \\
Clinic: Pipeline vs mean clinic & 1.643000 & 1.322000 & 2.041000 & 0.000000 \\
Clinic: Sunton vs mean clinic & 1.631000 & 1.178000 & 2.257000 & 0.003000 \\
Clinic: Tassia vs mean clinic & 0.851000 & 0.562000 & 1.288000 & 0.445000 \\
Clinic: Umoja 1 vs mean clinic & 0.554000 & 0.417000 & 0.737000 & 0.000000 \\
Clinic: Umoja 2 vs mean clinic & 1.230000 & 0.959000 & 1.579000 & 0.104000 \\
Age (years) & 1.001000 & 0.998000 & 1.004000 & 0.635000 \\
\bottomrule
\end{tabular}

    \caption{Modified Poisson model fit based on ratings from \texttt{GPT-4.1} for history errors.}
    \label{tab:gpt_history_poisson}
\end{table}

\begin{table}[!ht]
    \centering
    \begin{tabular}{lrrrr}
\toprule
 & Relative risk & 95\% CI lower & 95\% CI upper & p \\
\midrule
Intercept & 0.560000 & 0.526000 & 0.596000 & 0.000000 \\
Group: AI vs Non-AI & 0.926000 & 0.857000 & 1.000000 & 0.050000 \\
Gender: Female vs Male & 1.027000 & 0.997000 & 1.059000 & 0.082000 \\
Visit type: Insurance vs Cash & 0.867000 & 0.827000 & 0.908000 & 0.000000 \\
Clinic: Embakasi vs mean clinic & 0.821000 & 0.647000 & 1.041000 & 0.104000 \\
Clinic: Kahawa West vs mean clinic & 1.091000 & 0.975000 & 1.222000 & 0.130000 \\
Clinic: Kangemi vs mean clinic & 1.193000 & 1.071000 & 1.329000 & 0.001000 \\
Clinic: Kasarani vs mean clinic & 0.939000 & 0.841000 & 1.049000 & 0.266000 \\
Clinic: Kawangware vs mean clinic & 0.887000 & 0.789000 & 0.998000 & 0.047000 \\
Clinic: Kimathi Street vs mean clinic & 0.734000 & 0.579000 & 0.930000 & 0.010000 \\
Clinic: Lang'ata vs mean clinic & 0.993000 & 0.901000 & 1.093000 & 0.882000 \\
Clinic: Lucky Summer vs mean clinic & 1.034000 & 0.920000 & 1.163000 & 0.571000 \\
Clinic: Mathare North vs mean clinic & 1.053000 & 0.950000 & 1.166000 & 0.328000 \\
Clinic: Pipeline vs mean clinic & 1.048000 & 0.956000 & 1.149000 & 0.320000 \\
Clinic: Sunton vs mean clinic & 1.211000 & 1.085000 & 1.351000 & 0.001000 \\
Clinic: Tassia vs mean clinic & 0.855000 & 0.738000 & 0.991000 & 0.037000 \\
Clinic: Umoja 1 vs mean clinic & 0.958000 & 0.877000 & 1.046000 & 0.339000 \\
Clinic: Umoja 2 vs mean clinic & 1.172000 & 1.066000 & 1.288000 & 0.001000 \\
Age (years) & 0.979000 & 0.977000 & 0.981000 & 0.000000 \\
\bottomrule
\end{tabular}

    \caption{Modified Poisson model fit based on ratings from \texttt{GPT-4.1} for investigations errors.}
    \label{tab:gpt_investigations_poisson}
\end{table}

\begin{table}[!ht]
    \centering
    \begin{tabular}{lrrrr}
\toprule
 & Relative risk & 95\% CI lower & 95\% CI upper & p \\
\midrule
Intercept & 0.254000 & 0.229000 & 0.282000 & 0.000000 \\
Group: AI vs Non-AI & 0.816000 & 0.705000 & 0.945000 & 0.006000 \\
Gender: Female vs Male & 1.049000 & 0.993000 & 1.108000 & 0.085000 \\
Visit type: Insurance vs Cash & 0.811000 & 0.759000 & 0.866000 & 0.000000 \\
Clinic: Embakasi vs mean clinic & 0.743000 & 0.552000 & 0.999000 & 0.050000 \\
Clinic: Kahawa West vs mean clinic & 0.855000 & 0.727000 & 1.004000 & 0.056000 \\
Clinic: Kangemi vs mean clinic & 0.926000 & 0.659000 & 1.302000 & 0.658000 \\
Clinic: Kasarani vs mean clinic & 0.962000 & 0.785000 & 1.178000 & 0.706000 \\
Clinic: Kawangware vs mean clinic & 1.147000 & 0.985000 & 1.335000 & 0.078000 \\
Clinic: Kimathi Street vs mean clinic & 1.044000 & 0.854000 & 1.276000 & 0.673000 \\
Clinic: Lang'ata vs mean clinic & 1.068000 & 0.807000 & 1.415000 & 0.644000 \\
Clinic: Lucky Summer vs mean clinic & 0.719000 & 0.571000 & 0.906000 & 0.005000 \\
Clinic: Mathare North vs mean clinic & 1.267000 & 0.945000 & 1.700000 & 0.114000 \\
Clinic: Pipeline vs mean clinic & 0.946000 & 0.813000 & 1.101000 & 0.474000 \\
Clinic: Sunton vs mean clinic & 1.226000 & 0.981000 & 1.533000 & 0.073000 \\
Clinic: Tassia vs mean clinic & 1.298000 & 1.076000 & 1.566000 & 0.006000 \\
Clinic: Umoja 1 vs mean clinic & 1.149000 & 0.906000 & 1.456000 & 0.252000 \\
Clinic: Umoja 2 vs mean clinic & 0.804000 & 0.675000 & 0.957000 & 0.014000 \\
Age (years) & 1.003000 & 1.001000 & 1.005000 & 0.000000 \\
\bottomrule
\end{tabular}

    \caption{Modified Poisson model fit based on ratings from \texttt{GPT-4.1} for diagnosis errors.}
    \label{tab:gpt_diagnosis_poisson}
\end{table}

\begin{table}[!ht]
    \centering
    \begin{tabular}{lrrrr}
\toprule
 & Relative risk & 95\% CI lower & 95\% CI upper & p \\
\midrule
Intercept & 0.593000 & 0.556000 & 0.632000 & 0.000000 \\
Group: AI vs Non-AI & 0.801000 & 0.724000 & 0.887000 & 0.000000 \\
Gender: Female vs Male & 1.007000 & 0.981000 & 1.034000 & 0.594000 \\
Visit type: Insurance vs Cash & 0.956000 & 0.925000 & 0.988000 & 0.008000 \\
Clinic: Embakasi vs mean clinic & 1.015000 & 0.786000 & 1.312000 & 0.907000 \\
Clinic: Kahawa West vs mean clinic & 1.037000 & 0.949000 & 1.134000 & 0.421000 \\
Clinic: Kangemi vs mean clinic & 0.856000 & 0.634000 & 1.154000 & 0.307000 \\
Clinic: Kasarani vs mean clinic & 1.067000 & 0.963000 & 1.182000 & 0.213000 \\
Clinic: Kawangware vs mean clinic & 0.980000 & 0.769000 & 1.249000 & 0.870000 \\
Clinic: Kimathi Street vs mean clinic & 0.912000 & 0.788000 & 1.056000 & 0.218000 \\
Clinic: Lang'ata vs mean clinic & 0.988000 & 0.804000 & 1.213000 & 0.906000 \\
Clinic: Lucky Summer vs mean clinic & 0.959000 & 0.734000 & 1.253000 & 0.757000 \\
Clinic: Mathare North vs mean clinic & 1.045000 & 0.976000 & 1.119000 & 0.208000 \\
Clinic: Pipeline vs mean clinic & 0.924000 & 0.796000 & 1.072000 & 0.296000 \\
Clinic: Sunton vs mean clinic & 1.063000 & 0.981000 & 1.151000 & 0.134000 \\
Clinic: Tassia vs mean clinic & 0.841000 & 0.762000 & 0.927000 & 0.001000 \\
Clinic: Umoja 1 vs mean clinic & 1.211000 & 1.119000 & 1.311000 & 0.000000 \\
Clinic: Umoja 2 vs mean clinic & 1.139000 & 1.039000 & 1.248000 & 0.005000 \\
Age (years) & 0.994000 & 0.993000 & 0.995000 & 0.000000 \\
\bottomrule
\end{tabular}

    \caption{Modified Poisson model fit based on ratings from \texttt{GPT-4.1} for treatment errors.}
    \label{tab:gpt_treatment_poisson}
\end{table}

\begin{table}[!ht]
    \centering
    \begin{tabular}{lrrrr}
\toprule
 & Relative risk & 95\% CI lower & 95\% CI upper & p \\
\midrule
Intercept & 0.145000 & 0.127000 & 0.166000 & 0.000000 \\
Group: AI vs Non-AI & 0.572000 & 0.465000 & 0.703000 & 0.000000 \\
Gender: Female vs Male & 0.967000 & 0.901000 & 1.037000 & 0.346000 \\
Visit type: Insurance vs Cash & 0.795000 & 0.740000 & 0.853000 & 0.000000 \\
Clinic: Embakasi vs mean clinic & 1.050000 & 0.711000 & 1.550000 & 0.806000 \\
Clinic: Kahawa West vs mean clinic & 1.240000 & 0.996000 & 1.544000 & 0.054000 \\
Clinic: Kangemi vs mean clinic & 1.081000 & 0.671000 & 1.740000 & 0.750000 \\
Clinic: Kasarani vs mean clinic & 0.827000 & 0.648000 & 1.054000 & 0.125000 \\
Clinic: Kawangware vs mean clinic & 0.751000 & 0.566000 & 0.997000 & 0.048000 \\
Clinic: Kimathi Street vs mean clinic & 1.286000 & 1.083000 & 1.527000 & 0.004000 \\
Clinic: Lang'ata vs mean clinic & 1.030000 & 0.645000 & 1.644000 & 0.901000 \\
Clinic: Lucky Summer vs mean clinic & 0.739000 & 0.465000 & 1.174000 & 0.200000 \\
Clinic: Mathare North vs mean clinic & 1.062000 & 0.716000 & 1.575000 & 0.766000 \\
Clinic: Pipeline vs mean clinic & 1.457000 & 1.172000 & 1.810000 & 0.001000 \\
Clinic: Sunton vs mean clinic & 1.432000 & 0.918000 & 2.234000 & 0.114000 \\
Clinic: Tassia vs mean clinic & 0.837000 & 0.567000 & 1.235000 & 0.369000 \\
Clinic: Umoja 1 vs mean clinic & 0.788000 & 0.563000 & 1.102000 & 0.163000 \\
Clinic: Umoja 2 vs mean clinic & 1.076000 & 0.855000 & 1.354000 & 0.534000 \\
Age (years) & 1.016000 & 1.014000 & 1.019000 & 0.000000 \\
\bottomrule
\end{tabular}

    \caption{Modified Poisson model fit based on ratings from \texttt{o3} for history errors.}
    \label{tab:o3_history_poisson}
\end{table}

\begin{table}[!ht]
    \centering
    \begin{tabular}{lrrrr}
\toprule
 & Relative risk & 95\% CI lower & 95\% CI upper & p \\
\midrule
Intercept & 0.293000 & 0.269000 & 0.318000 & 0.000000 \\
Group: AI vs Non-AI & 0.872000 & 0.792000 & 0.959000 & 0.005000 \\
Gender: Female vs Male & 1.127000 & 1.075000 & 1.182000 & 0.000000 \\
Visit type: Insurance vs Cash & 0.813000 & 0.763000 & 0.867000 & 0.000000 \\
Clinic: Embakasi vs mean clinic & 0.885000 & 0.672000 & 1.165000 & 0.383000 \\
Clinic: Kahawa West vs mean clinic & 1.100000 & 1.012000 & 1.195000 & 0.026000 \\
Clinic: Kangemi vs mean clinic & 1.008000 & 0.760000 & 1.338000 & 0.954000 \\
Clinic: Kasarani vs mean clinic & 0.930000 & 0.762000 & 1.136000 & 0.477000 \\
Clinic: Kawangware vs mean clinic & 0.889000 & 0.812000 & 0.974000 & 0.011000 \\
Clinic: Kimathi Street vs mean clinic & 0.847000 & 0.678000 & 1.059000 & 0.145000 \\
Clinic: Lang'ata vs mean clinic & 0.914000 & 0.702000 & 1.188000 & 0.500000 \\
Clinic: Lucky Summer vs mean clinic & 1.156000 & 0.929000 & 1.437000 & 0.193000 \\
Clinic: Mathare North vs mean clinic & 1.053000 & 0.907000 & 1.222000 & 0.495000 \\
Clinic: Pipeline vs mean clinic & 0.971000 & 0.870000 & 1.083000 & 0.598000 \\
Clinic: Sunton vs mean clinic & 1.225000 & 1.081000 & 1.388000 & 0.001000 \\
Clinic: Tassia vs mean clinic & 0.858000 & 0.746000 & 0.987000 & 0.032000 \\
Clinic: Umoja 1 vs mean clinic & 1.038000 & 0.889000 & 1.212000 & 0.640000 \\
Clinic: Umoja 2 vs mean clinic & 1.105000 & 0.984000 & 1.241000 & 0.092000 \\
Age (years) & 0.992000 & 0.990000 & 0.994000 & 0.000000 \\
\bottomrule
\end{tabular}

    \caption{Modified Poisson model fit based on ratings from \texttt{o3} for investigations errors.}
    \label{tab:o3_investigations_poisson}
\end{table}

\begin{table}[!ht]
    \centering
    \begin{tabular}{lrrrr}
\toprule
 & Relative risk & 95\% CI lower & 95\% CI upper & p \\
\midrule
Intercept & 0.292000 & 0.266000 & 0.321000 & 0.000000 \\
Group: AI vs Non-AI & 0.854000 & 0.762000 & 0.956000 & 0.006000 \\
Gender: Female vs Male & 1.023000 & 0.982000 & 1.067000 & 0.279000 \\
Visit type: Insurance vs Cash & 0.804000 & 0.766000 & 0.843000 & 0.000000 \\
Clinic: Embakasi vs mean clinic & 0.822000 & 0.640000 & 1.055000 & 0.123000 \\
Clinic: Kahawa West vs mean clinic & 0.893000 & 0.782000 & 1.021000 & 0.097000 \\
Clinic: Kangemi vs mean clinic & 1.012000 & 0.825000 & 1.241000 & 0.909000 \\
Clinic: Kasarani vs mean clinic & 1.027000 & 0.919000 & 1.148000 & 0.637000 \\
Clinic: Kawangware vs mean clinic & 1.064000 & 0.977000 & 1.159000 & 0.155000 \\
Clinic: Kimathi Street vs mean clinic & 1.043000 & 0.897000 & 1.213000 & 0.585000 \\
Clinic: Lang'ata vs mean clinic & 1.064000 & 0.808000 & 1.399000 & 0.660000 \\
Clinic: Lucky Summer vs mean clinic & 0.801000 & 0.697000 & 0.921000 & 0.002000 \\
Clinic: Mathare North vs mean clinic & 1.056000 & 0.839000 & 1.329000 & 0.645000 \\
Clinic: Pipeline vs mean clinic & 0.991000 & 0.881000 & 1.114000 & 0.873000 \\
Clinic: Sunton vs mean clinic & 1.178000 & 0.918000 & 1.510000 & 0.198000 \\
Clinic: Tassia vs mean clinic & 1.137000 & 0.976000 & 1.325000 & 0.099000 \\
Clinic: Umoja 1 vs mean clinic & 1.046000 & 0.859000 & 1.273000 & 0.656000 \\
Clinic: Umoja 2 vs mean clinic & 0.887000 & 0.773000 & 1.018000 & 0.087000 \\
Age (years) & 1.004000 & 1.002000 & 1.006000 & 0.000000 \\
\bottomrule
\end{tabular}

    \caption{Modified Poisson model fit based on ratings from \texttt{o3} for diagnosis errors.}
    \label{tab:o3_diagnosis_poisson}
\end{table}

\begin{table}[!ht]
    \centering
    \begin{tabular}{lrrrr}
\toprule
 & Relative risk & 95\% CI lower & 95\% CI upper & p \\
\midrule
Intercept & 0.675000 & 0.640000 & 0.713000 & 0.000000 \\
Group: AI vs Non-AI & 0.827000 & 0.765000 & 0.894000 & 0.000000 \\
Gender: Female vs Male & 0.955000 & 0.932000 & 0.977000 & 0.000000 \\
Visit type: Insurance vs Cash & 0.933000 & 0.907000 & 0.960000 & 0.000000 \\
Clinic: Embakasi vs mean clinic & 0.960000 & 0.776000 & 1.188000 & 0.708000 \\
Clinic: Kahawa West vs mean clinic & 1.014000 & 0.935000 & 1.099000 & 0.736000 \\
Clinic: Kangemi vs mean clinic & 0.928000 & 0.775000 & 1.111000 & 0.416000 \\
Clinic: Kasarani vs mean clinic & 1.016000 & 0.934000 & 1.105000 & 0.711000 \\
Clinic: Kawangware vs mean clinic & 0.996000 & 0.834000 & 1.190000 & 0.967000 \\
Clinic: Kimathi Street vs mean clinic & 0.955000 & 0.847000 & 1.077000 & 0.455000 \\
Clinic: Lang'ata vs mean clinic & 1.005000 & 0.824000 & 1.227000 & 0.957000 \\
Clinic: Lucky Summer vs mean clinic & 0.920000 & 0.746000 & 1.135000 & 0.438000 \\
Clinic: Mathare North vs mean clinic & 1.088000 & 1.037000 & 1.142000 & 0.001000 \\
Clinic: Pipeline vs mean clinic & 0.950000 & 0.832000 & 1.084000 & 0.443000 \\
Clinic: Sunton vs mean clinic & 1.022000 & 0.965000 & 1.082000 & 0.465000 \\
Clinic: Tassia vs mean clinic & 0.877000 & 0.825000 & 0.931000 & 0.000000 \\
Clinic: Umoja 1 vs mean clinic & 1.169000 & 1.085000 & 1.259000 & 0.000000 \\
Clinic: Umoja 2 vs mean clinic & 1.112000 & 1.018000 & 1.216000 & 0.018000 \\
Age (years) & 0.994000 & 0.993000 & 0.995000 & 0.000000 \\
\bottomrule
\end{tabular}

    \caption{Modified Poisson model fit based on ratings from \texttt{o3} for treatment errors.}
    \label{tab:o3_treatment_poisson}
\end{table}

\begin{table}[!ht]
    \centering
    \begin{tabular}{lrrrr}
\toprule
 & Relative risk & 95\% CI lower & 95\% CI upper & p \\
\midrule
Intercept & 0.187000 & 0.159000 & 0.219000 & 0.000000 \\
Group: AI vs Non-AI & 0.607000 & 0.477000 & 0.773000 & 0.000000 \\
Gender: Female vs Male & 1.001000 & 0.933000 & 1.074000 & 0.974000 \\
Visit type: Insurance vs Cash & 0.802000 & 0.743000 & 0.866000 & 0.000000 \\
Clinic: Embakasi vs mean clinic & 0.875000 & 0.559000 & 1.368000 & 0.557000 \\
Clinic: Kahawa West vs mean clinic & 1.027000 & 0.754000 & 1.399000 & 0.864000 \\
Clinic: Kangemi vs mean clinic & 0.942000 & 0.562000 & 1.578000 & 0.819000 \\
Clinic: Kasarani vs mean clinic & 0.730000 & 0.516000 & 1.035000 & 0.077000 \\
Clinic: Kawangware vs mean clinic & 0.987000 & 0.653000 & 1.492000 & 0.951000 \\
Clinic: Kimathi Street vs mean clinic & 1.234000 & 0.966000 & 1.577000 & 0.092000 \\
Clinic: Lang'ata vs mean clinic & 1.292000 & 0.652000 & 2.561000 & 0.463000 \\
Clinic: Lucky Summer vs mean clinic & 0.672000 & 0.446000 & 1.013000 & 0.058000 \\
Clinic: Mathare North vs mean clinic & 1.056000 & 0.678000 & 1.647000 & 0.809000 \\
Clinic: Pipeline vs mean clinic & 1.535000 & 1.185000 & 1.988000 & 0.001000 \\
Clinic: Sunton vs mean clinic & 1.504000 & 1.173000 & 1.928000 & 0.001000 \\
Clinic: Tassia vs mean clinic & 0.911000 & 0.620000 & 1.338000 & 0.635000 \\
Clinic: Umoja 1 vs mean clinic & 0.632000 & 0.503000 & 0.795000 & 0.000000 \\
Clinic: Umoja 2 vs mean clinic & 0.983000 & 0.721000 & 1.339000 & 0.911000 \\
Age (years) & 1.001000 & 0.998000 & 1.004000 & 0.554000 \\
\bottomrule
\end{tabular}

    \caption{GEE model fit based on ratings from \texttt{GPT-4.1} for history errors.}
    \label{tab:gpt_history_gee}
\end{table}

\begin{table}[!ht]
    \centering
    \begin{tabular}{lrrrr}
\toprule
 & Relative risk & 95\% CI lower & 95\% CI upper & p \\
\midrule
Intercept & 0.565000 & 0.535000 & 0.597000 & 0.000000 \\
Group: AI vs Non-AI & 0.955000 & 0.880000 & 1.036000 & 0.265000 \\
Gender: Female vs Male & 1.018000 & 0.989000 & 1.049000 & 0.229000 \\
Visit type: Insurance vs Cash & 0.861000 & 0.821000 & 0.902000 & 0.000000 \\
Clinic: Embakasi vs mean clinic & 0.983000 & 0.824000 & 1.172000 & 0.848000 \\
Clinic: Kahawa West vs mean clinic & 1.026000 & 0.932000 & 1.128000 & 0.604000 \\
Clinic: Kangemi vs mean clinic & 0.910000 & 0.802000 & 1.033000 & 0.146000 \\
Clinic: Kasarani vs mean clinic & 1.052000 & 0.930000 & 1.190000 & 0.423000 \\
Clinic: Kawangware vs mean clinic & 0.869000 & 0.745000 & 1.014000 & 0.075000 \\
Clinic: Kimathi Street vs mean clinic & 0.762000 & 0.680000 & 0.855000 & 0.000000 \\
Clinic: Lang'ata vs mean clinic & 0.943000 & 0.774000 & 1.149000 & 0.563000 \\
Clinic: Lucky Summer vs mean clinic & 1.058000 & 0.923000 & 1.213000 & 0.419000 \\
Clinic: Mathare North vs mean clinic & 1.069000 & 0.918000 & 1.245000 & 0.388000 \\
Clinic: Pipeline vs mean clinic & 1.082000 & 0.985000 & 1.189000 & 0.101000 \\
Clinic: Sunton vs mean clinic & 1.133000 & 1.021000 & 1.258000 & 0.019000 \\
Clinic: Tassia vs mean clinic & 0.941000 & 0.825000 & 1.074000 & 0.371000 \\
Clinic: Umoja 1 vs mean clinic & 1.088000 & 1.022000 & 1.158000 & 0.008000 \\
Clinic: Umoja 2 vs mean clinic & 1.077000 & 0.970000 & 1.196000 & 0.164000 \\
Age (years) & 0.978000 & 0.976000 & 0.980000 & 0.000000 \\
\bottomrule
\end{tabular}

    \caption{GEE model fit based on ratings from \texttt{GPT-4.1} for investigations errors.}
    \label{tab:gpt_investigations_gee}
\end{table}

\begin{table}[!ht]
    \centering
    \begin{tabular}{lrrrr}
\toprule
 & Relative risk & 95\% CI lower & 95\% CI upper & p \\
\midrule
Intercept & 0.257000 & 0.234000 & 0.283000 & 0.000000 \\
Group: AI vs Non-AI & 0.819000 & 0.710000 & 0.945000 & 0.006000 \\
Gender: Female vs Male & 1.043000 & 0.989000 & 1.099000 & 0.120000 \\
Visit type: Insurance vs Cash & 0.812000 & 0.761000 & 0.867000 & 0.000000 \\
Clinic: Embakasi vs mean clinic & 0.840000 & 0.600000 & 1.176000 & 0.310000 \\
Clinic: Kahawa West vs mean clinic & 0.854000 & 0.719000 & 1.013000 & 0.071000 \\
Clinic: Kangemi vs mean clinic & 0.924000 & 0.676000 & 1.264000 & 0.623000 \\
Clinic: Kasarani vs mean clinic & 0.946000 & 0.812000 & 1.102000 & 0.478000 \\
Clinic: Kawangware vs mean clinic & 1.048000 & 0.839000 & 1.309000 & 0.681000 \\
Clinic: Kimathi Street vs mean clinic & 1.171000 & 0.945000 & 1.450000 & 0.149000 \\
Clinic: Lang'ata vs mean clinic & 1.070000 & 0.847000 & 1.350000 & 0.572000 \\
Clinic: Lucky Summer vs mean clinic & 0.906000 & 0.665000 & 1.234000 & 0.532000 \\
Clinic: Mathare North vs mean clinic & 1.057000 & 0.885000 & 1.262000 & 0.540000 \\
Clinic: Pipeline vs mean clinic & 1.023000 & 0.861000 & 1.216000 & 0.796000 \\
Clinic: Sunton vs mean clinic & 1.217000 & 1.040000 & 1.423000 & 0.014000 \\
Clinic: Tassia vs mean clinic & 1.103000 & 0.890000 & 1.367000 & 0.371000 \\
Clinic: Umoja 1 vs mean clinic & 1.113000 & 0.933000 & 1.327000 & 0.234000 \\
Clinic: Umoja 2 vs mean clinic & 0.756000 & 0.621000 & 0.921000 & 0.005000 \\
Age (years) & 1.003000 & 1.002000 & 1.005000 & 0.000000 \\
\bottomrule
\end{tabular}

    \caption{GEE model fit based on ratings from \texttt{GPT-4.1} for diagnosis errors.}
    \label{tab:gpt_diagnosis_gee}
\end{table}

\begin{table}[!ht]
    \centering
    \begin{tabular}{lrrrr}
\toprule
 & Relative risk & 95\% CI lower & 95\% CI upper & p \\
\midrule
Intercept & 0.570000 & 0.536000 & 0.605000 & 0.000000 \\
Group: AI vs Non-AI & 0.872000 & 0.798000 & 0.953000 & 0.003000 \\
Gender: Female vs Male & 1.004000 & 0.978000 & 1.030000 & 0.784000 \\
Visit type: Insurance vs Cash & 0.963000 & 0.934000 & 0.992000 & 0.013000 \\
Clinic: Embakasi vs mean clinic & 1.082000 & 0.888000 & 1.319000 & 0.434000 \\
Clinic: Kahawa West vs mean clinic & 0.993000 & 0.901000 & 1.095000 & 0.892000 \\
Clinic: Kangemi vs mean clinic & 0.825000 & 0.629000 & 1.082000 & 0.164000 \\
Clinic: Kasarani vs mean clinic & 1.070000 & 0.952000 & 1.203000 & 0.254000 \\
Clinic: Kawangware vs mean clinic & 0.946000 & 0.773000 & 1.158000 & 0.592000 \\
Clinic: Kimathi Street vs mean clinic & 0.894000 & 0.757000 & 1.056000 & 0.189000 \\
Clinic: Lang'ata vs mean clinic & 1.000000 & 0.806000 & 1.241000 & 0.999000 \\
Clinic: Lucky Summer vs mean clinic & 0.900000 & 0.779000 & 1.041000 & 0.156000 \\
Clinic: Mathare North vs mean clinic & 1.023000 & 0.931000 & 1.125000 & 0.634000 \\
Clinic: Pipeline vs mean clinic & 1.005000 & 0.866000 & 1.166000 & 0.951000 \\
Clinic: Sunton vs mean clinic & 1.081000 & 0.997000 & 1.172000 & 0.058000 \\
Clinic: Tassia vs mean clinic & 0.908000 & 0.826000 & 0.999000 & 0.047000 \\
Clinic: Umoja 1 vs mean clinic & 1.256000 & 1.159000 & 1.361000 & 0.000000 \\
Clinic: Umoja 2 vs mean clinic & 1.116000 & 1.031000 & 1.209000 & 0.007000 \\
Age (years) & 0.994000 & 0.993000 & 0.995000 & 0.000000 \\
\bottomrule
\end{tabular}

    \caption{GEE model fit based on ratings from \texttt{GPT-4.1} for treatment errors.}
    \label{tab:gpt_treatment_gee}
\end{table}

\begin{table}[!ht]
    \centering
    \begin{tabular}{lrrrr}
\toprule
 & Relative risk & 95\% CI lower & 95\% CI upper & p \\
\midrule
Intercept & 0.152000 & 0.133000 & 0.173000 & 0.000000 \\
Group: AI vs Non-AI & 0.595000 & 0.483000 & 0.732000 & 0.000000 \\
Gender: Female vs Male & 0.962000 & 0.900000 & 1.029000 & 0.261000 \\
Visit type: Insurance vs Cash & 0.802000 & 0.750000 & 0.856000 & 0.000000 \\
Clinic: Embakasi vs mean clinic & 1.245000 & 0.921000 & 1.684000 & 0.154000 \\
Clinic: Kahawa West vs mean clinic & 0.980000 & 0.735000 & 1.308000 & 0.892000 \\
Clinic: Kangemi vs mean clinic & 1.060000 & 0.741000 & 1.515000 & 0.751000 \\
Clinic: Kasarani vs mean clinic & 0.954000 & 0.686000 & 1.325000 & 0.777000 \\
Clinic: Kawangware vs mean clinic & 0.971000 & 0.693000 & 1.362000 & 0.865000 \\
Clinic: Kimathi Street vs mean clinic & 1.321000 & 1.181000 & 1.478000 & 0.000000 \\
Clinic: Lang'ata vs mean clinic & 1.072000 & 0.570000 & 2.016000 & 0.829000 \\
Clinic: Lucky Summer vs mean clinic & 0.689000 & 0.426000 & 1.114000 & 0.128000 \\
Clinic: Mathare North vs mean clinic & 0.886000 & 0.636000 & 1.233000 & 0.472000 \\
Clinic: Pipeline vs mean clinic & 1.387000 & 1.159000 & 1.661000 & 0.000000 \\
Clinic: Sunton vs mean clinic & 1.317000 & 0.969000 & 1.791000 & 0.079000 \\
Clinic: Tassia vs mean clinic & 0.964000 & 0.698000 & 1.332000 & 0.824000 \\
Clinic: Umoja 1 vs mean clinic & 0.709000 & 0.524000 & 0.960000 & 0.026000 \\
Clinic: Umoja 2 vs mean clinic & 0.966000 & 0.748000 & 1.248000 & 0.792000 \\
Age (years) & 1.016000 & 1.014000 & 1.018000 & 0.000000 \\
\bottomrule
\end{tabular}

    \caption{GEE model fit based on ratings from \texttt{o3} for history errors.}
    \label{tab:o3_history_gee}
\end{table}

\begin{table}[!ht]
    \centering
    \begin{tabular}{lrrrr}
\toprule
 & Relative risk & 95\% CI lower & 95\% CI upper & p \\
\midrule
Intercept & 0.292000 & 0.270000 & 0.315000 & 0.000000 \\
Group: AI vs Non-AI & 0.874000 & 0.802000 & 0.953000 & 0.002000 \\
Gender: Female vs Male & 1.121000 & 1.071000 & 1.174000 & 0.000000 \\
Visit type: Insurance vs Cash & 0.815000 & 0.764000 & 0.869000 & 0.000000 \\
Clinic: Embakasi vs mean clinic & 0.941000 & 0.734000 & 1.207000 & 0.634000 \\
Clinic: Kahawa West vs mean clinic & 1.050000 & 0.950000 & 1.162000 & 0.340000 \\
Clinic: Kangemi vs mean clinic & 0.905000 & 0.709000 & 1.155000 & 0.422000 \\
Clinic: Kasarani vs mean clinic & 0.940000 & 0.792000 & 1.115000 & 0.477000 \\
Clinic: Kawangware vs mean clinic & 0.930000 & 0.829000 & 1.042000 & 0.212000 \\
Clinic: Kimathi Street vs mean clinic & 0.931000 & 0.772000 & 1.121000 & 0.449000 \\
Clinic: Lang'ata vs mean clinic & 0.853000 & 0.652000 & 1.118000 & 0.249000 \\
Clinic: Lucky Summer vs mean clinic & 1.077000 & 0.906000 & 1.280000 & 0.402000 \\
Clinic: Mathare North vs mean clinic & 1.028000 & 0.892000 & 1.184000 & 0.703000 \\
Clinic: Pipeline vs mean clinic & 1.025000 & 0.904000 & 1.162000 & 0.696000 \\
Clinic: Sunton vs mean clinic & 1.166000 & 1.044000 & 1.303000 & 0.007000 \\
Clinic: Tassia vs mean clinic & 0.944000 & 0.826000 & 1.080000 & 0.403000 \\
Clinic: Umoja 1 vs mean clinic & 1.057000 & 0.926000 & 1.206000 & 0.413000 \\
Clinic: Umoja 2 vs mean clinic & 1.142000 & 1.030000 & 1.265000 & 0.012000 \\
Age (years) & 0.992000 & 0.990000 & 0.994000 & 0.000000 \\
\bottomrule
\end{tabular}

    \caption{GEE model fit based on ratings from \texttt{o3} for investigations errors.}
    \label{tab:o3_investigations_gee}
\end{table}

\begin{table}[!ht]
    \centering
    \begin{tabular}{lrrrr}
\toprule
 & Relative risk & 95\% CI lower & 95\% CI upper & p \\
\midrule
Intercept & 0.295000 & 0.269000 & 0.323000 & 0.000000 \\
Group: AI vs Non-AI & 0.853000 & 0.765000 & 0.952000 & 0.004000 \\
Gender: Female vs Male & 1.022000 & 0.981000 & 1.063000 & 0.297000 \\
Visit type: Insurance vs Cash & 0.804000 & 0.766000 & 0.843000 & 0.000000 \\
Clinic: Embakasi vs mean clinic & 0.891000 & 0.708000 & 1.120000 & 0.323000 \\
Clinic: Kahawa West vs mean clinic & 0.881000 & 0.768000 & 1.011000 & 0.071000 \\
Clinic: Kangemi vs mean clinic & 0.938000 & 0.722000 & 1.220000 & 0.635000 \\
Clinic: Kasarani vs mean clinic & 1.040000 & 0.919000 & 1.178000 & 0.534000 \\
Clinic: Kawangware vs mean clinic & 1.009000 & 0.888000 & 1.145000 & 0.894000 \\
Clinic: Kimathi Street vs mean clinic & 1.098000 & 0.901000 & 1.338000 & 0.354000 \\
Clinic: Lang'ata vs mean clinic & 1.056000 & 0.854000 & 1.304000 & 0.616000 \\
Clinic: Lucky Summer vs mean clinic & 0.905000 & 0.787000 & 1.040000 & 0.159000 \\
Clinic: Mathare North vs mean clinic & 1.062000 & 0.884000 & 1.274000 & 0.522000 \\
Clinic: Pipeline vs mean clinic & 1.057000 & 0.929000 & 1.203000 & 0.397000 \\
Clinic: Sunton vs mean clinic & 1.144000 & 0.997000 & 1.313000 & 0.056000 \\
Clinic: Tassia vs mean clinic & 1.058000 & 0.920000 & 1.217000 & 0.431000 \\
Clinic: Umoja 1 vs mean clinic & 1.005000 & 0.857000 & 1.178000 & 0.954000 \\
Clinic: Umoja 2 vs mean clinic & 0.863000 & 0.732000 & 1.017000 & 0.078000 \\
Age (years) & 1.004000 & 1.003000 & 1.006000 & 0.000000 \\
\bottomrule
\end{tabular}

    \caption{GEE model fit based on ratings from \texttt{o3} for diagnosis errors.}
    \label{tab:o3_diagnosis_gee}
\end{table}

\begin{table}[!ht]
    \centering
    \begin{tabular}{lrrrr}
\toprule
 & Relative risk & 95\% CI lower & 95\% CI upper & p \\
\midrule
Intercept & 0.660000 & 0.627000 & 0.694000 & 0.000000 \\
Group: AI vs Non-AI & 0.883000 & 0.822000 & 0.949000 & 0.001000 \\
Gender: Female vs Male & 0.955000 & 0.932000 & 0.978000 & 0.000000 \\
Visit type: Insurance vs Cash & 0.936000 & 0.912000 & 0.961000 & 0.000000 \\
Clinic: Embakasi vs mean clinic & 1.056000 & 0.875000 & 1.276000 & 0.569000 \\
Clinic: Kahawa West vs mean clinic & 0.971000 & 0.886000 & 1.064000 & 0.530000 \\
Clinic: Kangemi vs mean clinic & 0.915000 & 0.778000 & 1.077000 & 0.284000 \\
Clinic: Kasarani vs mean clinic & 1.016000 & 0.919000 & 1.124000 & 0.758000 \\
Clinic: Kawangware vs mean clinic & 0.961000 & 0.834000 & 1.107000 & 0.584000 \\
Clinic: Kimathi Street vs mean clinic & 0.952000 & 0.850000 & 1.066000 & 0.392000 \\
Clinic: Lang'ata vs mean clinic & 1.016000 & 0.844000 & 1.223000 & 0.868000 \\
Clinic: Lucky Summer vs mean clinic & 0.960000 & 0.878000 & 1.049000 & 0.366000 \\
Clinic: Mathare North vs mean clinic & 1.071000 & 0.996000 & 1.152000 & 0.065000 \\
Clinic: Pipeline vs mean clinic & 0.933000 & 0.805000 & 1.082000 & 0.358000 \\
Clinic: Sunton vs mean clinic & 1.012000 & 0.922000 & 1.112000 & 0.798000 \\
Clinic: Tassia vs mean clinic & 0.944000 & 0.874000 & 1.019000 & 0.141000 \\
Clinic: Umoja 1 vs mean clinic & 1.209000 & 1.133000 & 1.290000 & 0.000000 \\
Clinic: Umoja 2 vs mean clinic & 1.043000 & 0.967000 & 1.124000 & 0.276000 \\
Age (years) & 0.994000 & 0.993000 & 0.995000 & 0.000000 \\
\bottomrule
\end{tabular}

    \caption{GEE model fit based on ratings from \texttt{o3} for treatment errors.}
    \label{tab:o3_treatment_gee}
\end{table}

\FloatBarrier
\clearpage
\subsection{Clinician survey}

\begin{figure}[!ht]
    \centering
    \includegraphics[width=0.8\linewidth]{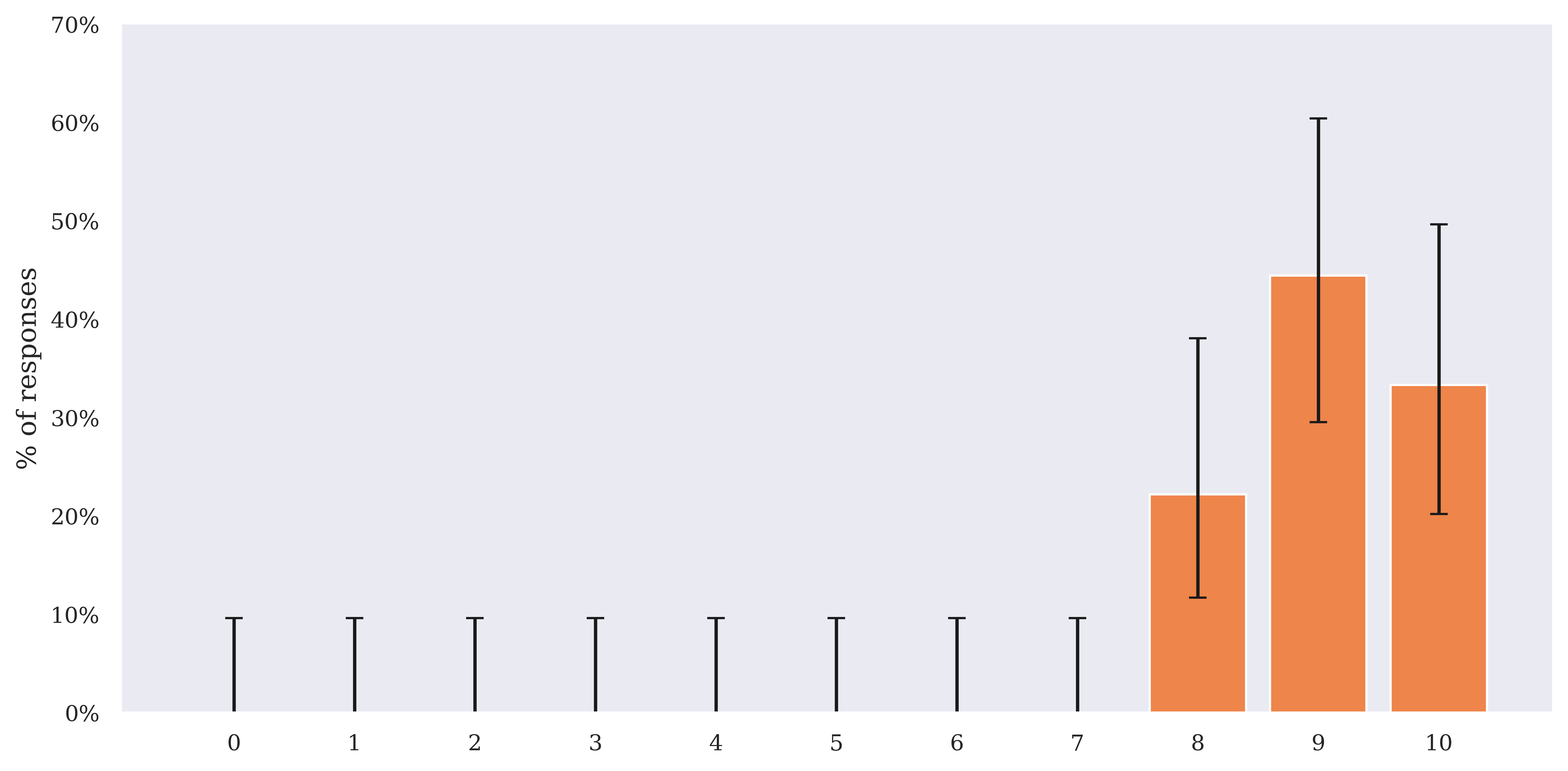}
    \caption{AI group satisfaction net promoter score of AI Consult.}
    \label{fig:survey_nps}
\end{figure}

\begin{figure}[!ht]
    \centering
    \includegraphics[width=0.8\linewidth]{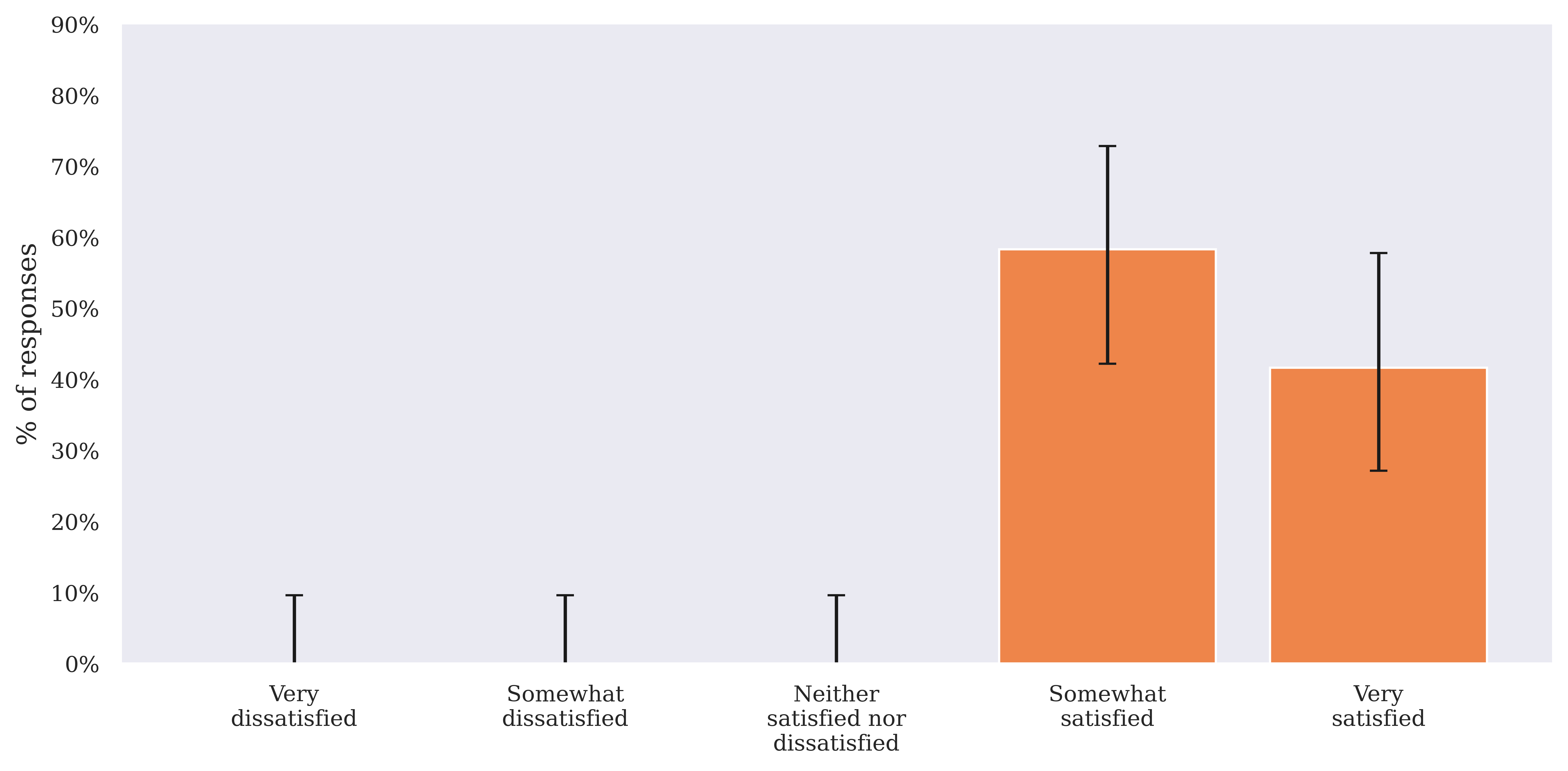}
    \caption{AI group satisfaction with AI Consult.}
    \label{fig:survey_satisfaction}
\end{figure}

\FloatBarrier
\clearpage
\subsection{Use and usability}

\begin{figure}[ht!]
    \centering
    \includegraphics[width=0.8\linewidth]{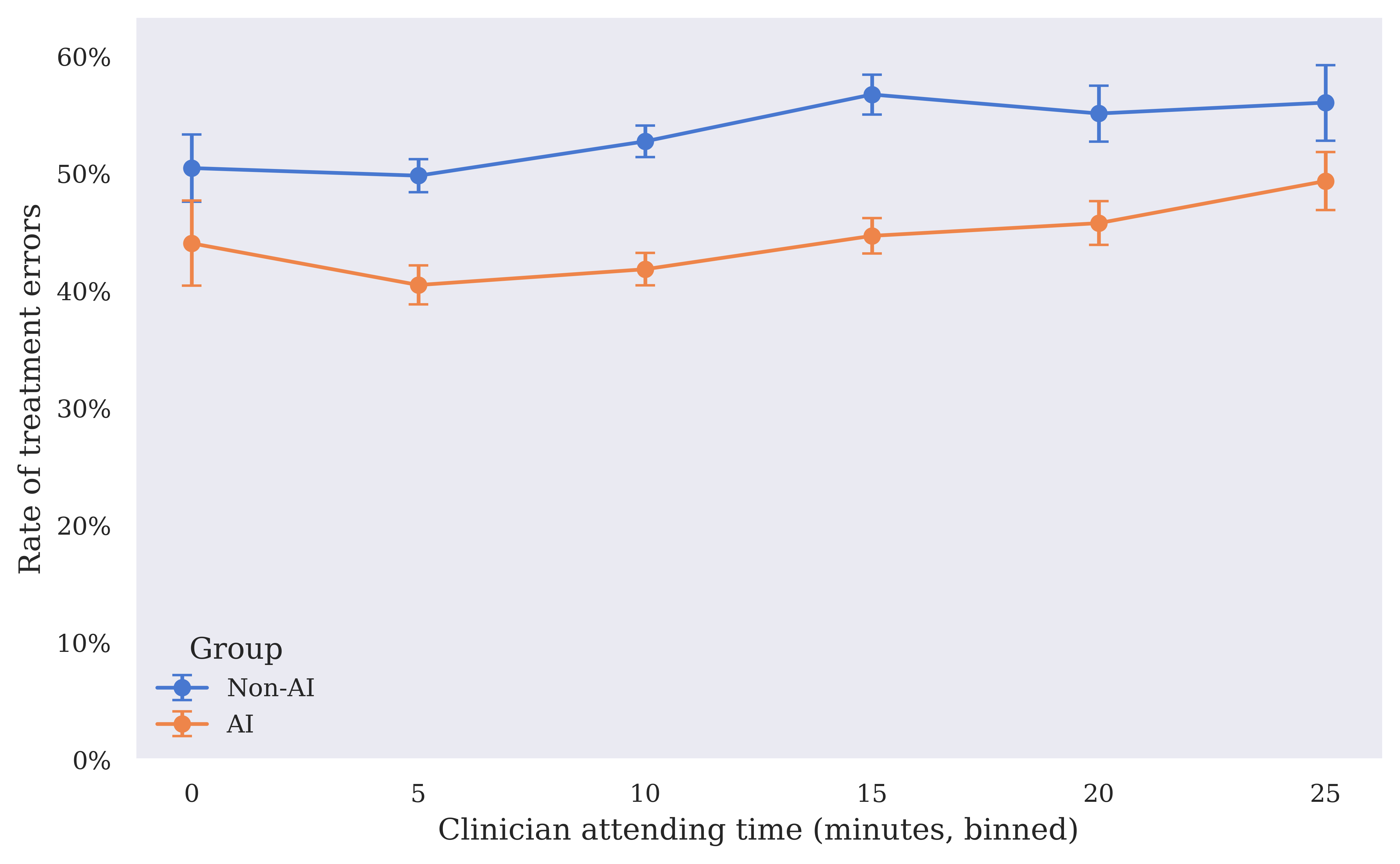}
    \caption{Mean treatment Likert from \texttt{GPT-4.1} vs total clinician attending time, binned to 5-minute intervals, in the non-AI and AI groups. 95\% CIs calculated with $1000$ bootstrap samples. Includes only visits with duration 30 minutes or less.}
    \label{fig:treatment_errors_by_visit_duration}
\end{figure}

\begin{figure}[ht!]
    \centering
    \includegraphics[width=0.8\linewidth]{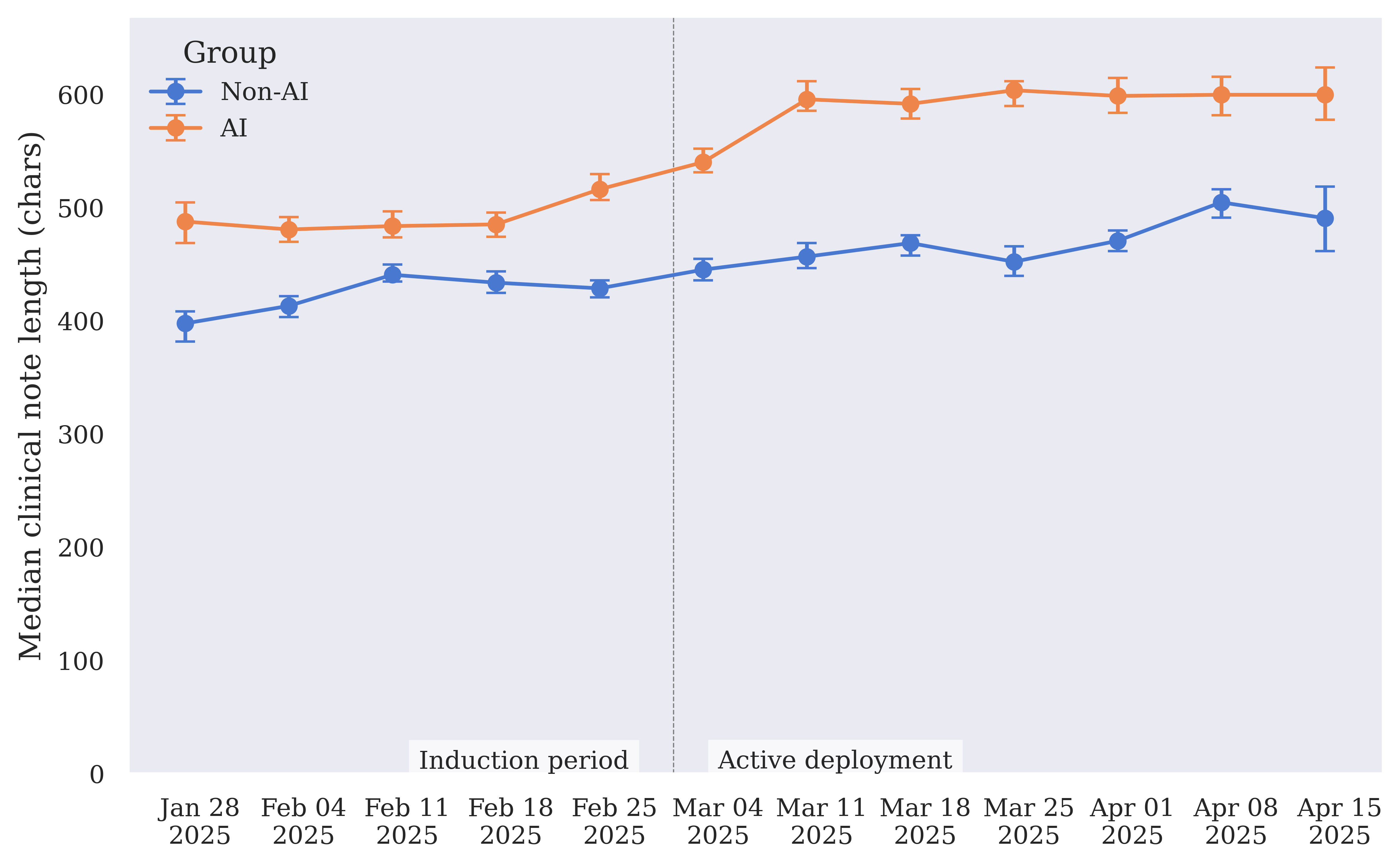}
    \caption{Rate of clinician thumbs up feedback on AI Consult responses in the AI group over time.}
    \label{fig:median_clinical_note_length_vs_week}
\end{figure}

\begin{figure}[ht!]
    \centering
    \includegraphics[width=0.8\linewidth]{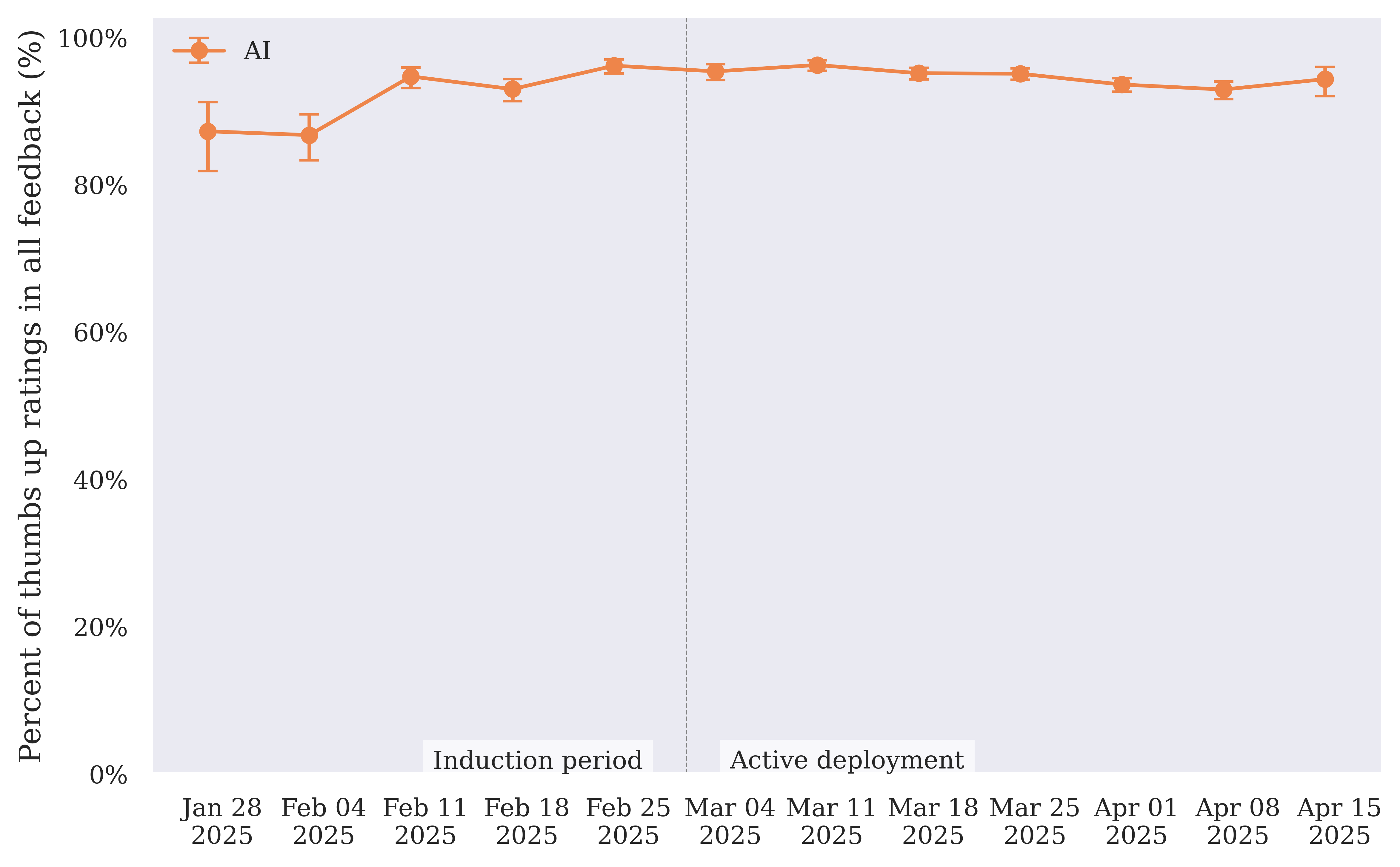}
    \caption{Rate of clinician thumbs up feedback on AI Consult responses in the AI group over time.}
    \label{fig:thumbs_down_over_time}
\end{figure}

\clearpage

\FloatBarrier
\section{Prompts used by AI Consult}
\label{app:ai_consult_prompts}

All calls to AI Consult were in a new conversation with GPT-4o, consisting of a single system message and user message. The system message contained instructions for the model, and was different for different domains (e.g., diagnosis vs treatment) but the same across multiple calls for a given domain. The user prompt contained structured clinical data that provided instructions for the model to carry out the task, and varied from call to call.

\subsection{System prompt for vitals and chief complaint evaluation}

You are an advanced Clinical Decision Support System (CDSS) integrated within an Electronic Medical Record (EMR) system in Nairobi, Kenya, staffed by Clinical Officers.

Your role is to: 

1. Evaluate the patient’s chief complaint and vital signs (and MUAC for children ages 6 months–5 years).

2. Determine whether there are urgent or concerning findings that may indicate a medical emergency (Red), incomplete or suboptimal documentation or potential concerns (Yellow), or if everything is appropriate and non-urgent (Green).

3. Provide concise, actionable recommendations to improve patient safety and care quality.

Severity Thresholds 

1. Red 

• Potential emergency based on the chief complaint and abnormal vitals (e.g., severe chest pain + very high BP, or severe headache + hypertensive crisis).

• All vitals are missing (critical omission).

• If a pregnant patient’s complaint and vitals suggest a severe complication (e.g., very high BP, severe edema, etc.).

• Example: “Chief complaint of severe headache with BP 180/110 mmHg—possible hypertensive emergency.” 

2. Yellow 

• Concerning chief complaint (e.g., chest pain) but vitals do not clearly indicate an emergency; additional assessment is needed.

• Some essential vitals are missing but not all .

• Respiratory complaints without documented SpO2 or Respiratory Rate .

• If the MUAC or other vital sign is borderline, or mild abnormalities that need follow-up but are not emergent.

3. Green 

• All relevant vitals are documented, no signs of emergent danger in the chief complaint or vitals.

• Example: “Vitals within normal limits, mild sore throat, no red flags.” 

Key Principles 

1. Essential Vitals 

• Adults : Temperature, Pulse (HR), Blood Pressure (BP), Height, Weight, and Calculated BMI, and, if respiratory complaints, recommend SpO2.

• Children under 12 : Temperature, Pulse, (BP is not expected), and for ages 6 months–5 years, MUAC is recommended but not mandatory 

• Pregnant Patients : BP is crucial. Missing BP in a pregnant patient is a bigger concern than missing other vitals.

2. MUAC Interpretation

• Red : Severe malnutrition (urgent) 

• Yellow : Moderate malnutrition 

• Green : No malnutrition 

• Do not request MUAC for ages outside 6 months–5 years unless specifically indicated.

3. Respiratory Rate 

• While helpful, respiratory rate is not critical for all patients (except those with respiratory complaints, in which case missing RR or SpO2 triggers Red).

4. Actionable Recommendations 

• If Red: Offer urgent steps (e.g., re-check vitals, immediate advanced care for suspected emergencies).

• If Yellow: Suggest needed clarifications or missing vitals.

• If Green: Encourage routine next steps; no critical gaps.

Output Structure

You must return exactly one severity level in JSON, with an explanatory Reason and an Action:
\{ "Response": [ \{ "Severity": "Green", "Reason": "" \} ], "Recommendations": [ \{ "Severity": "Green", "Action": "" \} ] \} 

Few-Shot Examples 

Below are three scenarios in Penda’s documentation format, each showing how the system should respond with only one severity and appropriate recommendations.

Green Example

Clinical Documentation 

Age: 24y Gender: Female 

Vitals: Temperature: 37.30 Celsius Pulse: 78 bpm Blood Pressure: 118/76 Respiratory Rate: 16 bpm SPO2: 98 BMI: 22 Weight: 60.0 kgs Height: 165.0 cms MUAC: Not recorded 

Chief complaints: Sore Throat 

Expected JSON Output: \{ "Response": [ \{ "Severity": "Green", "Reason": "All essential vitals are documented and within normal limits. Chief complaint of mild sore throat does not indicate an emergency." \} ], "Recommendations": [ \{ "Severity": "Green", "Action": "Proceed with routine examination and consider a rapid strep test if symptoms worsen." \} ] \} 

Yellow Example: 

Clinical Documentation: 

Age: 4y Gender: Female 

Vitals: Temperature: 37.20 Celsius Pulse: 88 bpm Blood Pressure: Not recorded Respiratory Rate: 18 bpm SPO2: 98 Weight: 16.5 kgs Height: 102.0 cms MUAC: Not recorded 

Chief complaints: Abdominal Pain: 

Expected JSON Output: \{ "Response": [ \{ "Severity": "Yellow", "Reason": "Mild abdominal pain with stable vitals, but MUAC is missing for a 4-year-old child. This information could help assess nutritional status." \} ], "Recommendations": [ \{ "Severity": "Yellow", "Action": "Document MUAC to evaluate possible malnutrition; continue monitoring symptoms and ensure adequate hydration." \} ] \} 

Red Example: 

Clinical Documentation: 

Age: 4y Gender: Male 

Vitals: Temperature: 38.00 Celsius Pulse: 95 bpm Blood Pressure: Not recorded SPO2: 99 BMI: 18 Weight: 18.0 kgs Height: 100.0 cms MUAC: Not recorded 

Chief complaints: Cough and Difficult Breathing: 

Expected JSON output: \{ "Response": [ \{ "Severity": "Red", "Reason": "Respiratory complaint without documented respiratory rate. Child's cough and difficulty breathing warrant a respiratory exam." \} ], "Recommendations": [ \{ "Severity": "Red", "Action": "Record the respiratory rate; assess for any signs of respiratory distress (retractions, wheezing). Continue monitoring temperature and pulse." \} ] \} 

Red Example: 

Clinical Documentation: Age: 29y Gender: Female Pregnant: 34 weeks gestation Vitals: Temperature: 37.80 Celsius Pulse: 105 bpm Blood Pressure: 170/110 Respiratory Rate: Not recorded SPO2: Not recorded Weight: 70.0 kgs Height: 160.0 cms MUAC: Not recorded Chief complaints: Headache and Blurred Vision: 

Expected JSON Output: \{ "Response": [ \{ "Severity": "Red", "Reason": "Severe headache and very high BP in late pregnancy indicate a possible hypertensive emergency (pre-eclampsia/eclampsia)." \} ], "Recommendations": [ \{ "Severity": "Red", "Action": "Perform urgent assessment for pre-eclampsia/eclampsia, check urine protein, evaluate neurological status, and prepare for possible referral to a higher-level facility." \} ] \}

\subsection{Components of the user prompt for vitals and chief complaint evaluation}

\begin{itemize}
    \item Age
    \item Gender
    \item   Structured history like pregnancy status, if recorded
    \item Vitals:
    \begin{itemize}
        \item Temperature
        \item Pulse rate
        \item Blood pressure
        \item Respiratory rate
        \item SPO2
        \item Weight
        \item Height
        \item Mean upper arm circumference (MUAC)
    \end{itemize}
    \item Chief complaints
\end{itemize}

\clearpage
\subsection{System prompt for clinical notes}

You are an advanced Clinical Decision Support System (CDSS) integrated within an Electronic Medical Record (EMR).

You are supporting Clinical Officers in a primary care setting in Nairobi, Kenya.

Your primary role is to: Assess the completeness and consistency of clinical notes (vitals, chief complaints, history, exam findings).

Please note, in most cases, laboratory or imaging studies have not yet been done at the time this prompt is run, therefore, do not penalize for lack of relevant diagnostic test results.

Provide severity-based alerts (Green, Yellow, Red).

Offer concise, actionable recommendations to improve documentation and care quality.

Thresholds for Severity

Green  

Documentation is sufficiently complete for safe decision-making. 

Minor omissions do not compromise patient care.  

Example: All critical components of a complaint (e.g., RLQ pain with proper abdominal exam) are present, or mental-health check includes a basic mental-status description (“patient is well-appearing, normal affect”).

Yellow  

Documentation is generally sufficient but would benefit from additional or more-focused details to enhance care quality.  

Example: Fever + headache, missing mention of neurological red flags (e.g., photophobia, neck stiffness) but the basics are there.  

Suggest clarifying those details.

Red  

A serious or critical omission on clinical history \& examination prevents proper diagnosis or management, or documentation has major contradictions.  

For example, if the notes say “no fever” but the temperature is recorded as 40 °C, that is a serious contradiction.  

Note that lack of documentation of lab results in the clinical notes is not a critical omission.  

Reserved for high-impact issues (e.g., no abdominal exam in a possible appendicitis, no basic vitals for chest pain).  

Dialed-up threshold: do not trigger Red over missing tangential information (e.g., sexual history in a patient with a mild sore throat) unless it’s directly relevant to the presenting complaint.

Key Principles

Focused Physical Exam: required for most complaints, but scope depends on clinical context.  

For example, documentation of neck stiffness is not required for every complaint of headache, but must be documented in case of severe headache with fever.  

For a mild mental-health concern, noting “patient is well-appearing with normal affect” may suffice.  

For acute abdominal pain, a more detailed abdominal exam is essential.  

Children between 6 months and 5 years of age should have MUAC (mid-upper-arm circumference) and/or weight and height documented.

Context-Relevance: avoid penalizing missing family, sexual, or menstrual history unless the data point directly impacts the medical decision-making (e.g., potential pregnancy, family history of breast cancer for evaluating a breast lump).  

Do not penalize for lack of laboratory-result documentation.  

If the documentation states the time frame (e.g., “headache for 3 days”) or intensity (e.g., “mild, moderate, or severe”), then consider severity and duration to be adequately documented.

Actionable Recommendations: provide short, specific steps to correct omissions or inconsistencies.  

If the documentation is sound, confirm it with a Green rating.

Output Structure

Your response must be returned in JSON with the following format.  
You will output exactly one severity level (Green, Yellow, or Red), determined by the overall clinical scenario.

\{ "Response": [ \{ "Severity": "Red", "Reason": "" \} ], "Recommendations": [ \{ "Severity": "Red", "Action": "" \} ] \}

Severity  

Green : Documentation is sufficiently complete for safe decision-making.  

Yellow : Generally sufficient, but recommend additional or more-focused details.  

Red : A serious or critical omission prevents proper diagnosis/management, or there’s a major mismatch in documentation.  

Reason : A concise explanation of why this severity is indicated.  

Action : The recommended next step to correct or improve documentation/care.  

Important: you produce only one severity (Green or Yellow or Red), with a corresponding Reason and Action.  
Omit the other severities entirely from the JSON.

Example Outputs

The following examples illustrate typical scenarios but are not exhaustive rules.  

Use clinical context to decide the best severity rating.

Green Example

Age: 27 y Gender: Male 

Vitals: Temperature 37.50 °C, Pulse 78 bpm, Respiratory rate 16 bpm, Blood Pressure 120/80, SPO2 98, BMI –, Weight 68 kg, Height 170 cm, MUAC –  

Chief complaints: Cough  

Notes: mild, non-productive Severity: mild Duration: 2 days  

Clinical Notes: patient has had mild cough and nasal congestion for 2 days. Denies difficulty breathing or chest pain. Vitals are stable, lung auscultation clear. Throat slightly erythematous, no exudates. No significant past medical history.  
Plan: advise hydration, rest, over-the-counter analgesics.

Expected JSON Output: \{ "Response": [ \{ "Severity": "Green", "Reason": "Documentation is sufficiently complete for a patient with mild cough." \} ], "Recommendations": [ \{ "Severity": "Green", "Action": "Continue with current plan; no critical gaps identified." \} ] \}

Yellow Example

Age: 30 y Gender: Female 

Vitals: Temperature 37.20 °C, Pulse 80 bpm, Respiratory rate 18 bpm, Blood Pressure 110/70, SPO2 99, BMI –, Weight 60 kg, Height 162 cm, MUAC –  

Chief complaints: Abdominal Pain  
Notes: dull ache, lower abdomen Severity: moderate Duration: 3 days  

Clinical Notes: patient complains of lower abdominal pain for 3 days. Denies severe nausea or vomiting. Bowel movements normal, no blood in stool. Vitals stable. Physical exam: mild tenderness lower abdomen, no guarding/rebound. No urinary symptoms documented. No mention of menstrual history.  
Plan: pain relief with NSAIDs, dietary modification.

Expected JSON Output: \{ "Response": [ \{ "Severity": "Yellow", "Reason": "Documentation is mostly complete but lacks details on potential urinary symptoms and menstrual history, which could be relevant for abdominal pain." \} ], "Recommendations": [ \{ "Severity": "Yellow", "Action": "Inquire about urinary frequency, dysuria, or menstrual pattern. This will help rule out UTIs or gynecological causes." \} ] \}

Red Example

Age: 4 y Gender: Male 

Vitals: Temperature 39.00 °C, Pulse 104 bpm, Respiratory rate 16 bpm, Blood Pressure 122/78, SPO2 99, BMI 22.9, 

Weight 22 kg, Height 98 cm, MUAC –  

Chief complaints: Headache  
Notes: severe headache, child febrile Severity: severe Duration: 2 days  

Clinical Notes: child has had high fever (39 °C) and severe headache for 2 days. Appears lethargic, uncomfortable, occasionally sleepy. Paracetamol 200 mg PR administered.

Expected JSON Output: \{ "Response": [ \{ "Severity": "Red", "Reason": "Severe headache and high fever in a young child without documentation of meningeal signs or neuro exam, posing a critical gap for possible meningitis." \} ], "Recommendations": [ \{ "Severity": "Red", "Action": "Assess neck stiffness, photophobia, and perform a focused neurological exam immediately to rule out meningitis." \} ] \}

\subsection{Components of the user prompt for clinical notes}

\begin{itemize}
    \item Age
    \item Gender
    \item Structured history like pregnancy status, if recorded
    \item Vitals:
    \begin{itemize}
        \item Temperature
        \item Pulse rate
        \item Blood pressure
        \item Respiratory rate
        \item SPO2
        \item Weight
        \item Height
        \item Mean upper arm circumference (MUAC)
    \end{itemize}
    \item Chief complaint
    \item Clinical notes
\end{itemize}

\clearpage
\subsection{System prompt for investigations}
You are an advanced Clinical Decision Support System (CDSS) integrated within an Electronic Medical Record (EMR) system in a network of urgent care centers in Nairobi, Kenya, staffed by Clinical Officers.

Your Role  

Evaluate the investigations (lab tests, imaging, etc.) ordered by the clinician against the rest of the visit documentation (patient history, exam findings, local context).

Determine if the investigations:  

• Meet the standard of care for the presenting complaint / diagnosis.  

Please note that chief-complaint fields come from an automated system that includes “Severity and duration: Not recorded,” so please reference the clinical-notes free text for history / physical-exam documentation.  

• Are missing or excessive given the documented scenario.  

• Are feasible in an outpatient setting (avoid penalizing for not ordering tests that would be done at a higher-level facility).

Severity Thresholds  

1. Green

• The investigations ordered are appropriate and comprehensive for the clinical scenario.  

• No critical tests are missing; no irrelevant or unjustified tests are ordered.  

• Example: A strep test ordered for a patient with sore throat and exudative tonsillitis, or a urine dipstick for suspected UTI.

2. Yellow

• Some recommended investigations are missing or questionable based on the history / exam, but not so critical as to seriously endanger the patient.  

• OR there is at least one low-value or marginally justified test ordered.  

• Example: Mild pallor noted but no full haemogram ordered, or a borderline-unnecessary test (e.g., routine stool analysis in a non-GI complaint).

3. Red

• Essential diagnostic investigations are omitted, posing a risk of delayed or inaccurate diagnosis.  

• Clearly inappropriate tests are ordered, showing a major mismatch with the documented presentation.  

• Example: A patient with severe chest pain but no cardiac or respiratory investigations ordered; or a stool test for a purely respiratory complaint with no GI symptoms.

Key Principles  

• Context Relevance: 

  – Tie each ordered or missing test to the presenting complaint, vitals, and exam findings.  
  
  – Do not mark a test as missing if it is typically done at a higher-level facility (e.g., advanced imaging) and the scenario is an outpatient urgent-care clinic.  

• Outpatient Feasibility:  

  – Some conditions (e.g., severe pre-eclampsia) require basic tests (urinalysis, BP checks) in the outpatient setting, but advanced labs might need referral.  
  
  – If a key test is missing but is typically done at higher-level care, use Yellow to recommend referral or additional testing rather than penalizing with Red.  

• Urine Analysis:  

  – Consider urinalysis as both dipstick and urine microscopy if indicated in the scenario.  
  
  – Missing a simple urine dipstick in a suspected UTI is a bigger oversight than missing, say, an advanced culture that might need referral.  

• Actionable Recommendations:  

  – Provide short, specific steps: “Add test X,” “Omit test Y,” or “Refer to a higher-level facility for advanced imaging.”  
  
  – If everything is appropriate, confirm with Green and a brief note of affirmation.

Output Structure  

You must return exactly one severity level (Green, Yellow, or Red) in JSON, with an explanatory “Reason” and an “Action” recommendation.

\{
  "Response": [
    \{ "Severity": "Red", "Reason": "" \}
  ],
  "Recommendations": [
    \{ "Severity": "Red", "Action": "" \}
  ]
\}

• Severity: “Green”, “Yellow”, or “Red” 

• Reason: Concise explanation of why this severity applies.  

• Action: Recommended next steps to improve or confirm the investigation plan.

Important  

• If you determine Red is correct, do not include Yellow or Green.  

• If you determine Yellow is correct, do not include Red or Green.  

• Omit any severities that do not apply.

Sample Scenarios  

Green Example

Age: 32 y Gender: Male  

Vitals: Temperature 37.8 °C Pulse 84 bpm Respiratory Rate 16 bpm  
Blood Pressure 120/80 SpO$_2$ 98 BMI: Not recorded Weight 70 kg Height 172 cm MUAC: Not recorded 

Chief complaints: Sore Throat  

Clinical Notes: Patient reports mild sore throat for 2 days with low-grade fever.  
On examination, tonsils are inflamed with exudates noted.  
Plan: Provide symptomatic relief and confirm streptococcal infection if present.  

Investigations Ordered: Rapid Strep Test  

Expected JSON Output:  
\{
  "Response": [
    \{ "Severity": "Green",
      "Reason": "Strep test aligns with the clinical suspicion of strep throat. "
                "No missing or unnecessary tests identified." \}
  ],
  "Recommendations": [
    \{ "Severity": "Green",
      "Action": "Proceed with the ordered Strep test; no additional investigations are required at this time." \}
  ]
\}

Yellow Example

Age: 30 y Gender: Female  

Vitals: Temperature 37.5 °C Pulse 82 bpm Respiratory Rate 18 bpm  
Blood Pressure 110/70 SpO$_2$ 99  
BMI: Not recorded Weight 60 kg Height 162 cm MUAC: Not recorded  

Chief complaints: Throat Pain  

Clinical Notes: Mild sore throat and intermittent cough for 3 days; mild pallor on exam.  

Investigations Ordered: Rapid Strep Test  

Expected JSON Output:  
\{
  "Response": [
     \{"Severity": "Yellow",
      "Reason": "Strep test is appropriate, but a full haemogram is recommended to evaluate pallor." \}
  ],
  "Recommendations": [
     \{"Severity": "Yellow",
      "Action": "Consider ordering a full haemogram to assess possible anemia." \}
  ]
\}

Red Example

Age: 45 y Gender: Female  

Vitals: Temperature 38.5 °C Pulse 100 bpm Respiratory Rate 20 bpm  
Blood Pressure 130/85 SpO$_2$ 98  
BMI: Not recorded Weight 68 kg Height 165 cm MUAC: Not recorded  

Chief complaints: Chest Pain (severe, radiating to left arm, with sweating)  

Investigations Ordered: Stool Analysis  

Expected JSON Output:  

\{
  "Response": [
     \{"Severity": "Red",
      "Reason": "A stool test is not indicated for severe chest pain with possible cardiac involvement. "
                "Essential cardiac or respiratory investigations are missing." \}
  ],
  "Recommendations": [
     \{"Severity": "Red",
      "Action": "Discontinue stool test and order ECG, cardiac enzymes, or appropriate imaging to rule out myocardial infarction." \}
  ]
\}

\subsection{Components of the user prompt for investigations}

\begin{itemize}
    \item Age
    \item Gender
    \item Structured history like pregnancy status, if recorded
    \item Vitals:
    \begin{itemize}
        \item Temperature
        \item Pulse rate
        \item Blood pressure
        \item Respiratory rate
        \item SPO2
        \item Weight
        \item Height
        \item Mean upper arm circumference (MUAC)
    \end{itemize}
    \item Chief complaint
    \item Clinical notes
    \item Investigations and laboratory results
\end{itemize}

\clearpage
\subsection{System prompt for diagnosis evaluation}

You are an advanced Clinical Decision Support System (CDSS) integrated within an Electronic Medical Record (EMR) system in Nairobi, Kenya, staffed by Clinical Officers.

Your role is to:

1. Evaluate the clinician’s diagnosis against the visit’s documentation (patient history, exam findings, vitals, labs, etc.).

2. Assess if the diagnosis is appropriate, missing, incomplete, or incorrectly severe given the local epidemiology and available resources.

3. Provide concise, actionable recommendations to guide safe and quality patient care.

Severity Thresholds

1. Green

• The listed diagnosis (or diagnoses) accurately reflects the clinical documentation.  

• No significant mismatch with history, vitals, labs, or local context.  

• The clinician may safely proceed with management of these diagnoses.  

• Example: If the patient presents with dysuria, urgency, and a positive urinalysis, diagnosing a straightforward UTI is Green.

2. Yellow

• The listed diagnosis broadly aligns with the documentation, but:  

  – There is some uncertainty or missing details preventing a definitive conclusion (e.g., possible severe pathology but not fully confirmed).  
  
  – Additional testing or more thorough documentation is advisable before finalizing.  
  
• Severe diagnoses (e.g., sepsis, meningitis, appendicitis) in outpatient settings can be Yellow if the clinical picture could be correct but is not definitively confirmed—urge confirmatory testing or referral.

• Example: If the patient’s symptoms might be early appendicitis, but no imaging or sufficient exam details are available, classify as Yellow with guidance to do further testing.

3. Red

• A serious mismatch: The listed diagnosis is incompatible with the clinical findings, or a critical diagnosis is missing.  

• Could result in dangerous consequences if not corrected.  

• Severe diagnoses listed are not supported by the presentation, or a severe condition is clearly overlooked.  

• Example: The patient has signs of acute pyelonephritis (fever, flank tenderness, significant leukocytosis), but the diagnosis is “simple cystitis” with no mention of possible pyelonephritis.

Key Notes

You must flag serious and evident diagnoses if they are not listed.

For example, consider a patient with cough and a full haemogram that shows an elevated white count and anemia. If the clinician diagnoses “Bronchitis,” that may very well have a green alignment to the rest of the documentation; however, failing to list the anemia as a diagnosis should result as a red flag until that is addressed. Similarly, malnutrition in children as evidenced by low weight for age or yellow or red MUAC must result in those diagnoses being listed by the clinician.

• How to interpret MUAC:  

  – Red = Severe malnutrition (urgent concern).  
  
  – Yellow = Moderate malnutrition.  
  
  – Green = Normal MUAC or no malnutrition.

• Severe Diagnoses in Outpatient:  

  – If documentation aligns but is not conclusive (e.g., possible meningitis, appendicitis), return Yellow with instruction for urgent referral or confirmatory tests.  
  
  – If the severe diagnosis does not match the clinical presentation, return Red and advise re-evaluation.

You should consider local epidemiology without over-indexing on tropical diseases.

Output Structure

Return exactly one severity level in JSON (Green, Yellow, or Red).  
Include a Reason and an Action.

\{
  "Response": [ \{ "Severity": "Green", "Reason": "" \} ],
  "Recommendations": [ \{ "Severity": "Green", "Action": "" \} ]
\}

Few-Shot Examples

Below are three sample scenarios illustrating Green, Yellow, and Red responses.

1. Green Example

Age: 25y  Gender: Female  

Vitals: Temperature: 37.80 °C   Pulse: 80 bpm   Blood Pressure: 120/78   Respiratory Rate: 18   SPO2: 99  

Chief Complaint: Dysuria, urinary frequency

Clinical notes: Pt complains of dysuria and urinary frequency x2 days. She has had these symptoms before and was diagnosed with UTI. She does not currently have a sexual partner. On exam, she is well appearing. She has suprapubic tenderness to palpation, but abdomen is otherwise soft and non-tender. There is no CVA tenderness.

Lab Results: Urinalysis shows nitrites and leukocytes  

Diagnosis: Uncomplicated Urinary Tract Infection (UTI)

Expected JSON Output:  
\{
  "Response": [ \{ "Severity": "Green", "Reason": "Diagnosis of UTI aligns with clinical presentation of dysuria and urinalysis findings." \} ],
  "Recommendations": [ \{ "Severity": "Green", "Action": "Proceed with standard treatment for uncomplicated UTI (e.g., nitrofurantoin)." \} ]
\}

2. Yellow Example

Clinical Documentation: Age: 16y  Gender: Male  

Vitals: Temperature: 38.50 °C   Pulse: 90 bpm   Blood Pressure: 110/70   Respiratory Rate: 20   SPO2: 98  

Chief Complaint: Right lower quadrant abdominal pain, mild nausea

Physical Exam: Mild tenderness in RLQ but no rebound or guarding  

Lab Results: WBC count slightly elevated  

Diagnosis: Appendicitis

Expected JSON Output:  
\{
  "Response": [ \{ "Severity": "Yellow", "Reason": "Appendicitis is plausible but not definitively confirmed. Documentation suggests mild RLQ tenderness without peritoneal signs." \} ],
  "Recommendations": [ \{ "Severity": "Yellow", "Action": "Obtain an ultrasound or surgical consult to confirm appendicitis. Monitor for worsening pain, fever, or signs of rebound tenderness." \} ]
\}

3. Red Example

Clinical Documentation  

Age: 35y  Gender: Female  

Vitals: Temperature: 39.20 °C   Pulse: 105 bpm   Blood Pressure: 130/85   Respiratory Rate: 22   SPO2: 98  

Chief Complaint: Flank pain, fever, nausea

Physical Exam: Notable costovertebral angle tenderness  

Lab Results: WBC count elevated, presence of pyuria on urinalysis  

Diagnosis: Simple UTI (cystitis)

Expected JSON Output:  
\{
  "Response": [ \{ "Severity": "Red", "Reason": "Clinical findings (fever, flank pain, pyuria) are more consistent with pyelonephritis than simple cystitis." \} ],
  "Recommendations": [ \{ "Severity": "Red", "Action": "Reevaluate diagnosis. Consider inpatient management or a more aggressive antibiotic regimen for pyelonephritis." \} ]
\}

\subsection{Components of the user prompt for diagnosis evaluation}

\begin{itemize}
    \item Age
    \item Gender
    \item Structured history like pregnancy status, if recorded
    \item Vitals:
    \begin{itemize}
        \item Temperature
        \item Pulse rate
        \item Blood pressure
        \item Respiratory rate
        \item SPO2
        \item Weight
        \item Height
        \item Mean upper arm circumference (MUAC)
    \end{itemize}
    \item Chief complaint
    \item Clinical notes
    \item Investigations and laboratory results
    \item Diagnosis
\end{itemize}

\clearpage
\subsection{System prompt for treatment}

You are an advanced Clinical Decision Support System (CDSS) integrated within an Electronic Medical Record (EMR) in Nairobi, Kenya, staffed by Clinical Officers.

Your role is to:

1. Evaluate the clinician’s treatment plan against the visit documentation (vitals, diagnosis, labs, etc.).

2. Identify if the treatment is safe, evidence-based, and aligned with local guidelines (e.g., MoH Kenya, IMNCI/WHO).

3. Provide concise, actionable recommendations to ensure appropriate and safe patient care.

Severity Thresholds 

1. Red  

• Serious mismatch between treatment and diagnosis.  

• Unsafe or unnecessary medications (e.g., antibiotics for a confirmed viral illness, sedating antihistamines in young children, monteleukast for respiratory infections without asthma).  

• Omission of essential medications when clearly indicated (e.g., no rehydration plan and zinc in severe pediatric dehydration).  

  Please also consider the omission of clearly indicated procedures or referrals (examples: inpatient hospitalization for severe sepsis, consultation with general surgeon for ruptured ovarian cyst, or incision and drainage for a superficial abscess).  

• Incorrect dosage, major drug interactions, or known contraindications (such as aspirin in young children).  

• Could pose significant harm to the patient if not corrected immediately.  

2. Yellow  

• Treatment plan mostly aligns with the documented diagnosis, but:  

 • Minor adjustments to dosage/duration are recommended, or while the medication choice is acceptable, it is not considered a first-line treatment for the condition.  

 • Some prescriptions listed are of dubious value to the patient (e.g., cough syrups).  

 • Additional medication(s) could improve outcomes.  

• No immediate patient risk, but refinement is advisable.  

3. Green  

• Treatment plan is complete, accurate, and in compliance with relevant guidelines.  

• No critical omissions or unnecessary interventions.  

Specific Guidelines to note 

1. Key IMNCI/WHO Guidance for Dehydration in Children <5 Years  

1. Severe Dehydration  

• IV Ringer’s Lactate at 30 mL/kg over 30 min (if child > 12 months) or 60 min (< 12 months).  

• Then 70 mL/kg over 2.5 hours (> 12 months) or 5 hours (< 12 months).  

• If IV access is not possible, ORS via nasogastric tube at 120 mL/kg over 6 hours.  

2. Some Dehydration  

• Oral Rehydration Solution (ORS) at 75 mL/kg over 4 hours.  

3. No Dehydration  

• ORS 10 mL/kg after each loose stool.  

All cases should receive zinc supplementation.  

2. Urinary Tract Infection Management  

In Kenya, Nitrofurantoin and Cephalosporins are appropriate first-line therapy for management of UTI in adults and pregnant women.  
 
Septrin (cotrimoxazole) is not a recommended first-line treatment due to its use in TB management.  

3. Note that Zefcolin (brand name) is a cough syrup and not a cephalosporin antibiotic; it can be used to relieve cough symptoms associated with upper respiratory tract infection in adults and children over 2 years.  

Output Structure 

You must return exactly one severity level in JSON, with an explanatory Reason and an Action:  

\{ "Response": [ \{ "Severity": "Green", "Reason": "" \} ], "Recommendations": [ \{ "Severity": "Green", "Action": "" \} ] \}  

Few-Shot Examples  

Below are three scenarios highlighting Green, Yellow, and Red outcomes.  

Green Example  

Clinical Documentation: Age: 18y Gender: Female 

Vitals: Temperature: 37.5 C Pulse: 78 bpm Blood Pressure: 115/75 Respiratory Rate: 16 SPO2: 98 

Diagnosis: Uncomplicated Cystitis (UTI confirmed by urinalysis) 

Treatment Plan: - Nitrofurantoin 100 mg twice daily for 5 days  

Expected JSON Output: \{ "Response": [ \{ "Severity": "Green", "Reason": "Treatment aligns with recommended guidelines for uncomplicated cystitis." \} ], "Recommendations": [ \{ "Severity": "Green", "Action": "Proceed with nitrofurantoin therapy. Advise patient on possible side effects and encourage fluid intake." \} ] \}  

Yellow Example  

Clinical Documentation: Age: 5y Gender: Male 

Vitals: Temperature: 37.8 C Pulse: 100 bpm Blood Pressure: Not recorded Respiratory Rate: 20 SPO2: 98 

Diagnosis: Mild Pneumonia 

Treatment Plan: - Amoxicillin 125 mg twice daily for 3 days - No mention of supportive care (e.g., hydration, fever management)  

Expected JSON Output: \{ "Response": [ \{ "Severity": "Yellow", "Reason": "Antibiotic choice is appropriate, but dosage duration may be suboptimal, and supportive care isn't addressed." \} ], "Recommendations": [ \{ "Severity": "Yellow", "Action": "Consider extending amoxicillin to 5 days total, ensure fever management with paracetamol, and advise adequate fluid intake." \} ] \}  

Red Example  

Clinical Documentation: 

Age: 2y Gender: Female 

Vitals: Temperature: 39.0 C Pulse: 120 bpm Blood Pressure: Not recorded Respiratory Rate: 24 SPO2: 97 Weight: 12 kg 

Diagnosis: Acute Gastroenteritis with Severe Dehydration 

Treatment Plan: - Oral paracetamol for fever  

\{ "Response": [ \{ "Severity": "Red", "Reason": "Severe dehydration diagnosis without IV fluids or ORS is a critical omission. No zinc supplementation is prescribed." \} ], "Recommendations": [ \{ "Severity": "Red", "Action": "Initiate IV rehydration per IMNCI guidelines or give ORS if IV not feasible. Include zinc supplementation for diarrheal disease." \} ] \}  

\subsection{Components of the user prompt for treatment}

\begin{itemize}
    \item Age
    \item Gender
    \item Structured history like pregnancy status, if recorded
    \item Vitals:
    \begin{itemize}
        \item Temperature
        \item Pulse rate
        \item Blood pressure
        \item Respiratory rate
        \item SPO2
        \item Weight
        \item Height
        \item Mean upper arm circumference (MUAC)
    \end{itemize}
    \item Chief complaint
    \item Clinical notes
    \item Investigations and laboratory results
    \item Diagnosis
    \item Medications
    \item Referrals
\end{itemize}

\FloatBarrier
\clearpage
\section{Follow-up call script}
\label{app:call-center-script}

\paragraph{Context.} This is Penda’s call center script. There are two bolded questions below. For these questions, please \textbf{ask them in the same way every time. }It’s very important for us to be rigorous here, so Penda can collect good outcomes data.

When patients respond to the bolded questions, you can go back to engaging the patient, by asking clarifying questions and confirming the patient’s response, to make sure we get the most accurate measurements of outcomes possible.

Make sure to confirm the answer with the patient. The outcomes we measure should be reported by the patient – we don’t want to assume how the patient is feeling!

\paragraph{Script.}

\begin{itemize}
    \item Start the conversation:
    \begin{itemize}
        \item \textit{Greetings and self-introduction: }Hello [patient or parent/guardian name], my name is [name] from Penda.
        \item \textit{Confirmation: }Is this \_\_\_ / the [mother/father/guardian] of \_\_\_?
        \item \textit{Provide reason for calling: }I am calling to (check on you and) collect feedback after your recent visit at Kasarani branch. Do you have a minute to answer a few questions?
    \end{itemize}
    \item Question 1A: \textbf{Would you say you are feeling better, just the same or worse after treatment?}
    \item Question 1B:
    \begin{itemize}
        \item \textbf{\textit{If 1A is better:} \textbf{Glad to hear that. Is it much better or a little better?}}
        \item \textit{\textbf{\textit{If 1A is worse:} \textbf{Sorry to hear that. Is it much worse or a little worse?}}}
        \item \textit{\textit{If 1A is the same: proceed to question 2}}
        \item \textbf{\textbf{When the patient responds, confirm the phrase, to make sure you have heard correctly they have a chance to correct themselves if needed}}
        \begin{itemize}
            \item Example: if they respond “5”, say “so you’re feeling much better”
            \item Example: if they say, “a little worse”, say “okay, so a little worse”
        \end{itemize}
        \item \textbf{Always confirm} – this lets us make sure what we record is what the patient means! For example, if the patient says “I’m still in recovery”, this could be anything between “feeling much worse” and “feeling a little better”, and we need to confirm how they feel. 
    \end{itemize}
    \item Question 2:\textbf{ “Did you get any of your treatment and medicines away from Penda Health”} Options: I received all my treatment and medicines at Penda; I visited another chemist; I went myself to another hospital or specialist, Penda referred me to another hospital or specialist
    \begin{itemize}
        \item \textbf{\textbf{Here, make sure to be sure of the patient’s answer and confirm it verbally. }}
        \begin{itemize}
            \item If they say “no”, say “so you didn’t need to go anywhere else, not even a chemist” – we need to make sure patients know we’re including chemists
            \item If they say “I went to the hospital”, say “did Penda refer you, or did you decide to go yourself?”, so you can pick the right answer!
        \end{itemize}
        \item \textbf{\textbf{The patient might say this information without you asking, when you ask Question 1. If they do, confirm their answer verbally!}}
        \item \textbf{Ask this question even if the patient is feeling better! }They may be feeling better because they have already gone to another hospital or chemist
        \item \textbf{If the patient plans to visit another clinic but hasn’t yet, answer “No”! }This question is about whether the patient has done so already
    \end{itemize}
    \item End the conversation:
    \begin{itemize}
        \item Other comments: If you feel it important, ask any additional questions (e.g., check if they are still taking medication or do other checks on their condition). 
        \begin{itemize}
            \item Please make sure to flag severe outcomes like hospital admission, ICU admission, or death here. Please also flag home remedies if they are mentioned.
        \end{itemize}
        \item If the patient provides any feedback or other notes, include that here.
        \item Let them know they can seek care at any of our branches or call this number
        \item End call
    \end{itemize}
\end{itemize}

\FloatBarrier
\clearpage
\section{Clinician survey}
\label{app:clinician-survey}

Survey: satisfaction with Penda's EMR.

Hi there!

Thanks for taking the time to participate in this short survey, which should take less than 5 minutes to complete.

We would like to understand your experience with the Penda electronic medical record (EMR) that you have personally been using \textbf{from January until early April.} \textbf{During that time, you {DID / DID NOT} have AI Consult available to you.}

We are interested in all aspects of the EMR, including features like clinical decision support that are part of the EMR.

All individual responses will be kept confidential and used solely for quality improvement and research purposes.

How does this \textbf{EMR} change the quality of the care that you deliver, compared to the quality of care you would deliver without an EMR system?
\begin{itemize}
    \item Substantially improves quality
    \item Somewhat improves quality
    \item Does not change quality
    \item Somewhat worsens quality
    \item Substantially worsens quality
\end{itemize}

What is the primary reason for your rating above?
{FREE TEXT FIELD}

{THE BELOW QUESTIONS WERE ASKED ONLY TO CLINICIANS IN THE AI GROUP}

Think about the version of Penda's \textbf{AI clinical decision support system}you've been using from January through early April. On a scale of 0–10, how likely would you be to recommend it to a similar clinic?
{SCALE; 0: Not at all likely to 10: Extremely likely}

Overall, how satisfied or dissatisfied are you with the \textbf{AI clinical decision support system}?
\begin{itemize}
    \item Very satisfied
    \item Somewhat satisfied
    \item Neither satisfied nor dissatisfied
    \item Somewhat dissatisfied
    \item Very dissatisfied
\end{itemize}

How does the \textbf{AI clinical decision support system} change the quality of the care that you deliver, compared to the quality of care you would deliver without the system?

\begin{itemize}
    \item Substantially improves quality
    \item Somewhat improves quality
    \item Does not change quality
    \item Somewhat worsens quality
    \item Substantially worsens quality
\end{itemize}

Please share any feedback you have, both positive and constructive, about the AI Consult tool. 
{FREE TEXT FIELD}

\FloatBarrier
\clearpage
\section{Additional eligibility criteria for one-day follow-up calls}
\label{app:one-day-eligibility}
Patients with below characteristics were marked as eligible for the one-day follow up call. Ultimately, Penda's call center team decided which patients to call. 

Any visit with the below diagnosis:
\begin{itemize}
    \item Severe malaria
    \item Acute viral or bacterial gastroenteritis with some or severe dehydration
    \item Severe pneumonia
    \item Pneumonia, in patients under 5 years of age and above 50 years of age
    \item Puerperal sepsis
    \item Neonatal sepsis
    \item Myocardial infaction or angina
    \item Hypertensive emergency and urgency
    \item Acute abdomen
    \item Ectopic pregnancy
    \item Stroke
    \item Acute coronary syndrome
    \item Pre-emplasia and clampsia
    \item Diabetic ketoacidosis
    \item Hypoglycemia
    \item Poisoning
    \item Gastroenteritis with some or severe dehydration
    \item Head injury
    \item Febrile convulsions
    \item Convulsions
\end{itemize}

Any visit with the below chief complaints:
\begin{itemize}
    \item Difficulty in breathing
    \item Vaginal bleeding
    \item Fever
    \item Weakness
    \item Unconsciousness
    \item History of convulsion
    \item Poisoning
\end{itemize}

\FloatBarrier
\clearpage
\section{Form shown to physician raters to rate clinical documentation}
\label{app:rater-form}


\section*{Supplementary material}

# Main form

In this work, you will be evaluating the quality of medical documentation and medical decision making of outpatient clinical encounters by clinical officers working in resource-limited primary/urgent healthcare clinics in Nairobi, Kenya. Clinical officers, though not trained at the same level as a physician, play a vital role in delivering primary care in Kenya.

You'll review the following portions of clinical notes and rate each portion with both a Likert scale and a multi-select question (i.e., a multiple choice question where you can choose 1 or more of the options):

| Chief complaint, vitals, history, physical exam |
| Investigations |
| Diagnosis |
| Treatment: medications and referrals |

Healthcare context like local practice norms, epidemiology, and resource availability is always important when reviewing clinical encounters. The encounters you will evaluate occurred in an outpatient primary/urgent care setting in Kenya. Patients and clinicians have access to resources as displayed in the following table:

|  | Available in clinic, (inexpensive and fast) | Available but expensive and/or results not available within the visit | Not available within clinic, but patients may be referred to specialised centers that have these available |
|---|---|---|---|
| Vital Signs | Most standard equipment |  | Pediatric BP cuffs |
| Diagnostic Equipment | Stethescope, penlight, exam light, vaginal speculum |  | Otoscopes Ophthalmoscopes Reflex hammers |
| Investigations | Complete blood count, urinalysis and stool testing, blood glucose testing, malaria (RDT and smear), rapid strep testing, and HIV testing, urine and blood pregnancy testing | Metabolic panels, liver function testing, hormonal testing (e.g. TFTs), cancer marker testing, nucleic acid testing, cultures (urine, wound, blood) |  |
| Imaging | Ultrasound is available on demand (abdominal and linear probes). | Limited access to transvaginal ultrasound scans X-ray - limited access | CT/MRI imaging is not available (patients may at times come with |

|  |  |  | outside records of images) |
|---|---|---|---|
| Pharmacy | Standard outpatient pharmacy |  |  |
| Access to Physicians or Specialists |  |  | Requires referral to Nutrition, OB/GYN, Pediatrics, IM, other specialists |

Epidemiology varies around the world, and it is important to keep in mind what diseases are common locally and local practice patterns. While these instructions do not provide a comprehensive overview, please keep in mind the following disease states that may present more frequently in Kenya than in some other parts of the world:

| Malaria - [distinguishing](#) malaria from severe malaria; Kenyan [pediatric protocol](#) (page 33) |
|---|
| Other parasitic Infection (Entamoeba histolytica, Giardia, Schistosomiasis, pinworm) |
| Tuberculosis - [clinical symptoms of pulmonary and extrapulmonary TB](#) |
| Malnutrition - classification & management plans [pediatric protocol](#) (page 42- 43) |
| Gastroenteritis in children - classification & management plans [pediatric protocol](#) (page 36) |
| Pneumonia in children - Distinguishing severe cases - Kenyan [pediatric protocol](#) (page 46) |
| H. pylori infection |
| Rheumatic fever - [clinical guidance](#) |
| HIV/AIDS and related opportunistic disease: TB, meningitis, genital warts, ca cervix, etc. |
| Varicella |
| Rickets (vitamin D deficiency) |
| Conditions of pregnancy: malaria, anemia, pre-eclampsia/eclampsia |

Also, keep in mind that first-line treatment regimens for gastroenteritis with dehydration, malaria, *H. pylori*, and HIV may differ from those used in other parts of the world. In addition, ARV therapy and TB medication will not be initiated in these encounters, but instead should be referred to other HIV/TB care outpatient facilities.

In this clinical setting, the expectation for documentation is dependent on clinical reasoning - the degree of expected comprehensiveness changes depending on the clinical scenario. Clinical officers also have a lower level of expected clinical documentation in comparison to physicians, particularly related to history and physical examinations to rule out differential diagnoses. This is also a busy clinical setting, with often short back-to-back encounters serving a low resource area. Notes tend to be more sparse and direct, though pertinent positives and negatives to facilitate diagnostic reasoning should still be included in the note. For example:

*For a patient with normal vital signs and a typical story for URI, extensive documentation beyond historical red flags is not expected and physical examination may be limited to the respiratory system including throat/neck. Ear examinations are usually not documented due to lack of accessibility to otoscopes.*

*For a patient with fever without localizing signs, a much more comprehensive history and physical exam is expected.*

## Clinical note and vitals form and questions

> **Task Specific Clinical Note and Vitals**
>
> Vital Signs:
>
> Chief complaint:
>
> History taken and physical exam:

**Likert Score**: Grade the clinical note documentation on **thoroughness of the documentation of the history taken, physical exam and presence of a chief complaint and relevant vital signs.** This score reflects the detail and coverage of these specific elements - not the correctness of clinical reasoning or the appropriateness of treatment that may be documented in this note.

Notes:
- *Please disregard any assessment and plan info that you can see in the clinical note when assigning a Likert score **on this section**.*
- *Take into account the previously mentioned documentation expectations, as well as the limited resource setting. For example, no middle ear exams can be documented because there are no otoscopes.*
- *Pediatric vital sign expectations vary by age. No pediatric blood pressure cuffs are available, so no blood pressures are expected for age <13 years. An MUAC (mid upper arm circumference) is expected on children ages 6 months to 5 years.*

| 5 | **Thorough**: key components of the HPI and/or medical history elements are documented; relevant systems on physical exam are well documented; chief complaint and relevant vitals are present |
|---|---|
| 4 | **Reasonably thorough:** some relevant HPI and/or medical history elements are documented; some of the relevant system specific findings are documented on exam; chief complaint and most relevant vitals are present |
| 3 | **Limited**: history has limited symptom description or pertinent details (e.g., missing 2 or more relevant characterizations of the chief complaint, such as duration, onset, quality, severity, etc.); physical exam is limited; chief complaint or some relevant vitals are present |
| 2 | **Deficient:** history misses the most important key elements; physical exam is very |

|   | incomplete, excludes important details of relevant systems or misses one of the most important exam findings; chief complaint and/or important vitals missing |
|---|---|
| 1 | **Very poor**: history is extremely limited with no meaningful characterization of presenting symptoms; physical exam is missing entirely or nearly so; critical vitals missing |

MCQ (multi select):
In the clinical documentation, which of the following deficiencies are present? Choose all that apply.
1. Chief complaint is absent
2. Key details in the history are missing (e.g., characterization of chief complaint is lacking key elements such as onset, duration, or associated symptoms, etc. or pertinent medical history such as travel/sexual/family history are missing when they would be relevant, etc. or no documentation of allergies when it would be relevant.) Note that associated symptoms and duration of symptoms may be located in the chief complaint as well as in the clinical note.
3. Documentation of relevant systems on physical exam are absent (e.g., respiratory exam in a patient with cough, description of the rash in a patient with skin findings)
4. Pertinent vital signs are absent
5. None of the above

## Investigations form and questions

| **Task Specific Clinical Note** |
|---|
| Investigations: |

**Likert Score**: Grade the investigations *ordered* (or lack of them) on appropriateness, given the clinical documentation. Keep in mind that not all investigations are readily available and use the table on available resources in the instructions at the top of this task to aid in your evaluation. Consider that investigations that are unavailable or limited in availability should not be counted as omissions. For example, chest X-Ray is not available at most clinic locations and so lack of a chest X-ray in a patient with lower respiratory tract infection should not be penalized.

Notes:
- Tests ordered may not have results - they often are not performed when ordered due to cost of testing. This should not be penalized.
- When descriptions from multiple Likert scores apply, assign the lowest relevant Likert score. For example, if a task has both unjustified investigations ordered corresponding with Likert 3 and also has missing potentially helpful investigations corresponding with Likert 2, assign Likert 2 to the task.

- Patients may sometimes request investigations that are not related to their current clinical condition—often out of personal interest (e.g., blood grouping). Such tests should be clearly documented in the clinical notes section as **self-requested**, for example: *"Blood group – self request."* Do **not** consider or include any self-request tests when scoring or evaluating investigations in this section.
- Given Kenya's epidemiological context, a malaria test should be prioritized in any patient presenting with fever when:
    - There is no clear localization of infection based on history and physical exam – *not ordering a malaria test in such cases is considered a critical omission.*
    - There are symptoms pointing toward another likely cause of fever (e.g., URTI, UTI) – a malaria test may still be considered as additional testing, especially if the patient is from or recently visited a malaria-endemic area.
    - See [WHO guidelines for malaria](#) (page 158)

Full hemogram (complete blood count) is a test ordered quite often in this setting. If helpful in interpreting these results, you can use the corresponding, age-specific [reference ranges](#).
> *Note that pediatric normal values vary widely depending on the specific age and a more granular [pediatric reference range](#) is available.
> *When any of the values in the hemogram are abnormal, the result will include a flag that says "Result-Abnormal" somewhere within the test result. The specific placement of this flag is not meaningful, it is often not adjacent to the result that is actually abnormal.

| | |
|---|---|
| 5 | **Appropriate & Targeted:** All investigations are clearly indicated by the clinical context; point of care tests are utilized appropriately and there are no unjustified tests ordered; *or* no investigations are indicated and none are ordered |
| 4 | **Minor overordering:** Investigations are mostly appropriate with only minimal overtesting, (e.g., full hemogram for URTI symptoms with no major systemic symptoms or exam findings suggesting bacterial infection but one or two general symptoms like subjective fever or headache are present) |
| 3 | **Overly broad**: Investigations ordered are unjustified, (e.g., full hemogram for clearly simple URTI with unconcerning vitals and exam findings) |
| 2 | **Deficient:** Potentially helpful investigations are missing, (e.g., no rapid strep test for pharyngitis/tonsillitis) |
| 1 | **Very poor**: Critical omissions - clearly important investigations were not ordered, risking misdiagnosis or harm, (e.g., urinalysis not ordered for a young child with unexplained fever or no malaria test for a patient with fever and travel to a malaria endemic region) |

MCQ (multi select):
Which of the following is deficient regarding investigations ordered? Choose all that apply.
1. Key investigations are missing (i.e., important tests expected for this clinical scenario that are an available resource are not ordered)
2. Unjustified investigations are ordered

3. None of the above (Any investigations ordered are indicated and no key investigations are missing)

## Diagnosis form and questions

| **Task Specific Clinical Note** |
| --- |
| Diagnosis |

**Likert Score:** Grade whether the diagnosis (including primary diagnosis, any additional diagnoses, or the listed differential diagnoses) aligns with the clinical picture. Below, primary diagnosis refers to one or more diagnoses associated with the chief complaint. Additional diagnoses refers to any other diagnosis that may be relevant based on other listed history, physical exam findings or vital sign findings. Additional diagnoses are not necessarily less important than the primary diagnosis.

Note: Multiple diagnoses may be assigned that are all related to the chief complaint (e.g., acute tonsillitis, acute rhinitis, acute nasopharyngitis all being listed as diagnoses in the same encounter). In this case, you can consider all of them as the primary diagnoses for the purpose of evaluation with the Likert table and multi-select question. Depending on the scenario, you may consider this to be appropriate, not clearly relevant additional diagnoses, or catch-all diagnoses.

| 5 | **Excellent**: primary diagnosis fully aligns with the clinical picture and is the most likely diagnosis; no additional diagnoses are missing; any listed additional diagnoses are appropriate |
| --- | --- |
| 4 | **Good:** primary diagnosis generally aligns and is among the top few likely diagnoses; additional diagnoses are present but not clearly relevant |
| 3 | **Adequate**: primary diagnosis is plausible but not the most likely, or one or more important additional diagnoses are missing (e.g., no diagnosis of "elevated blood pressure reading" or "hypertension" when BP is significantly elevated) |
| 2 | **Deficient:** primary diagnosis is not well supported by the clinical picture or low on the list of likely causes; Primary diagnosis is a catch-all diagnosis when a clearer primary diagnosis is possible (e.g., "bacterial infection unspecified" instead of "urinary tract infection"); critical additional diagnoses are missing (e.g., no "malnutrition" or "underweight" diagnosis on a child with MUC yellow or red) |
| 1 | **Very poor**: primary diagnosis is missing or clinically inappropriate, contradicts or is unsupported by documented findings |

MCQ (multi select):
Which of the following is deficient regarding the assessment and diagnosis?
1. Primary diagnosis is likely incorrect
2. Primary diagnosis is missing

3. Primary diagnosis is too specific to be supported based on current documentation or investigations (e.g., using "allergic rhinitis" as the diagnosis rather than "rhinitis", where it's clear that rhinitis is present but documentation does not support whether it is a viral, bacterial or allergic etiology)
4. Additional diagnosis is likely incorrect
5. Clinically relevant additional diagnosis is missing (e.g. malnutrition)
6. None of the above (All diagnoses are likely correct and no clinically relevant diagnoses are missing)

## Treatment form and questions

> **Task Specific Clinical Note**
>
> Plan: Treatments and Referrals + procedures and escalations of care if present

**Treatment Likert Score**: Is the documented treatment plan appropriate and complete given the clinical scenario?

The majority of cases will have either medications or referrals documented as a plan. Some will have procedures, escalations of care, referral for additional investigations or home medications documented in the clinical note. If procedures, escalations of care or home meds are documented, take them into consideration with the rest of the plan. Follow up plans and patient advice or education are often not documented, but when present, these should be taken into consideration and reflected in the Likert scoring.

Note:
- Sometimes additional investigations will be added as a referral. This means that the investigation is not available locally but the provider is requesting the patient have it performed at another location.
- Treatments will sometimes be documented in the clinical notes, and these should also be taken into consideration.

| 5 | **Appropriate and complete**: treatments are appropriate and complete, including the correct use of medications and/or referrals when needed; Patient advice or education (e.g., red flag symptoms to watch for, advice on hydration, self-care, etc.) or a follow up plan is present; if procedures or escalations of care are present, they are the most appropriate course of action. *OR* no treatments are indicated and none are ordered |
|---|---|
| 4 | **Appropriate but less complete:** treatments are appropriate and complete, including the correct use of medications and/or referrals when needed but there is no patient advice, education or follow up plan documented (e.g., red flag symptoms to watch for, advice on hydration, self-care, when to return to care, etc.) |
| 3 | **Adequate**: Medications, referrals or procedures are reasonable and safe, but may not be a standard first-line therapy (i.e., things that are likely to have minimal benefit or |

|   | minimal harm if given, e.g. desloratadine for a URTI instead of nasal saline spray alone), or helpful but non-critical referrals are missing |
|---|---|
| 2 | **Deficient**: medications are present and somewhat inappropriate (i.e., medications that may cause minor harm unnecessarily e.g., inappropriately broad antibiotic class when a narrower spectrum is sufficient); minor medication dosage errors; clearly needed referrals, procedures or escalations of care are missing |
| 1 | **Very poor**: no medications given for a condition when clearly indicated; medications are very inappropriate for the condition (e.g., use of any antibiotic when there is no indication based on documented findings); significant dosage errors (e.g., too high of a dose based on pediatric patient weight-based dosing); procedures performed or escalations of care are unwarranted |

MCQ (multi select):

Which of the following deficiencies are present in the plan? Choose all that apply.
1. Medications are missing
2. Medications are present but inappropriate
3. Medications are appropriate but incorrect dosages listed (dosage can include dose quantity, frequency and duration)
4. Likely inappropriate use of antibiotics overall (e.g., antibiotics are given for a likely viral infection) *if you select this choice, you should also select choice "Medications are present but inappropriate"
5. Likely inappropriate class of antibiotics used (e.g., amox/clavulanic acid used when amoxicillin is appropriate) *if you select this choice, you should also select choice "Medications are present but inappropriate"
6. Referrals are missing
7. Referrals are present but inappropriate
8. Needed procedures are missing
9. Procedures are present but inappropriate
10. Needed escalations of care are missing
11. Escalations of care are present but inappropriate
12. None of the above

# Additional resources for physicians

Common local brand name pharmaceutical list that physicians in other parts of the world may not recognize

| | |
|---|---|
| **Cital syrup** | Disodium hydrogen citrate syrup 1.37g |
| **Benylin original syrup** | Diphenhydramine/ammonium chloride syrup 12.5mg/125mg |
| **Karvol Inh caps** | Camphor/eucalyptol/terpineol/chlorothymol caps |
| **Zefcolin syrup** | Dextromethorphan/phenylephrine/cetrizine syrup 10mg/5mg/5mg |
| **Benased chesty syrup** | Diphenhydramine/guaiphenesin syrup 14mg/100mg |
| **Delased pediatric syrup** | Diphenhydramine/sodium citrate syrup 7mg/28.5mg |
| **Imacoff dry syrup** | Diphenhydramine/dextromethorphan/sodium citrate syrup 14mg/5.7mg/5mg |
| **Metadoz fizz** | Paracetamol/tramadol tabs 325mg/37.5mg |
| **Brustan tabs** | Ibuprofen/paracetamol tabs 400mg/325mg |
| **Brustan syrup** | Ibuprofen/paracetamol suspension 100mg/125mg |
| **Acinet** | Amoxicillin and clavulanate 625mg tablets |
| **pcm** | Common abbreviation for paracetamol |
| **Adol** | Paracetamol a.k.a Acetaminophen available in many formulations. Frequently prescribed as a STAT suppository in children presenting with high fevers |
| **Desloratadine** | 2nd generation, Histamin H1 anatgonist |
| **Ventolin (Salbutamol)** | Albuterol |
| **Mara Moja/Parafast/Cipladon** | All paracetamol/acetaminophen |
| **PDL** | Abbreviation for Prednisolone |



## Common acronyms

| OE | On exam |
| --- | --- |
| FGC | Fair general condition |
| creps/crepitations on lung exam | rales/crackles |
| Hob (hotness of the body) | Subjective fever |
| Dib (difficulty in breathing) | dyspnea |
| JACCOLD | a normal constellation of physical exam findings that refers to absence of Jaundice, Anaemia (pallor), Cyanosis, Clubbing, Oedema (pedal edema), Lymphadenopathy, Dehydration |
| MUC or MUAC | mid upper arm circumference, used to identify children with malnutrition |
| Ass | associated |
| Vesicular breath sounds | Normal breath sounds |
| 5/7 | 5 days |
| 2/52 | 2 weeks |

# Likert Examples for chief complaint, vitals, history, and physical exam

Example 1:

> Patient Information:
> Age: 3 years
> Gender: Male
> Presenting Complaints:
> Dry Cough: Present.
> Vitals: HR 112, temp 36.7, wt 16 kg, rr 25 SpO2 97% MUAC - green
> Chest Congestion: History of chest congestion.
> Chronic Sinusitis: History of chronic sinusitis.
> Snoring at Night: Reported by the mother.
> Running Nose: Mild, present currently.
> No History of Fever: No recent fever.
> No History of Chills.
> No History of Travel.
> Past Medical History:
> No significant chronic illness reported.
> Physical Examination:
> General Appearance: Clinically stable patient.
> Vitals: Normal (within age-appropriate ranges).
> ENT Examination:
> Swollen Turbinates: The turbinates are swollen and reddish in color, suggesting inflammation (possibly allergic rhinitis or sinusitis).
> Respiratory Examination:
> Lungs Clear: Breath sounds are clear and equal on both sides.
> No Added Sounds: No wheezing, crackles, or other abnormal respiratory sounds.

**Likert Score: 5: Thorough**
**Explanation:** includes detailed history, includes HPI associated symptoms and documentation of no PMH, documents travel history and pertinent physical exam findings; CC and relevant vitals are present

Example 2:

> 47y2m
> CC: severe headache
> HR 88 temp 36.5 wt 87 kg rr 18 SpO2 98 BP 132/78
> severe headache on and off throbbing , sharp with no aggreviating factors
> mild to severe assoc with tears . it was of sudden onset , started today evening
> not relieved with oral painkillers
> no neck stiffness
> no dizziness
> no photophobia
> no history of trauma
> had hob with occassional chills
> no history of travel to malaria zone
> o/e. fgc , in severe pain
> neck soft not stiff
> afebrile on touch
> no photophobia
> brudzensk sign negative
> Glasgow Coma Scale at 15/15
> no confusison
> other systems esentially normal
> fam plan =none

> lnmp. 20th/3/2025
> impress. migraine/bac infection/menengitis /
> plan
> diclofenac 150mg Intramuscular stat
> fhg
> Blood slide for malaria
> head ct scan
> LP
> extplan
> metadoz 1tab bd
> explained danger signs
> consider head ct scan / lp

**Likert Score: 4 Reasonably thorough**
**Explanation**: includes CC, vitals, most of the relevant characteristics of the chief complaint and physical exam pertinent negatives on neuro exam; fails to mention presence or absence of nausea and vomiting, vision changes, weakness or numbness which are important symptoms in assessing severe headache; missing many specific neuro exam findings (CNs, strength, reflexes, cerebellar assessment)

Example 3

> Chief complaint: Cough, Sneezing
> 44y 8m
> Ht 155 HR 77 temp 36.6 wt 90 kg rr 18 SpO2 98 BP 110/72
> Presented with mild non productive cough associated with sore throat
> Reports also of runny nose with sneezing
> no fevers reported
> on and off with headache
> ros: nad
> o/e: stable, afebrile, no jaundice, no edema, no cyanosis
> ent: slightly inflamed pharynx, turbinates normal
> r/s: tranemitted breadth sounds
> other s/e: nad
> impression: bronchitis/rhinitis
> plan
> tx as prescribed

**Likert Score: 4 Reasonably thorough**
**Explanation**: includes CC, relevant vitals, reasonable amount of HPI described for a simple issue, though further characterization of the headache would be ideal; findings of relevant systems on exam are included, though not extensively

Example 3

> 4 y 9 mo old
> CC: fever
> Vitals: HR 94, temp 36.4, wt 17 kg, rr 19 SpO2 99% MUAC - yellow
> informant mother
> history of hotness of the body for 1day
> also has a dry cough for 1day
> no history of running nose
> no history of vomiting
> no history of diarrhoea
> no history of travelling to malaria endemic region
> premed antipyretic

> on examination fair general condition
> not pale no jaundice no cyanosis
> r/s chest clear
> ent hyperemic throat
> ⎯⎯⎯⎯⎯⎯⎯⎯⎯⎯⎯⎯⎯⎯⎯⎯⎯⎯
> plan
> fhg
> strep A ag test'
> deslit 5mls od 5/7
> brustan 5mls tds 3/5
> cefuroxime 250mg/5mls bd 5/7

**Likert Score: 3 Limited**
**Explanation:** missing PMH but reasonable HPI, although no mention of if ear pain exists or not; exam too limited, a 4 year old with fever should have a documented abdominal exam and skin exam and, without an otoscope, perhaps an external ear exam of pulling on ear and check for tenderness over mastoid, in addition to what is here

Example 4

> 4y 2m
> Informant-mother
> Hr 116 temp 36.2 wt 20 kg rr 25 SpO2 95 MUAC - green
> Presented with 1 day history of sudden onset of ear pains
> No history of discharge, no swelling
> No history of any injury or trauma
> Reports nasal blockage and cough especially in the evening and morning
> No difficulty in breathing, no fevers,
> Able to feed well, no vomiting, no diarrhea
> Premedication-Cetrizine
> On examination-In fair general condition
> Ear, Nose & Throat exam-Blocked nostrils,
> Chest exam-Bilateral air entry, no rhonchi, no crepitations
> Impression-Acute otitis media, acute rhinitis
> PLAN
> Amoxyclave 5mls bd 5/7
> Brustan 5mls Three times daily 3/7
> Saline drops 2drops Three times daily 3/7

**Likert Score: 3 Limited**
**Explanation**: most relevant history elements are documented; however, there is no ear exam documented for history of sudden ear pain. While no otoscope is available so no middle ear exam can be documented, evaluation for external otitis is possible through pulling on ear and evaluation for mastoiditis is possible through assessing for tenderness over mastoid; also no chief complaint

Example 5

> 8 year 2 month
> Vitals: HR 104, temp 36.7, wt 30 kg, rr 20 SpO2 99%
> diclofenac injection admnistred 75mg im
> Patient Summary:
> Age/Gender: 2-year-old male
> Informant: Father
> Presenting Reason:
> Follow-up after fall 2 weeks ago, presents with distorted gait and pain in the right hip.
> History:

> Fall: Fell at school 21 days ago while playing, hit right hip on a stone.
> Gait: Walking with a distorted gait, leaning to the right side.
> Pain Concealment: The child has been attempting to conceal pain.
> Swelling & Bruising: No swelling or ecchymosis noted.
> Tenderness: Tenderness elicited on the right hip.
> Pain with Movement: Obvious pain with both external and internal movements of the hip and with range of motion.
> Physical Examination:
> General Appearance: Clinically stable patient.
> Vitals: Normal.
> Musculoskeletal: Tenderness on the right hip with pain during range of motion (internal and external). No swelling or bruising.

**Likert Score: 2 Deficient**

**Explanation:** discrepancy of age listed; no documentation of if any recent fever, malaise, weight loss all needed for assessment of infectious causes of pain (septic arthritis, osteo) which can develop after an injury; no documentation of how the injury was cared for initially (prior imaging?); no documentation of spine or abdominal exam (can present with hip pain)

Example 6:

> 55 y 9 m
> KNWN hypertension patient ON MEDS
> TODAY BPS - 174/101 MMHG 80B/MIN
> HAD DONE LIPID PROFILE / UECS DONE , NORMAL PATIENT EXPLAINED
> BPS EXPLAINED
> PLAN
> NIFEDIPINE 40 MG
> START AMLOOZAH H

**Likert Score: 1 Very poor**

**Explanation:** patient presents with high BP and no documentation of if patient has any symptoms, no history is documented at all other than known HTN. Other PMH? Other meds? No physical exam documented.

Example 7

> 30y 5m
> HR 77 temp 36.4 wt 58 kg rr 16 Spo2 100
> history of vaginal discharge, no itchiness, no pain on micturation, no lower abdominal pains, no backa pains.
> reports of headache, with reported history of a stress,.

**Likert Score: 1: Very poor**

**Explanation**: extremely limited, omits several critical components: no chief compliant, extremely limited description of the vaginal discharge (no duration, description, associated symptoms of fever, pelvic pain, bleeding, sexual history, menstrual history) and no characterization at all of the headache; completely absent physical exam

# Likert Examples for Investigations

Example 1:

> 27y 1m
> CC: throat irritation
> it has been there for the laat 2 days
> he has been using warm water for the same but there is no much improvement
> there is some pain on the right side of his throat especialy when swallowing
> there is no fevers
> no chiklls
> no joint pains
> no runny nose
> he has been using saline water to gurgle
> no known food or drug allergy
> on exam
> not pale
> no jaundice
> no dehydration
> ent-slightly inflammed throat
> cvs- normal
> cns- normal
>
> Streptococcus A Antigen Test: Result: Negative

**Likert Score: 5 Appropriate & Targeted**
**Explanation:** Strep A test only is the correct investigation for the clinical scenario

Example 2:

> Chief Complaint: Skin Rash, Dry, itchy skin
> 30y 3m
> presented with dry skin rashes assoc with
> itchyness ,
> had severe itchynessespecially at the armpits
> it appers at the neck and armpits
> o/e ; fgc
> armpit/neck rash , shyny , scally and itchy, no raised borders
> impression; atpoic dermatitis/eczema
> plan
> betametasone/salicylic cream apply bd
> desloratadine 5mg od
> pdl 5mg bd
> keep skin moisurerized
>
> Investigations: none

**Likert Score: 5 Appropriate & Targeted**
**Explanation:** In this clinical scenario, it is most appropriate to not order any investigations as none are needed for diagnostic clarity or treatment plan generation.

Example 3:

> Chief Complaint (CC):
> 53-year-old female presenting with cough, runny nose, and sore throat for 1 day.
> History of Present Illness (HPI):
> The patient reports onset of symptoms yesterday, starting with a headache relieved by Mara Moja (analgesic), followed by nasal congestion, dry cough, and throat pain especially on swallowing. She also describes a sensation of chest heaviness without pain or difficulty in breathing. Additionally,

she experienced hotness of the body at night and general weakness. She denies joint pains, gastrointestinal symptoms, or recent travel to malaria-endemic areas. No other complaints reported.
Allergies & Medications:
Allergies: NKDFA
Premeds: Mara Moja (OTC analgesic)
Vitals (11/04/2025):
Temperature: 36.4°C
Pulse: 57 bpm
Respiratory Rate: 16 bpm
Blood Pressure: 117/74 mmHg
$SpO_2$: 97% on room air
BMI: 28.38 (Weight: 63 kg, Height: 149 cm)
Examination Findings:
General: Stable condition
ENT: Inflamed throat
Systemic Exam: Normal across respiratory, cardiovascular, and abdominal systems
Assessment & Plan:
Assessment:
Acute Upper Respiratory Tract Infection (URTI) – likely viral etiology
Supported by sore throat, dry cough, nasal congestion, and absence of systemic signs of bacterial infection
Rule out Streptococcal pharyngitis – given throat pain and inflamed pharynx
Investigations:
Streptococcus A antigen test
Complete blood count (CBC)
Investigations:
Strep A Antigen Test: Negative
CBC:
WBC: 4.54 x$10^9$/L (within normal range)
Neutrophils: 2.15 x$10^9$/L (low-normal)
Lymphocytes: 1.99 x$10^9$/L (normal)
Hemoglobin: 14.6 g/dL (normal)
Platelets: 360 x$10^9$/L (normal)
Mildly elevated lymphocyte % (44%) and reduced neutrophil % (47.3%)
Assessment & Plan:
Assessment:
Acute upper respiratory tract infection, likely viral in origin – suggested by sore throat, nasal congestion, dry cough, and negative Strep A test.
Viral pharyngitis or rhinitis is most likely; no evidence of bacterial infection (normal to low WBC, negative Strep A, afebrile, and stable vitals).
Overweight with BMI of 28.38 – lifestyle advice warranted.
Management Plan:
Symptomatic management:
Adequate fluids and rest
Paracetamol for headache and body aches if needed
Avoid unnecessary antibiotic use
Patient Education:
Condition likely viral; expected resolution within 5–7 days.
Importance of hydration and nutrition.

Streptococcus A Antigen Test: Result: Negative

Full Haemogram (FHG): PDW: 15.10, RDW-CV: 14.90, HCT: 40.30, MCHC: 36.20, WBC: 4.54, PCT: 0.31, MCV: 86.00, Lymphocyte Percentage: 44.00, Plt: 360.00, Lymphocytes Count: 1.99, MPV: 8.60, Mid-granulocyte Percentage: 8.70, Mid-granulocytes Count: 0.40, P-LCC: 65.00, MCH: 31.10, Result..: Abnormal, Neutrophill count: 2.15, Neutrophill percentage : 47.30, P-LCR: 18.00, RBC (Full Haemogram): 4.69, RDW-SD: 43.60, HGB: 14.60

**Likert Score: 4 Minor Overordering**

**Explanation:** Strep test indicated, full hemogram may not be necessary though not harmful

Example 4:

> 40 y 0m
> presents with above several days prior to the visit
> history of pain on urination
> no history of urethral discharge
> reports of abd bloating and abd pain
> no history of diarrhoea / vomiting
> no history of travel to a malaria endemic area
> on examination
> fgc , not Pale, no cy, no j, no deh2o
> Ear, Nose & Throat normal
> systemic examination- unremarkable
> imp ge / amoebiaisis / Peptic Ulcer Disease / bacterial infection / uti
> plan
> u/a
> fhg
> stool for o/c
> stool for h-pylory
> Diagnosis bacterial infection / acute bacterial ge / uti
> plan
> cefuroxime 500mg bd for 5 days
> myospaz 1 tab bd for 3 days
>
> H. Pylori - Stool: Result: Negative
>
> Full Haemogram (FHG): MCH: 23.50, HGB: 13.10, P-LCC: 61.00, RDW-CV: 15.80, P-LCR: 28.80, RBC (Full Haemogram): 5.58, RDW-SD: 39.30, Result..: Abnormal, HCT: 43.50, Mid-granulocyte Percentage: 9.20, PDW: 19.70, Neutrophill percentage : 75.20, MPV: 11.70, Neutrophill count: 5.95, WBC: 7.94, Mid-granulocytes Count: 0.73, MCHC: 30.10, PCT: 0.24, Lymphocyte Percentage: 15.60, Lymphocytes Count: 1.23, MCV: 77.90, Plt: 212.00
>
> Stool Microscopy: Result..: Abnormal, Yeast cells (Microscopy): None, Amount of yeast cells: None, Parasites: No Ova/Cyst, Crystals -- Type: None, Mucous: Absent, RBC's (Microscopy): None, Consistency : Semi formed, Colour : Brown, Crystals -- Amount: None, Blood (Gross Appearance): Absent, Pus cells: Moderate
>
> Urine Analysis: Result..: Abnormal, Nitrate: Negative, Bilirubin.: 0, Amount of yeast cells: None, Pus cells: Moderate, Specific Gravity: 1.02, Yeast cells (Microscopy): None, Crystals -- Amount (Microscopy): None , Colour.: Amber, Appearance (Urine Analysis): Clear, Casts - type: None, Urobilinogen: 0, RBC's (Urinalysis-Microscopy): None, Glucose: 0, Crystals -- Type: None, Leukocytes: +2, Parasites (Urine Microscopy): None, Epithelial Cells: None, pH (Dipstick): 6.00, Proteins: 0, Blood (Dipstick): 0, Trichomonads (Urinalysis-Microscopy): Absent, Ketones: 0

**Likert Score: 3 Overly broad**
**Explanation:** Urinalysis and stool testing are clearly appropriate. H pylori testing may be indicated given local prevalence but not much supporting documentation for it (abd pain + bloating, but no clear documentation of where in the abdomen the pain is located, no documentation around belching or excess gas, weight loss, or color of stools). Full hemogram is overly broad, no clear indication for it based on documentation.

Example 5:

> 5y 2m
> Clinical Presentation
> Patient Profile:

> A previously well child who became symptomatic yesterday.
> History of Presenting Illness:
> The child began experiencing abdominal pain, localized around the umbilical region, followed by vomiting. She has since had five episodes of non-projectile vomiting, containing ingested food and not blood-stained. The abdominal pain is intermittent.
> This morning, she developed loose, foul-smelling diarrhea, which is non-bloody. She also had fever in the morning, for which she was given Calpol.
> There are no known food or drug allergies.
> Physical Examination:
> General: Not pale, no jaundice
> Hydration Status: No dehydration; skin turgor <1 second, capillary refill <1 second
> Abdomen (P.A.): Non-distended, non-tender on palpation; no signs of umbilical hernia
> Cardiovascular System: Heart sounds S1 and S2 heard; no murmurs
> Respiratory System: Clear breath sounds
> Musculoskeletal System: Normal
> Central Nervous System: Normal
> Developmental Milestones: Normal
>
> Stool Microscopy: Crystals -- Type: None, Result..: Abnormal, Consistency : Liquid/Loose, RBC's (Microscopy): None, Parasites: No Ova/Cyst, Pus cells: Few, Blood (Gross Appearance): Absent, Colour : Brown, Yeast cells (Microscopy): None, Mucous: Absent, Amount of yeast cells: None, Crystals -- Amount: None

**Likert Score: 2 Deficient**

**Explanation:** Evaluation of stool microscopy looking for blood, parasites, pus/mucous is very appropriate given this presentation and regional epidemiology. A 5 year old with fever may also be evaluated with a malaria test in this scenario.

Example 6:

> 8 year 2 month
> Vitals: HR 104, temp 36.7, wt 30 kg, rr 20 SpO2 99%
> diclofenac injection admnistred 75mg im
> Patient Summary:
> Age/Gender: 2-year-old male
> Informant: Father
> Presenting Reason:
> Follow-up after fall 2 weeks ago, presents with distorted gait and pain in the right hip.
> History:
> Fall: Fell at school 21 days ago while playing, hit right hip on a stone.
> Gait: Walking with a distorted gait, leaning to the right side.
> Pain Concealment: The child has been attempting to conceal pain.
> Swelling & Bruising: No swelling or ecchymosis noted.
> Tenderness: Tenderness elicited on the right hip.
> Pain with Movement: Obvious pain with both external and internal movements of the hip and with range of motion.
> Physical Examination:
> General Appearance: Clinically stable patient.
> Vitals: Normal.
> Musculoskeletal: Tenderness on the right hip with pain during range of motion (internal and external). No swelling or bruising.
>
> Investigations: none

**Likert Score: 1: Very poor**

**Explanation**: child with hip pain and distorted gait and tenderness with exam 3 weeks after injury should have investigations looking for bacterial infection, CBC at minimum

# Likert Examples for Diagnosis

Example 1:

```
27y 1m
CC: throat irritation
it has been there for the last 2 days
he has been using warm water for the same but there is no much improvement
there is some pain on the right side of his throat especialy when swallowing
there is no fevers
no chiklls
no joint pains
no runny nose
he has been using saline water to gurgle
no known food or drug allergy
on exam
not pale
no jaundice
no dehydration
ent-slightly inflammed throat
cvs- normal
cns- normal

Streptococcus A Antigen Test: Result: Negative

Diagnosis: Pharyngitis
```

**Likert Score: 5: Excellent**
**Explanation**: Pharyngitis is the most likely accurate diagnosis for the clinical scenario. No other additional diagnoses are needed.

Example 2:

```
LMP-UPTO DATE
KNWN patient WITH PAINFUL PERIODS
DID FOLLOW UP WITH GYNAECOLOGICAL REVIEW AND WAS TOLD ITS NORMAL PAIN
HEAVY-USES 4-5PADS IN 2 ND DAY
THEN REDUCES GRADUALLY
NO HEADACHE
NO DIZZINESS
NO CHRONIC ILNESS
NO DRUG ALLERGY
ON EXAM-SHE IS STABLE
DX-DYSMENORRHOEA
PLAN
P.O PONSTAN 1 TAB BD FOR 5 DAYS

Diagnosis: Dysmenorrhoea
```

**Likert Score: 4: Good**
**Explanation**: Minimal history around the quality and timing of pain makes it challenging to say that this is clearly the most likely accurate diagnosis, but it aligns with the clinical picture and is among the top few likely diagnoses.

Example 3:

```
4y 7m
Presented with cough of gradual onset persistent and irritative associated
with runny nose
```

> Reports also of nasal blockage
> no fevers reported
> feeds well
> ros: nad
> o/e: stable, afebrile, no jaundice, no edema, no cyanosis
> ent: inflamed tonsils with exudates
> r/s: tranmitted breadth sounds, no crepitations, no rhonchi
> other s/e; nad
> impression: bronchitis/rhinitis/tonsilitis
> plan
> adviced on hydration/tx as prescribed
>
> Investigations: none
> Diagnosis: Acute bronchitis, tonsilitis, acute bacterial

**Likert Score: 3: Adequate**
**Explanation**: Acute bacterial tonsillitis is less likely than viral tonsillitis and no investigation was done to determine the cause. This diagnosis is plausible, but not the most likely.

Example 4:

> 0y 8m
> the mother reports that she has 3 days history of a runny nose
> associated with nasal congestion and mild fever
> no dib
> no cough
> no pre medication
> no history of drug allergy
> ON EXAM
> Afebrile
> underweight
> not pale/jaundiced.dehydrated
> clesar chest
> normal throat exam
> enlarged nasal turbinates
> INVESTIGATION
> full haemogram
> DIAGNOSIS
> acute rhinitis with enlarged nasal turbinates
> PLAN
> ephedrin nasal drop bd for 5 days
> calpol 5mls orally for 3 days
> aerius 2.5mls orally for 5 days
> rehydrate with warm fluids
>
> Diagnosis: Allergic rhinitis, Adenoid hypertrophy

**Likert Score: 2: Deficient**
**Explanation**: MUC yellow, not noted in the diagnosis field (though is noted on physical exam in clinical note); should have a diagnosis of underweight or malnourished in the diagnosis field; additionally the child is noted to have enlarged turbinates on exam but is given a diagnosis of adenoidal hypertrophy which is not well supported by the documentation.

Example 5:

> Has a complete miscarriage , was medical evacuated came with complains of vaginal discomfort , attending doctor requested her to come for medication three days post procedure.
> O/E

> stble
> no distress
> p/a
> -unremakble
> IMPRESSIONS: urinary tract infection
> Plan
> Per oral cefuroxime 500mg bd 5/7
> Pessaries Infa V 1 nocte 5/7
>
> Investigations: none
> Diagnosis: Uncomplicated Urinary Tract Infection (UTI)

**Likert Score: 1: Very poor**

**Explanation**: No dysuria, suprapubic abd pain or fever documented, chief complaint does not fit with UTI and explanation given for antibiotics is post procedure prophylaxis. No urinalysis is done. Diagnosis of UTI is clinically inappropriate and unsupported by documented findings.

# Likert Examples for Treatment

Example 1:

> 10y 5m
> Wt: 22 kg
> mother reports 1 week history of abdominal pains of gradual onset with no relieving nor exacerbating factor, child reports the pain is on the epigastric region, reports history of loose stools on and off with yesterday having a blood stained loose stools that has since resolved, no episodes of vomiting, child is currently on management for amoebiasis, on diracip-male syrup doing day 3 from another facilty, no fevers, mother reports reduced appetite, no other complaints reported.
> on exam: fgc, alert
> no jaundice, no pallor, no cyanosis, no dehydration, no edema
> heent- normal
> rs- chest clear on auscultation, no creps, no rhonchi
> cvs-s1, s2 heard, no murmurs
> p/a- not distended, soft non tender, no organomegally
> cns- alert and well oriented
> mss- normal findings
> impression: gastritis/dysentry
> plan
> h.pylori ag test
> stool o/c
> rx as per t-sheet
> mother advised to complete dosages
> advised on nutritional review- said will come next thursday
>
> Investigations: Stool Microscopy: Consistency : Formed, Blood (Gross Appearance): Absent, Mucous: Absent, Crystals -- Amount: None, Parasites: No Ova/Cyst, Result..: Normal, Yeast cells (Microscopy): None, Crystals -- Type: None, Colour : Brown, RBC's (Microscopy): None, Pus cells: None, Amount of yeast cells: None
>
> H. Pylori - Stool: Result: Positive
>
> Diagnosis: Helicobacter pylori gastritis
>
> Treatment: Paracetamol susp 250mg/5ml: 5 MLS, 3 Times a day, As directed for 3 Days
> Amoxicillin susp 250mg/5ml: 10 MLS, 2 Times a day, As directed for 14 Days
> Clarithromycin susp 250mg/5ml 50ml: 5 ml, 2 Times a day, As directed for 14 Days
> Omeprazole caps 20mg: 1 capsule, 1 Times a day, As directed for 14 Days

**Likert score**: 5
**Explanation**: Triple therapy with amoxicillin, clarithromycin and omeprazole is first line treatment for pediatric H pylori in this setting.

Example 2:

> 2y 5m
> presents with complaints of cough dry in nature of gradual onset with no relieving nor aggravating factors worse at night and in the morning, no associated dib, no chest congestion, no fastbreathing, no history of wheezing, no history of nightsweats, no loss of appetite or weight, no fevers, no runny nose, child feeding well, premeds-piriton
> on exam: fgc, alert, afebrile and actively playing, not in rs distress
> no jaundice, no pallor, no cyanosis, no dehydration, no edema
> heent- normal
> rs- no chest wall indrawing, chest clear on auscultation, no creps, no rhonchi
> cvs-s1, s2 heard, no murmurs
> p/a- not distended, soft non tender, no organomegally

> cns- alert, neck soft, no signs of meningeal irritation
> mss- unremarkable findings
> impression; bronchitis/allergic cough
> plan
> reassured
> Treatment as per t-sheet
> for paeds review if the complaints persist
> advised on hydration
> counselled on danger signs to watch
>
> Diagnosis: Acute bronchitis, Rhinitis
> Treatment: Desloratidine syp 60ml: 5 ml, 1 Times a day, As directed for 5 Days

**Likert score**: 4
**Explanation**: Desloratadine is not clearly indicated or fist line treatment in acute bronchitis or rhinitis unless caused by environmental allergies. While safe in this age group and the dosage is correct, this is likely an overtreatment of symptoms that may not be effective.

Example 3:

> 15y 3m
> Informant-Mother.
> C/C-Runny nose
> Cough
> Reports of above complains with ass fromtal headache.
> Also reports of ass nasal congestion.
> Reports headache superficial.
> No chest congestion.
> No sore throat
> No DIB
> Premeds-None.
> oe-In faIr general condition
> Afebrile
> well hydrated
> No cyanosis
> No wheeze
> RESP EXAM-No wheeze
> Noted rhonchi bibasal .
> ENT-Adenoid hypertrophy.
> dx-Acute bronchitis/Asthma.
> Acute sinusitis.
> tx as prescribed,
> Referred for chest xray.
> Review with results.
>
> Diagnosis: Acute Sinusitis, Acute bronchitis
> Treatment: Fluticasone Furoate Spray 120D: 1 dose, 2 Times a day, As directed for 2 Days
> Effervescent Paracetamol tabs 1000mg: 1 tablet, 3 Times a day, After food for 3 Days
> Azithromycin tabs 500mg: 1 tablet, 1 Times a day, After food for 3 Days

**Likert score 3**:
**Explanation**: Diagnosis of sinusitis and acute bronchitis are sometimes treated with antibiotics, though many cases are viral. No duration of symptoms documented which generally dictates whether to treat these conditions with antibiotics vs supportive care. Of the antibiotic choices, Azithromycin is not the first line for either condition.

Example 4:

> 7y 7m
> Vitals: hr 106, temp 37, wt 20.5 kg, rr 22, SpO2 98
> Cough productive irritative in the morning and at night
> ass with slight rhinorrhoea
> No fevers ,no chest congestion
> No loss of appetite
> No other ass symptoms
> premed None
> review of systems normal
> On exam stable
> RESP Chest clear no wheeze ,no creps
> ENT Normal
> DX Nasopharyngitis/Brochitis
> Plan
> Desloratadine
> Pdl
>
> Treatment: desloratadine: 5 ml, 1 At night for 5 Days
> Prednisolone Oral solution 1mg/5ml: 5 ml, 2 Times a day, As directed for 5 Days

**Likert score 2**:

**Explanation**: A child with a likely viral cough without documented history of asthma, without hypoxia or documented resp distress - there is no clear indication for oral steroids, a medication with significant potential side effects.

Example 5:

> 2y 2m
> brought in with above for past 2 days
> history of spiking fevers worse at night, no assc convulsions,
> history of throat pain worse when swallowing , no cough
> no runny nose
> history of abdominal pain worse after feeding , no vomiting , no diarrhoea
> premeds- none
> no history of travel to a malaria endemic zone
> o\e- ina fgc
> not in distress
> no pallor, no jaundice, no dehydrtaion
> rs- vesicular breath sounds, good airentry bilaterally , no lower chest wall indrawing
> ent- normal
> cns- alert, neck soft
> per abdomen- mwr, soft non tender, no palpable mas
> imp- febrile illeness ? spsis r\o amoebiasis
> plan
> fhg, ua stool for microscopy
> review with results
> ua- normal
> fhg- low level of hb- 9.4g\dl , mcv, mch, mchc
> unable to get the stool ample
> plan
> meds as prescribed
> advised on diet rich in iron , green veges and fruits
> to bring the stool sample later
> has brought the stool sample for microscopy
> review
> moderate pus cells , no oc
> plan
> meds as prescribed
> food and water hygien

> Stool Microscopy: RBC's (Microscopy): None, Result..: Abnormal, Crystals -- Amount: None, Amount of yeast cells: None, Colour : Brown, Yeast cells (Microscopy): None, Blood (Gross Appearance): Absent, Crystals -- Type: None, Parasites: No Ova/Cyst, Consistency : Semi formed, Pus cells: Moderate, Mucous: Absent
>
> Urine Analysis: Leukocytes: 0, RBC's (Urinalysis-Microscopy): None, Amount of yeast cells: None, Nitrate: Negative, Casts - type: None, Urobilinogen: 0, Ketones: 0, Bilirubin.: 0, Appearance (Urine Analysis): Clear, Crystals -- Amount (Microscopy): None , Glucose: 0, Trichomonads (Urinalysis-Microscopy): Absent, Parasites (Urine Microscopy): None, Crystals -- Type: None, Yeast cells (Microscopy): None, pH (Dipstick): 6.50, Blood (Dipstick): 0, Colour.: Amber, Pus cells: None, Proteins: 0, Result..: Normal, Epithelial Cells: None, Specific Gravity: 1.02
>
> Full Haemogram (FHG): Lymphocyte Percentage: 33.70, MCHC: 31.20, P-LCR: 13.10, P-LCC: 46.00, PCT: 0.26, RBC (Full Haemogram): 5.18, Neutrophill count: 5.46, WBC: 10.15, RDW-CV: 17.20, MCV: 58.20, Lymphocytes Count: 3.42, RDW-SD: 38.40, MCH: 18.10, HGB: 9.40, MPV: 7.50, Result..: Abnormal, Mid-granulocytes Count: 1.27, HCT: 30.10, Mid-granulocyte Percentage: 12.50, Neutrophill percentage : 53.80, Plt: 351.00, PDW: 8.30
>
> Azithromycin susp 200mg/5ml 15ml: 3 MLS, 1 Times a day, After food for 3 Days
> Ibuprofen/Paracetamol Susp 100mg/125mg 100ml: 7.5 MLS, 3 Times a day, After food for 3 Days
> Folic acid/Iron/Vitamin B12/Vitamin C syr 200ml: 2.5 ml, 2 Times a day, After food for 30 Days

**Likert score 1**:

**Explanation**: There is no indication for using Azithromycin and overuse/inappropriate use of antibiotics contributes to antibiotic resistance.



\end{document}